\UseRawInputEncoding
\documentclass[11pt]{article}

\usepackage[preprint]{acl}
\usepackage{booktabs}  % 负责画漂亮的“三线表”横线 (\toprule, \midrule, \bottomrule)
\usepackage{graphicx}  % 负责调整表格宽度 (\resizebox)
\usepackage{times}
\usepackage{latexsym}
\usepackage{rotating}
\usepackage{xcolor}
\usepackage{xstring}
\usepackage{tabularx}
\usepackage{array}
\usepackage{float} % ← 导言区加这行，提供 [H] 定位
\usepackage{placeins} % ← 导言区加这个用于 FloatBarrier
\usepackage{algorithm}
\usepackage{algpseudocode}
\usepackage[most]{tcolorbox} % “most” 包含 breakable、enhanced、skins 等
\tcbuselibrary{breakable}
\usepackage{amsmath,amssymb}
\usepackage{amsmath}  % 数学公式基础包（通常必备）
\usepackage{bm}       % 专门用于加粗希腊字母和数学符号的包 (Bold Math)
\usepackage{longtable}
\usepackage{booktabs}
\usepackage{refcount}
\usepackage{xcolor}
\usepackage{colortbl} % 用于排版优美的表格
\definecolor{negc}{gray}{0.6} % 定义负相关的灰色
\usepackage{cuted}   % 提供 strip 环境，双栏中插入全宽内容
\usepackage[utf8]{inputenc}
\usepackage{xcolor}    % tcolorbox 依赖颜色
\usepackage{tcolorbox} % 画盒子
\usepackage{caption}   % 支持 \captionof{figure}{...}
\usepackage{microtype}

\usepackage{inconsolata}

\usepackage{graphicx}
\newtcblisting{fullpromptlisting}[1]{
  breakable,
  enhanced,
  width=\textwidth,
  colback=gray!4,
  colframe=black!70,
  boxrule=0.5pt,
  arc=1mm,
  left=1.2mm,
  right=1.2mm,
  top=1mm,
  bottom=1mm,
  title={#1},
  fonttitle=\bfseries,
  before skip=8pt,
  after skip=8pt,
  listing only,
  listing engine=listings,
  listing options={
    basicstyle=\ttfamily\small,
    breaklines=true,
    columns=fullflexible,
    keepspaces=true,
    showstringspaces=false
  }
}
\title{HACHIMI: Scalable and Controllable Student Persona Generation via Orchestrated Agents}

\author{
  \textbf{Yilin Jiang}\textsuperscript{1,2}, 
  \textbf{Fei Tan}\textsuperscript{1}\thanks{Corresponding author.}, 
  \textbf{Xuanyu Yin}\textsuperscript{1}, 
  \textbf{Jing Leng}\textsuperscript{1}, 
  \textbf{Aimin Zhou}\textsuperscript{1,3}
  \\
  \textsuperscript{1}East China Normal University, Shanghai, China \\
  \textsuperscript{2}The Hong Kong University of Science and Technology (Guangzhou), Guangzhou, China \\
  \textsuperscript{3}Shanghai Innovation Institute, Shanghai, China \\
  \small{
    \texttt{ftan@mail.ecnu.edu.cn}
  }
}

\begin{document}
\maketitle
\begin{abstract}
Student Personas (SPs) are emerging as infrastructure for educational LLMs, yet prior work often relies on ad-hoc prompting or hand-crafted profiles with limited control over educational theory and population distributions. We formalize this as Theory-Aligned and Distribution-Controllable Persona Generation (TAD-PG) and introduce HACHIMI, a multi-agent \emph{Propose--Validate--Revise} framework that generates theory-aligned, quota-controlled personas. HACHIMI factorizes each persona into a theory-anchored educational schema, enforces developmental and psychological constraints via a neuro-symbolic validator, and combines stratified sampling with semantic deduplication to reduce mode collapse. The resulting HACHIMI-1M corpus comprises \textbf{1 million} personas for Grades 1--12. Intrinsic evaluation shows near-perfect schema validity, accurate quotas, and substantial diversity, while external evaluation instantiates personas as student agents answering CEPS and PISA~2022 surveys; across 16 cohorts, math and curiosity/growth constructs align strongly between humans and agents, whereas classroom-climate and well-being constructs are only moderately aligned, revealing a fidelity gradient. All personas are generated with Qwen2.5-72B, and HACHIMI provides a standardized synthetic student population for group-level benchmarking and social-science simulations. Resources available at \url{https://github.com/ZeroLoss-Lab/HACHIMI}.
\end{abstract}

\begin{comment}
    学生角色模型正成为教育领域大型语言模型的基础设施，但现有研究要么手工构建原型，要么临时性地向模型提供提示，难以控制理论契合度或群体层面的分布特性。我们将此挑战形式化为"理论对齐与分布可控的角色生成"（TAD-PG），并提出多智能体"提案-验证-修订"框架HACHIMI，该框架能生成理论锚定且配额可控的学生角色。HACHIMI将每个角色分解为高级教育模式，运用神经符号验证器强制约束，并通过分层采样与语义去重避免模式坍缩。生成的HACHIMI-1M语料库包含100万个1-12年级角色，附带标签及构念驱动的叙事文本。内在评估显示其模式有效性近乎完美，配额匹配精准，且具备显著多样性。在外部应用中，基于HACHIMI人物模型生成的智能体可完成CEPS及PISA-2022调查问卷；在16个分层样本组中，数学相关构念与好奇心构念与真实学生的相关系数均超过0.9。本研究首次构建了理论支撑、分布可控的人物模型语料库及框架，为教育领域大型语言模型的标准化群体评估提供了基础。
\end{comment}
\section{Introduction}

As artificial intelligence increasingly permeates the educational domain, particularly within personalized tutoring systems and teacher training simulations, the demand for high-quality, scalable Student Personas (SPs) has escalated into a piece of critical infrastructure. High-fidelity SPs serve as the cornerstone for driving educational dialogue simulation, evaluating the effectiveness of AI pedagogical strategies, and conducting virtual user testing~\cite{markel2023gpteach, zhang2025simulating}. However, the construction of student personas has traditionally been a fundamental bottleneck: it relies heavily on small-scale qualitative methods such as surveys, interviews, and direct observations~\cite{cooper1999inmates, pruitt2003personas}. While these manual methods offer deep insights, their prohibitive costs and inherent unscalability render them incapable of meeting the demands of modern data-driven research for simulating large-scale, highly heterogeneous student populations.

Advances in Large Language Models (LLMs) have ushered in a paradigm shift towards automated persona generation. LLM-based approaches have demonstrated potential in generating semi-structured, dynamic student personas~\cite{li2016persona, zhang2018personalizing}. However, this early generation paradigm quickly revealed limitations when attempting large-scale batch production: models are prone to \emph{intra-profile inconsistency}, i.e., self-contradiction and within-profile inconsistency within a single persona under one-shot long-context generation, often accompanied by instruction instability and formatting deviations~\cite{mundler2024self,li2024contradoc}. Although interventions such as Retrieval-Augmented Generation~\cite{lewis2020retrieval} and memory frameworks~\cite{park2023generative} can alleviate some consistency issues, they have not fundamentally addressed two deeper pedagogical challenges: the lack of cohort-level distributional control and explicit education-theory alignment.

Furthermore, existing educational datasets exhibit systematic misalignment on this issue. While works such as MathDial~\cite{macina2023mathdial} and Book2Dial~\cite{wang2024book2dial} have filled gaps in educational dialogue data through human-machine pairing or textbook-based generation, they prioritize \textit{interaction outcomes} over the \textit{personas themselves}. They remain insufficient in grounding personas within explicit educational frameworks (e.g., motivation, self-regulation, specific misconception patterns) and in controlling population-level distributions. These role-play or transcript-based datasets~\cite{stasaski2020cima, suresh2022talkmoves} further underscore the urgent need for methods to construct explicit student personas that are highly consistent, theory-aligned, and representative.

To address these challenges, we introduce HACHIMI, a theory-integrated multi-agent framework for student persona generation. This framework systematically addresses the three major challenges in persona generation through a collaborative \emph{Propose--Validate--Revise} workflow. HACHIMI internalizes structured pedagogical theories as Hard Constraints, anchors persona attributes to evidence, and enforces diversity and quota objectives to achieve cohort-level control. Empirically, HACHIMI achieves terriffic intrinsic controllability, and recovers cohort-level structure with a clear \textit{fidelity gradient} on CEPS and PISA 2022: alignment is strongest on school-facing constructs, yet weaker on latent well-being and family-dynamics patterns.

Our contributions are threefold: (i) We introduce a new task: Theory-Aligned and Distribution-Controllable student persona generation, formally defining the necessary constraint frameworks and distributional objectives. (ii) We propose HACHIMI, a multi-agent “Propose-Validate-Revise” framework that automatically fuses educational theory validation with diversity governance to ensure both pedagogical validity and compliance of the personas. (iii) We release a large-scale, theory-driven student persona dataset generated by this framework and show its utility for downstream educational dialogue data synthesis, connecting to broader evidence that synthesis quality and synthesis policy can matter as much as raw quantity in synthetic-data pipelines~\cite{zhao2025more,zhan2026mathsmith}.

\section{Related Work}
\subsection{Classic Student Persona Modeling}
In the educational field, the construction of learner personas has traditionally drawn upon methodologies from HCI and instructional design, whereby a limited set of archetypal student profiles are crafted to inform system development, curriculum design, and simulation studies. Early work employed qualitative instruments such as interviews, surveys, and observational data to generate semi-structured narrative profiles characterized by dimensions such as learner motivation, prior knowledge, and contextual constraints~\cite{cooper1999inmates, pruitt2003personas, carroll2000five}.

With the emergence of learning analytics, persona construction evolved to incorporate data-driven clustering of learner behaviors (e.g., in MOOC platforms), thereby enabling more representative archetypes~\cite{kizilcec2013deconstructing, breslow2013studying}. Concurrently, HCI research introduced scalable techniques for persona generation by leveraging telemetry and click-stream data to establish archetypal learner segments~\cite{zhang2016data, jung2018apg, jansen2022data, salminen2022use, nielsen2019personas}. However, despite methodological advances, traditional student persona generation frameworks remain constrained by static representations, reliance on manual curation, and limited capacity for large-scale, behaviorally consistent synthetic learner generation.
\begin{comment}
在教育领域中，学习者角色的构建传统上借鉴了人机交互与教学设计的方法论，通过构建有限数量的典型学生画像来指导系统开发、课程设计及模拟研究。早期研究采用定性工具——包括访谈、问卷和观察数据——生成半结构化叙事型画像，其特征维度涵盖学习动机、先验知识及情境限制等要素~\cite{cooper1999inmates, pruitt2003personas, carrol1999five}。
随着学习分析技术的兴起，学习者画像构建发展为基于数据驱动的学习者行为聚类（例如在MOOC平台中），从而能够构建更具代表性的典型画像~\cite{kizilcec2013deconstructing, breslow2013studying}。与此同时，人机交互研究通过利用遥测数据和点击流数据建立典型学习者分群，引入了可扩展的角色模型生成技术~\cite{zhang2016data, jung2018automatic, jansen2022data, salminen2022use, nielsen2013personas}。然而，尽管方法论取得进展，传统学生角色生成框架仍受限于静态表征、依赖人工整理，且在生成大规模行为一致的合成学习者方面能力有限。
\end{comment}
\subsection{Student Persona Generation via LLMs}
\label{sec:Student Persona Generation via LLMs}
Initial LLM-based persona generation typically uses direct role prompting, e.g., instructing the model to \texttt{act as a student} or “write a learner profile”~\cite{li2016persona, zhang2018personalizing}. More broadly, prompt design and prompt-based adaptation have been shown to substantially affect zero-shot behavior, controllability, and task transfer across NLP settings~\cite{lu2023what,liu2023deeply,lu2023punifiedner}. Although highly scalable, this method frequently suffers from self-contradiction in one-shot long-context generation.~\cite{li2024contradoc,mundler2024self}. Subsequent work introduced schema-based controls (e.g., contradiction-aware rewriting, trait-consistency filters)~\cite{song2020generate, kim2020will}, and retrieval-augmented generation (RAG) to anchor persona attributes in external evidence~\cite{lewis2020retrieval}. Memory-augmented agent architectures preserve persona states across interactions for longitudinal coherence~\cite{park2023generative}. Reason-and-act pipelines coordinate writing, planning and tool use for multi-step synthesis~\cite{yao2023react}. Controlled decoding methods (e.g., PPLM, GeDi) impose attribute filters for safer, more population-constrained outputs~\cite{dathathri2020plug, krause2021gedi}. Yet, two key gaps remain: lack of cohort-level distributional control and absence of education-theory-anchored validation, which hinder large-scale, theory-aligned and reliable student persona generation.

\begin{comment}
早期基于大型语言模型的方案依赖直接角色提示，要求模型”扮演学生”或撰写半结构化个人简介，这种方法源于基于角色条件的对话建模~\cite{li2016persona,zhang2018personalizing}。尽管具有极强的可扩展性，但这类单次提示生成的个人简介往往连贯性薄弱且性格特征归因不稳定。为解决此问题，后续研究引入基于模式或一致性的控制机制——例如矛盾感知重写或自我监控模块——以减少角色特质的漂移~\cite{song2020generate,kim2020will}。检索增强生成（RAG）技术进一步将人物档案属性锚定于外部证据而非自由补全，从而提升生成角色的真实度~\cite{lewis2020retrieval}。记忆增强型智能体框架通过跨交互维护学习者人格状态，从而保障学习过程中的纵向一致性~\cite{park2023generative}。针对大规模多步骤合成与评估，推理-行动管道协调人格构建、规划与工具使用~\cite{yao2022react}。并行发展中，受控生成模型通过属性过滤器或鉴别器引导解码，实现群体层级约束与更安全的输出~\cite{dathathriplug,krause2021gedi}。尽管取得这些进展，现有管道仍鲜少在大规模群体中整合分布式锚定或基于教育理论的验证机制。
\end{comment}

\begin{figure*}[t]
  \centering
  \resizebox{\textwidth}{0.48\height}{%
    \includegraphics{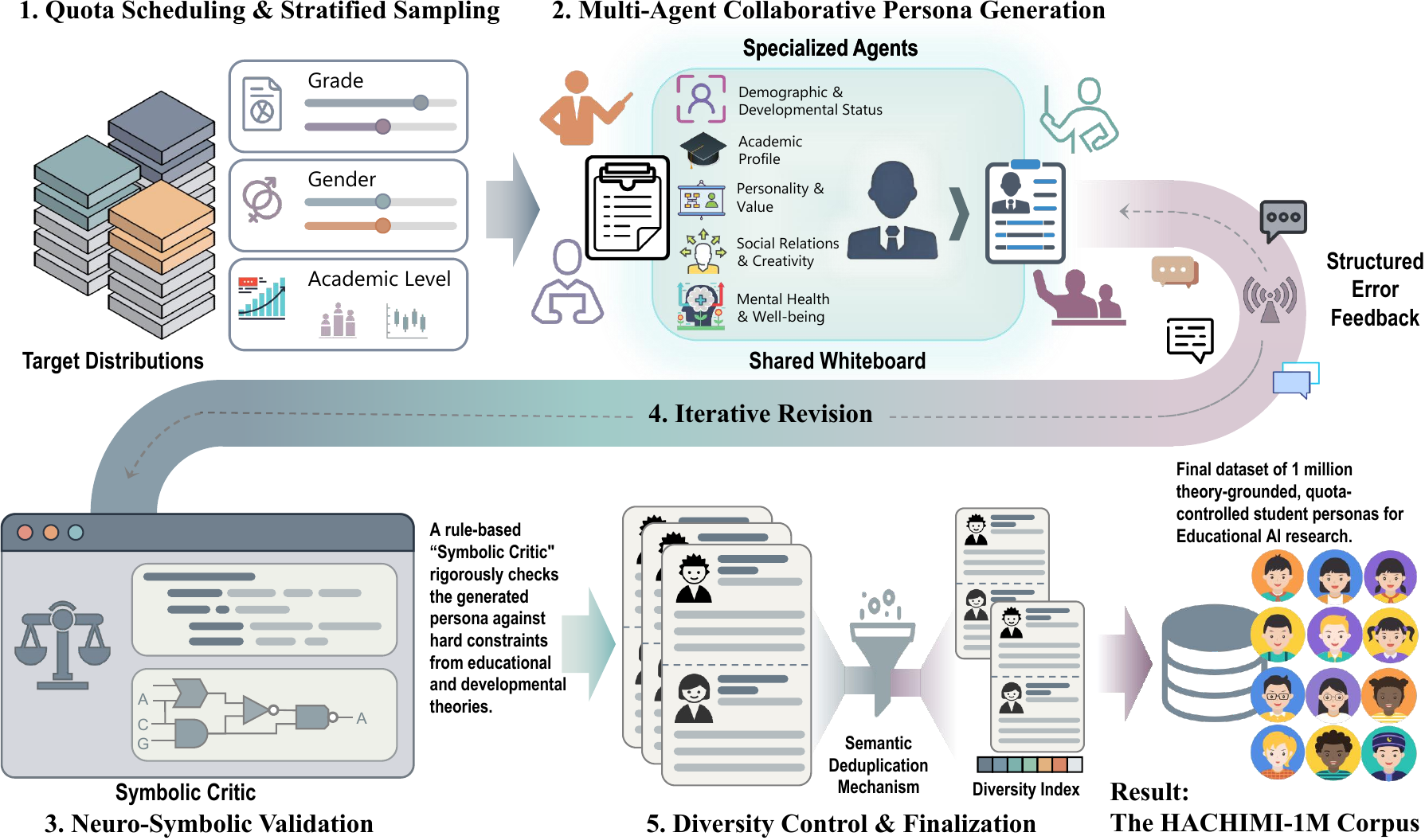}%
  }
  \caption{\textbf{HACHIMI pipeline overview.} From target distributions (grade/gender/academic level), steps (1)--(5) produce the HACHIMI-1M corpus.}
  \label{fig:hachimi-overview}
\end{figure*}

\subsection{Student Persona Datasets}
\label{sec:Student Persona Datasets}
High-fidelity student simulation hinges on high-quality student persona data, yet such resources remain scarce~\cite{wang2024book2dial}. MathDial addresses this gap by pairing real teachers with LLMs acting as students, collecting tutoring dialogues to simulate learner misconceptions and scaffolding behavior~\cite{macina2023mathdial}, while Book2Dial generates cross-subject educational conversations from textbook materials to alleviate data scarcity~\cite{wang2024book2dial}. Other datasets approximate learner personas through dual-role simulation~\cite{stasaski2020cima}, classroom transcription~\cite{suresh2022talkmoves,demszky2023ncte}, or real-world teacher–student chat logs~\cite{caines2020tscc}. While large-scale educational logs are abundant, such as the Open University Learning Analytics Dataset (OULAD)~\cite{kuzilek2017oulad}, ASSISTments’ online homework logs~\cite{selent2016assistments}, the PSLC DataShop from learning sciences~\cite{stamper2010datashop}, and EdNet’s self-directed learning traces~\cite{choi2020ednet}, they typically lack explicit schemas for learner personas. Student modeling efforts remain structurally rootless and difficult to validate in the absence of declarative constructs~\cite{farooq2025representing,tseng2024two}.
\begin{comment}
\subsection{学生角色数据集}
高保真学生模拟本质上依赖于高质量的学生角色数据，但此类资源仍十分稀缺~\cite{wang2024book2dial}。MathDial通过将真实教师与扮演学生的LLM配对，收集辅导对话以模拟学习者的误解和支架行为，从而弥补这一缺口~\cite{macina2023mathdial}；而Book2Dial则从教材材料中生成跨学科教育对话，以缓解数据稀缺问题~\cite{wang2024book2dial}。其他数据集通过双角色模拟~\cite{stasaski2020cima}、课堂转录~\cite{suresh2022talkmoves,demszky2023ncte}或记录真实师生对话~\cite{caines2020tscc}来近似学习者角色。尽管存在大量教育日志资源，如开放大学学习分析数据集（OULAD）~\cite{kuzilek2017oulad}、 ASSISTments在线作业日志~\cite{selent2016assistments}、学习科学领域的PSLC数据商店~\cite{stamper2010datashop}以及EdNet自主学习轨迹~\cite{choi2019ednet}——均缺乏针对学习者角色的明确模式。因此，在缺乏动机、自我调节或误解等声明性构造的情况下，学生建模工作仍处于结构性根基缺失的状态，难以进行有效验证~\cite{farooq2025representing,tseng2024two}。
\end{comment}

\section{HACHIMI Framework}
  \subsection{Problem Formulation}
  \label{sec:Problem Formulation}
    While prior work often treats persona generation as a byproduct of prompt engineering or dataset annotation, we explicitly define the task as Theory-Aligned and Distribution-Controllable Student Persona Generation (TAD-PG). This task requires producing a set of personas that are: (i) aligned with high-level educational objectives, (ii) internally coherent in discourse and traits, and (iii) matched to predefined population-level distributions.

  %   \label{sec:问题定义}尽管先前的研究常将人物角色生成视为提示工程或数据集标注的副产品，我们明确将该任务定义为理论对齐且分布可控的学生人物角色生成（TAD-PG）。该任务要求生成满足以下条件的人物角色集：(i) 与高层次教育目标保持一致，(ii) 在话语和特征上具有内在一致性，(iii) 匹配预定义的人口层级分布。

% This formulation motivates the design of our proposed HACHIMI framework, which systematically enforces these constraints during persona generation.
  \subsection{Theory-Anchored Persona Schema}
\label{sec:Theory-Anchored Persona Schema}

In designing student personas, we adopt \emph{high-level education} as the guiding orientation: LLM-based educational models should go beyond answer accuracy and knowledge mastery to support value formation, personalized support, creativity development, and mental well-being~\cite{kasneci2023chatgpt,durlak2011impact}. These dimensions are treated as core educational outcomes in both Chinese curriculum standards under the \emph{core-competencies–holistic development} paradigm~\cite{lin2017core} and the OECD vision of learner agency, creativity, and well-being~\cite{oecd2019learningcompass}.

To operationalize this vision at the student-persona level, we propose a theory-anchored persona schema that decomposes each persona into five complementary components based on the OECD Learning Compass~\cite{oecd2019learningcompass}: (1) \textbf{demographic \& developmental status}, covering grade level and major moral-development phases~\cite{kohlberg1977moral}; (2) \textbf{academic profile}, summarizing strong/weak subject clusters and an achievement tier; (3) \textbf{personality \& value orientation}, aligned with value-education and character-formation constructs~\cite{lapsley2006character,berkowitz2005works}; (4) \textbf{social relations \& creativity}, capturing interaction patterns and creative problem-solving characteristics~\cite{rubin2006peer,runco2004creativity,beghetto2014classroom}; and (5) \textbf{mental health \& well-being}, reflecting emotional functioning and support systems~\cite{weare2011mental,durlak2011impact}. See Appendix~\ref{app:persona-schema} for design details. Together, these five components provide observable carriers that \textit{instantiate} the four high-level educational capacities on the student side, spanning both relatively stable traits and interaction-driven states.
 
Within the TAD-PG task, the schema functions both as a structured constraint interface for theory-aligned, quota-controlled generation and as the state space for student agents, so that changes in values, personalization, creativity, and mental health can be read out as quantitative signals of high-level educational impact.

\begin{comment}
    \subsection{理论锚定型人物角色框架}
  在设计学生角色模型时，我们以”高层次教育”为根本导向：基于大型语言模型的教育模式不应仅追求答案准确性或知识掌握度的提升，而应服务于价值塑造、个性化支持、创造力发展及心理健康等深层目标~\cite{kasneci2023chatgpt,durlak2011sel}。在”核心能力-全面发展”范式下的多国课程标准~\cite{corecompetency2016,lin2017core}，以及经合组织提出的学习者自主性、创造力与幸福感愿景~\cite{oecd2019learningcompass}中，这些维度均被定位为核心教育成果而非附属副产品。

  为将此宏观愿景落实到学生个体层面，我们提出基于理论的个性化模型框架。每个学生模型均分解为五个互补维度~\cite{corecompetency2016,oecd2019learningcompass}：(1) \textbf{人口统计与发展状态}，涵盖年级水平及主要道德发展阶段~\cite{kohlberg1981moral}; (2) \textbf{学业档案}，通过优势/薄弱学科集群与成就等级概括学习状态；(3) \textbf{人格与价值取向}，契合当代价值教育与品格塑造理论框架~\cite{lapsley2006character,berkowitz2005character}; (4) \textbf{社交关系与创造力}，涵盖互动模式与创造性问题解决特征~\cite{rubin2006peer,runco2004creativity,beghetto2014classroom}; 以及 (5) \textbf{心理健康与幸福感}，反映情绪功能与支持系统~\cite{weare2011mentalhealth,durlak2011sel}。设计细节详见附录\ref{app:persona‐schema}。这些组件共同构成学生端四项高阶教育能力的可观测载体：既包含相对稳定的特征，也涵盖可通过交互演变的状态。

在TAD-PG任务中，该人格模式既作为理论驱动、配额控制生成机制的结构化约束接口，亦作为学生代理的状态空间，使价值观、个性化、创造力及心理健康的变化能够转化为高阶教育影响的量化信号。
\end{comment}

\subsection{Multi-Agent Generation Architecture}
\label{sec:architecture}

We cast student persona generation as a \textit{constrained optimization task} that reconciles LLM flexibility with the requirements of educational theory. The HACHIMI architecture employs a collaborative society of agents governed by three mechanisms, ensuring educational validity and structural consistency. Figure~\ref{fig:hachimi-overview} gives an end-to-end overview; here we describe the three mechanisms, with stage-wise implementation details in Appendix~\ref{app:hachimi-overview}. We also provide the canonical prompt templates for all generation agents, together with the shared preamble, validator prompts, and responsible-field mapping, in Appendix~\ref{app:agent-prompts}.

\begin{comment}
3.3 多智能体生成架构

我们将学生画像的生成视为一项约束优化任务，旨在协调大语言模型（LLM）的生成灵活性与教育心理学的严谨要求。HACHIMI 架构采用了一个由三种核心机制驱动的协作智能体群，以确保教育学上的有效性和结构上的一致性。
\end{comment}

\paragraph{Mechanism I: Modular Generation via Shared Whiteboard.}
Generating a \emph{holistic student} in a single pass often leads to \textit{intra-profile inconsistency} in long contexts, where different parts of the same persona may become self-contradictory or semantically misaligned~\cite{li2024contradoc,mundler2024self}. We therefore factorize personas into the five components in \S~\ref{sec:Theory-Anchored Persona Schema}. To avoid fragmentation across independently generated components, we anchor all agents to a \textbf{Shared Whiteboard}~\cite{park2023generative}, a dynamic context that lets agents condition sequentially on peers’ intermediate states.

\begin{comment}
机制一：基于共享白板的模块化生成

在单一的潜在空间中建模”全人学生”往往会导致内在属性不一致”现象，包括同一角色内的自我矛盾和叙事断裂等问题 [Xi et al., 2023]。为了缓解这一问题，我们将画像空间因子分解为 3.2 节中定义的五个正交子域（如学业、心理社会、价值观）。
然而，独立的生成方式面临逻辑割裂的风险。我们通过将所有智能体锚定到一个共享白板（Shared Whiteboard）上来解决这一问题 [Park et al., 2023]。这个动态上下文窗口允许智能体按顺序基于同伴的状态调节自身输出。例如，心理智能体会显式关注白板上的学业智能体输出，以合成与学生学业层级因果一致的压力水平。这种方法有效地维护了跨维度的一致性，防止了在非连通多智能体系统中常见的语义矛盾。
\end{comment}

\paragraph{Mechanism II: Neuro-Symbolic Constraint Satisfaction.}
To mitigate stochastic LLM hallucinations, we adopt a \textbf{Propose-Validate-Revise} workflow inspired by iterative self-correction~\cite{madaan2023self}. Neural agents generate \emph{creative narratives}, while a rule-based Validator acts as a \textbf{Symbolic Critic} that rigorously checks drafts against educational axioms. Unlike simple heuristic filtering, we formalize theoretical alignment as satisfaction of logical predicates: demographic attributes are mapped to developmental stages (Piagetian~\cite{piaget1964cognitive}; Eriksonian~\cite{erikson1963childhood}) via \emph{deterministic topology}, and trait descriptions are validated with set-theoretic logic. Upon violations, the system returns structured error signals to the relevant agents and iterates until all constraints are satisfied. We implement this critic as a rule-based Validator that checks the generated profile against an executable constraint set. The full executable rule set (R1--R15) is reported in Appendix~\ref{app:validator-rules}.

\paragraph{Mechanism III: Stratified Sampling and Diversity Control.}
To mitigate \emph{mode collapse} (i.e., models converging to generic, averaged personas), we replace random sampling with \textbf{Stratified Sampling}. We enforce a uniform distribution over academic proficiency strata, ensuring balanced representation across student groups and serving as a \textbf{Conditional Variable} that propagates influence to downstream attributes (e.g., self-efficacy). To further maximize diversity, we apply \textbf{Semantic Deduplication} via Locality-Sensitive Hashing (LSH)~\cite{charikar2002similarity}, mapping narratives into hash space and strictly removing semantically redundant duplicates to preserve a heterogeneous student population.

\begin{comment}
机制三：分层采样与多样性控制

为了对抗”模式坍塌”——即模型收敛于平庸、同质化的画像，我们用分层采样（Stratified Sampling）取代了随机采样。我们在不同的学业能力阶层上强制执行均匀分布，并将其作为条件变量，将因果影响传播至后续属性（如自我效能感），确保了从精英学生到后进学生的均衡代表性。
此外，为了最大化语料库的多样性，我们应用了基于局部敏感哈希（LSH）的语义去重（Semantic Deduplication）机制 [Charikar, 2002]。通过将生成的叙述映射到高维哈希空间，我们严格修剪表现出高语义冗余的画像，从而有效地确保数据集捕捉到高度异质化的学生群体。
\end{comment}

\subsection{The HACHIMI-1M Corpus}
\label{sec:dataset}

As the primary artifact of our framework, we present the \textbf{HACHIMI-1M Corpus}, comprising \textbf{1,000,000} synthetic student personas spanning Grades 1 through 12. To the best of our knowledge, this represents the largest publicly available dataset of student profiles explicitly anchored in educational theory. 

We include a persona example in Appendix~\ref{app:persona-sample}. All personas are generated with Qwen2.5-72B, and HACHIMI-1M can be produced at scale with \emph{$\sim$3{,}200 H100 GPU-hour} compute (Appendix~\ref{app:gen-eff}).

\begin{comment}
3.4 HACHIMI-1M 语料库
作为我们框架的主要产出，我们展示了 HACHIMI-1M 语料库，包含跨越 1-12 年级的 1,000,000 个合成学生画像。据我们所知，这是目前最大的、明确基于教育理论的学生档案公开数据集。
\end{comment}

\paragraph{Hybrid Semi-Structured Representation.}
Unlike traditional datasets comprised of interaction logs or purely unstructured text, HACHIMI-1M employs a \textbf{hybrid semi-structured format}. Each entry integrates two distinct data types:
\begin{enumerate}
    \item \textbf{Categorical Labels:} Deterministic attributes for Piagetian and Eriksonian developmental stages and academic tiers. These serve as structured metadata for filtering and retrieval.
    \item \textbf{Construct-driven Narratives:} Natural language descriptions for complex attributes drawn primarily from the \emph{personality \& value orientation}, \emph{social relations \& creativity}, and \emph{mental health \& well-being} components of our schema. These narratives are conditioned on specific psychological constructs, ensuring interpretability and theoretical consistency.
\end{enumerate}

\begin{comment}
混合半结构化表征
与由交互日志（如点击流）或纯非结构化文本组成的传统数据集不同，HACHIMI-1M 采用”混合半结构化格式”。每个条目整合了两种不同的数据类型：
    \item \textbf{分类标签：} 用于表征皮亚杰与埃里克森发展阶段及学术层级的确定性属性。这些标签作为结构化元数据，用于信息筛选与检索。
    \item \textbf{构念驱动叙事：} 针对价值观、创造力及心理健康等复杂属性的自然语言描述。这些叙事基于特定心理构念生成，确保可解释性与理论一致性。
\end{comment}

\paragraph{Distribution via Stratified Sampling.}
To ensure robust coverage across the student ability spectrum, we strictly employ a \textbf{Stratified Sampling} strategy. We enforce a uniform distribution ($\sim$250{,}000 profiles per tier) across the four academic proficiency levels defined in \S \ref{sec:architecture}. This approach effectively ensures that underrepresented groups, like struggling learners, are oversampled relative to their real-world frequency. This balanced distribution provides a standardized benchmark for evaluating the capabilities of educational AI systems. Offline descriptive summaries of the realized Grade $\times$ Academic Level and Grade $\times$ Gender distributions are reported in Appendix~\ref{app:hachimi-dist-summary}.

\section{Evaluation Methods}

\subsection{Evaluation Goals and Research Questions}
\label{sec:eval-goals}

Our evaluation aims to answer a central question: \emph{to what extent do HACHIMI personas and persona-based student agents behave like real students at the group level while satisfying the TAD-PG task?}

We therefore focus on three complementary goals:
(i) checking whether the generated persona collection satisfies the theory-anchored schema and quota targets,
(ii) testing group-level consistency with the China Education Panel Survey (CEPS)~\cite{ceps_followup_2014_2015}, and
(iii) investigating their transferability across broader geographical regions on the PISA 2022 dataset~\cite{oecd2023pisa2022db}.

Based on these goals, we formulate the following research questions:

\textbf{RQ1}: Without relying on external data, does HACHIMI produce persona collections that respect the theory-anchored schema, match target quotas for grade, gender, academic achievement and psychological risk, and remain semantically diverse rather than collapsing into a few templates?

\textbf{RQ2}: When personas are instantiated as student agents in a CEPS Grade~8 shadow survey, do agent cohorts reproduce the item-wise relative differences across real-student cohorts?

\textbf{RQ3}: Do similar group-level consistency patterns hold on PISA~2022 constructs across world regions, suggesting that HACHIMI captures educational regularities rather than overfitting to CEPS?

\subsection{Intrinsic Evaluation of TAD-PG Persona Collections}
\label{sec:intrinsic-eval}

Before bringing in real student data, we first evaluate the HACHIMI persona collection itself with a lightweight offline evaluator. At the profile level, we distinguish \emph{hard errors} (e.g., missing required fields, invalid fixed-label formats, or clear age--grade/developmental-stage violations) from \emph{warnings} (e.g., partial dimension coverage, mild formatting problems, or softer consistency mismatches). We define \textbf{schema validity} as the fraction of personas that contain no hard errors, and we additionally report the warning rate. At the corpus level, we measure \textbf{quota satisfaction} by comparing the empirical distributions of grade, gender, academic level, and psychological risk against the scheduler targets; \textbf{diversity} by corpus-level lexical diversity indicators; and \textbf{redundancy} by approximate near-duplicate detection over long-text fields. All metric definitions, thresholds, and implementation details are reported in Appendix~\ref{app:intrinsic-eval}.

\subsection{Shared Protocol for External Group-level Evaluation}
\label{sec:shared-external-protocol}

Both CEPS and PISA follow the same external evaluation logic: we first construct cohort-level reference patterns from real-student data, then instantiate HACHIMI personas as student agents, aggregate their shadow-survey responses by matched cohorts, and finally compare the two sides at the cohort level.

\begin{figure}[t]
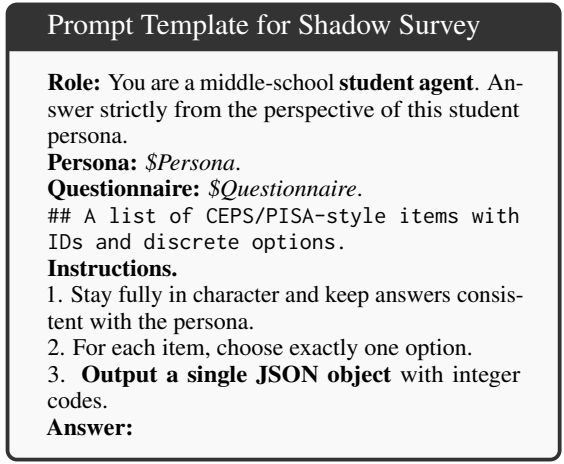

\centering
\begin{tcolorbox}[
  width=0.95\linewidth,
  colback=gray!5,
  colframe=black!80,
  title=Prompt Template for Shadow Survey
]
\footnotesize
\textbf{Role:} You are a \textbf{student agent}. Answer strictly from the perspective of this student persona.\\
\textbf{Persona:} \textit{\$Persona}.\\
\textbf{Questionnaire:} \textit{\$Questionnaire}.\\
\texttt{\#\# A list of CEPS/PISA-style items with IDs and discrete options.}\\
\textbf{Instructions.}\\
1. Stay fully in character and keep answers consistent with the persona.\\
2. For each item, choose exactly one option.\\
3. \textbf{Output a single JSON object} with integer codes.\\
\textbf{Answer:}
\end{tcolorbox}
\caption{Immersive role-playing prompt template used for HACHIMI student agents when answering CEPS- and PISA-based shadow surveys.}
\label{fig:prompt_template}
\end{figure}

\paragraph{Persona-side instantiation and aggregation.}
On the HACHIMI side, personas are not grouped by subjective reading of free-form narratives. Instead, for each evaluation setting, we first filter the corpus to the required scope, then assign each eligible persona to the target cohort (s) using validated structured fields together with fixed parsing rules, and finally perform stratified quota sampling to obtain structurally matched agent populations. Each sampled persona is instantiated as a DeepSeek-V3.2-based student agent and prompted under the immersive role-playing scheme in Figure~\ref{fig:prompt_template} to complete the corresponding shadow survey. The responses are mapped back to numeric codes and aggregated by cohort.

\paragraph{Shared consistency metrics.}
For any evaluated item or construct, the real-student side and the agent side each yield a cohort-level mean vector. We compare these vectors using Pearson correlation $r$ to measure alignment in absolute group means (linear trend consistency), and Spearman correlation $\rho$ to measure alignment in the relative ordering of cohorts (rank-order consistency). The two datasets differ in how cohorts and targets are defined, but they share this comparison protocol.

\subsection{Group-level Consistency on CEPS (Chinese Grade 8)}
\label{sec:ceps-consistency}

We use the CEPS Grade~8 data as a nationally representative real-student reference and test whether HACHIMI agent cohorts reproduce the corresponding group-level regularities.

\paragraph{Real-student cohorts on CEPS.}
On the CEPS side, we focus on Grade~8 students and derive three stratification variables: academic achievement level (high / medium / low / poor), gender (M / F)\footnote{\label{fn:mf}M/F follows the survey coding, used only for cohort matching; not intended to imply gender is binary.}, and psychological risk (high / low) from the CES-D depression scale in CEPS. Combining these three axes yields $4 \times 2 \times 2 = 16$ mutually exclusive cohorts. For each CEPS item selected for evaluation, we compute the mean response within each cohort, yielding a 16-dimensional vector of cohort means that serves as the real-world reference pattern for that item. Detailed preprocessing, variable construction, and cohort definitions are given in Appendix~\ref{app:ceps-pipeline-preprocessing}.

\paragraph{Persona-side CEPS cohort assignment and CEPS-based shadow survey.}
On the persona side, we first retain only Grade~8 personas. Each retained persona is then deterministically assigned to one of the 16 CEPS-style cohorts by reading its validated \texttt{Academic Level} and \texttt{Gender} fields and by extracting the psychological-risk signal from the validated \texttt{Mental Health} field under fixed parsing rules that map the text to the same binary high/low risk split used for cohort matching. Only personas whose three stratification variables can be stably parsed are included in the CEPS analysis. We then perform stratified quota sampling and draw a fixed sample of $200$ personas for each cohort. To obtain behaviour that is comparable to CEPS, we construct a shadow survey by selecting Grade~8 CEPS items that can be reasonably inferred from a textual persona. We retain attitudinal, perceptual, and self-reported behavioural items, and exclude purely factual items that cannot be inferred from the persona. The sampled personas are then instantiated and evaluated under the shared protocol in Section~\ref{sec:shared-external-protocol}, producing agent-side cohort statistics in exactly the same format as the human data.

\paragraph{CEPS comparison target.}
Under this setup, each CEPS shadow-survey item yields one human-side and one agent-side 16-dimensional cohort-mean vector. We apply the shared consistency metrics above and report the resulting correlation distributions in \S~\ref{sec:rq2-results}. Low-level implementation details remain in Appendix~\ref{app:ceps-pipeline-CMdetails}.

\subsection{Group-level Consistency on PISA 2022}
\label{sec:pisa-consistency}

We next investigate whether the CEPS-based consistency patterns transfer to an international assessment context on the PISA~2022 student dataset~\cite{oecd2023pisa2022db}. The overall comparison protocol is the same as above, while the main differences lie in the choice of constructs, regional grouping, and cohort definition.

\paragraph{Real-student constructs and cross-regional cohorts.}
On the PISA side, we select indices capturing students' motivation, affect, well-being, and creativity. Countries are grouped into five macro-regions (East Asia, Western Europe, Southern Europe, Latin America, Middle East), and within each region we form $4 \times 2 \times 2 = 16$ cohorts by crossing academic achievement quartiles, gender (M / F), and a binary psychological-risk indicator derived from the mental-health scale in PISA. For every region--construct pair, we compute unweighted cohort means, yielding a 16-dimensional cohort-mean vector as the real-student reference pattern. See Appendix~\ref{app:pisa-vars-preprocessing},~\ref{app:pisa-cohorts} for variable lists and thresholds.

\paragraph{PISA-based shadow survey.}
For each construct, we design a short PISA-based shadow survey by translating one or a few representative items into Chinese while preserving the original response scales. HACHIMI personas are then evaluated under the shared protocol in Section~\ref{sec:shared-external-protocol}, with PISA-side grouping and matched cohort sampling, and responses are aggregated by region and cohort. See Appendix~\ref{app:pisa-shadow} for item selection and coding.

\paragraph{PISA comparison target.}
For each region and construct, this procedure yields one human-side and one agent-side 16-dimensional cohort-mean vector. We apply Pearson and Spearman correlations and report the resulting cross-regional distributions in \S~\ref{sec:rq3-pisa}; see Appendix~\ref{app:pisa-shadow} for aggregation and implementation details.

\section{Results}

We report results from \emph{inside-to-outside}. We first examine whether the generated persona collection itself satisfies the TAD-PG requirements (RQ1). We then move to external evaluation, first in a single national educational setting with CEPS (RQ2), and then in a cross-regional setting with PISA 2022 (RQ3). Finally, we compare HACHIMI against a protocol-matched one-shot baseline to isolate the gains brought by the generation framework itself.

\subsection{Intrinsic Properties of TAD-PG Persona Collections}

We begin with RQ1, which asks whether HACHIMI can generate persona collections that are structurally valid, quota-faithful, and semantically diverse before any comparison with real-student data. To answer this, we run the intrinsic evaluator (Appendix~\ref{app:intrinsic-eval}) on a random, corpus-wide sub-corpus of $\sim$150{,}000 HACHIMI personas.

The results are uniformly strong. \textbf{Schema validity is near-perfect}: no sampled persona triggers any hard error, and only $0.06\%$ receive soft warnings. At the corpus level, \textbf{quota targets are met almost exactly} (KL $\approx 0$), indicating that the scheduler-controlled generation process remains tightly aligned with the intended marginal distributions. At the same time, \textbf{diversity remains high}: distinct-1/2 are approximately $0.40/0.83$, no SimHash near-duplicates are detected, and cross-component overlap remains low. In addition, \textit{anchor alignment} between academic tier and the values / creativity / mental-health components is consistently strong, suggesting that the long-text descriptions track the intended academic-level conditioning in a stable and interpretable way.

Taken together, these findings provide a positive answer to \textbf{RQ1}: HACHIMI does not merely generate well-formed personas, but generates a corpus that simultaneously satisfies structural constraints, realizes the target quota design, and avoids semantic collapse. See Appendix~\ref{app:intrinsic-eval-results} for full results.

\subsection{Group-level Consistency on CEPS}
\label{sec:rq2-results}

We next move to external evaluation in a single educational system and ask whether HACHIMI personas, once instantiated as student agents, can recover the group-level structure observed in real CEPS Grade~8 students. Following Section~\ref{sec:ceps-consistency}, we evaluate a set of CEPS targets that combine construct aggregates and focal items: six indices aggregated from related items (depression, parental strictness, teacher attention, misbehaviour, prosocial behaviour, and school bonding) and five stand-alone central items (aspiration, future confidence, parental expectations / pressure, and self-rated health). For each target, we correlate the 16-dimensional cohort-mean vectors between students and agents; see Appendix~\ref{app:ceps-constructs} for definitions.

\begin{figure*}[t!] 
\centering % e.g., two overlaid histograms or violin plots for Pearson and Spearman 
\includegraphics[width=0.77\linewidth]{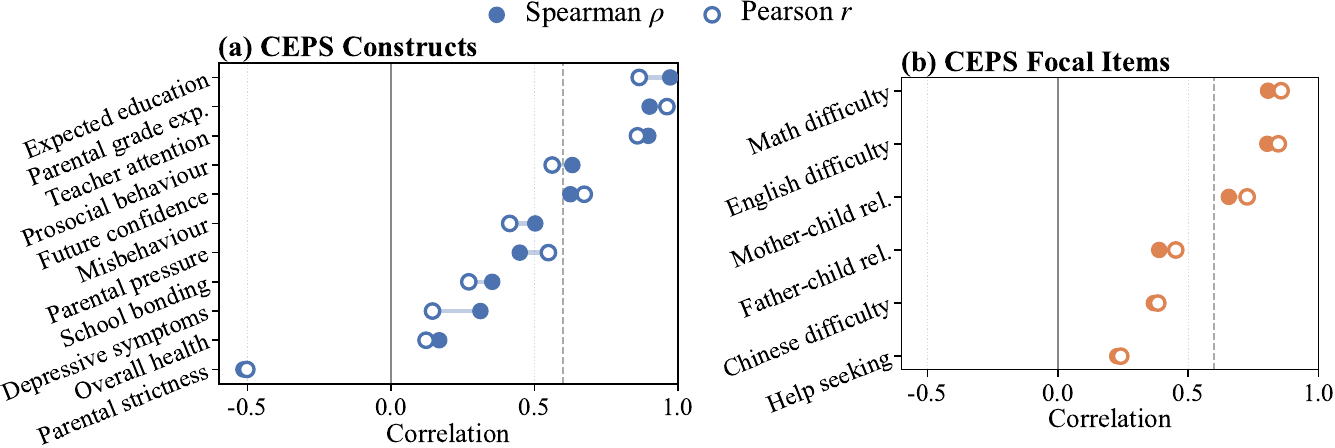} 
\caption{Pearson $r$ and Spearman $\rho$ between human and HACHIMI cohort means for each CEPS target.} 
% 图标题翻译：人类与 HACHIMI 群体均值在基于 CEPS 的构念和题目上的皮尔逊和斯皮尔曼相关系数分布。 
\label{fig:ceps-corr-hist} 
\end{figure*}

Figure~\ref{fig:ceps-corr-hist} summarizes these correlations for CEPS constructs and focal items. The pattern is clear: \textbf{alignment is strongest for school-facing and academically grounded constructs}. Educational aspirations (\texttt{w2b18}) and parental achievement expectations (\texttt{w2a27}) are highly aligned (Spearman $\rho \geq 0.90$, Pearson $r \geq 0.86$), and aggregated teacher attention is similarly strong (Spearman $\rho \approx 0.90$, Pearson $r \approx 0.86$). Prosocial behaviour and future confidence reach moderate-to-high consistency (Spearman $\rho \approx 0.63$), while misbehaviour frequency and parental-expectation pressure remain in the moderate range.

By contrast, constructs tied more closely to latent well-being and family dynamics are harder to recover from static personas. School bonding, depressive symptoms, self-rated health, and especially parental strictness show weak or even negative correlations, indicating that these dimensions are less directly inferable from the persona state used in the shadow-survey setting.

The same pattern appears at the item level. Perceived difficulty in mathematics (\texttt{w2b02}) and English (\texttt{w2b04}) is highly aligned (Spearman $\rho \approx 0.81/0.80$, Pearson $r \approx 0.86/0.85$), and mother--child relationship quality (\texttt{w2a23}) is also aligned (Spearman $\rho \approx 0.66$, Pearson $r \approx 0.73$). This suggests that the agents can recover cohort-level gradients in academic stress and some coarse-grained relational signals, even when alignment is not equally strong across all psychosocial constructs.

Overall, the CEPS results provide a positive answer to \textbf{RQ2}: when instantiated as student agents, HACHIMI personas recover a substantial portion of the group-level structure seen in real Grade~8 students. At the same time, they reveal a clear \textit{fidelity gradient}: alignment is strongest for observable, school-facing constructs, and weaker for more latent well-being and family-related patterns.

\subsection{Cross-regional Consistency on PISA 2022}
\label{sec:rq3-pisa}

We then ask whether the CEPS pattern generalizes beyond a single country and survey system. Using the protocol in Section~\ref{sec:pisa-consistency}, we evaluate human--agent consistency across five PISA 2022 regions.

Across regions, human--agent cohort correlations are decisively above chance for most constructs. Math-related and curiosity / growth constructs usually exceed $0.6$ and often exceed $0.8$: \texttt{MATHEFF} stays at $r > 0.95$ across all five regions, and \texttt{CURIOAGR} is similarly high ($r \gtrsim 0.85$). At a broad level, the correlations exhibit a tiered profile: very strong for math and curiosity-related constructs, moderate for classroom climate and belonging, and weak or negative for several well-being and workload indicators. This closely mirrors the CEPS pattern in \S~\ref{sec:rq2-results}, suggesting that the same \textit{fidelity gradient} persists across datasets rather than being specific to one survey source.

\begin{figure}[t]
  \centering
  \includegraphics[width=.95\linewidth]{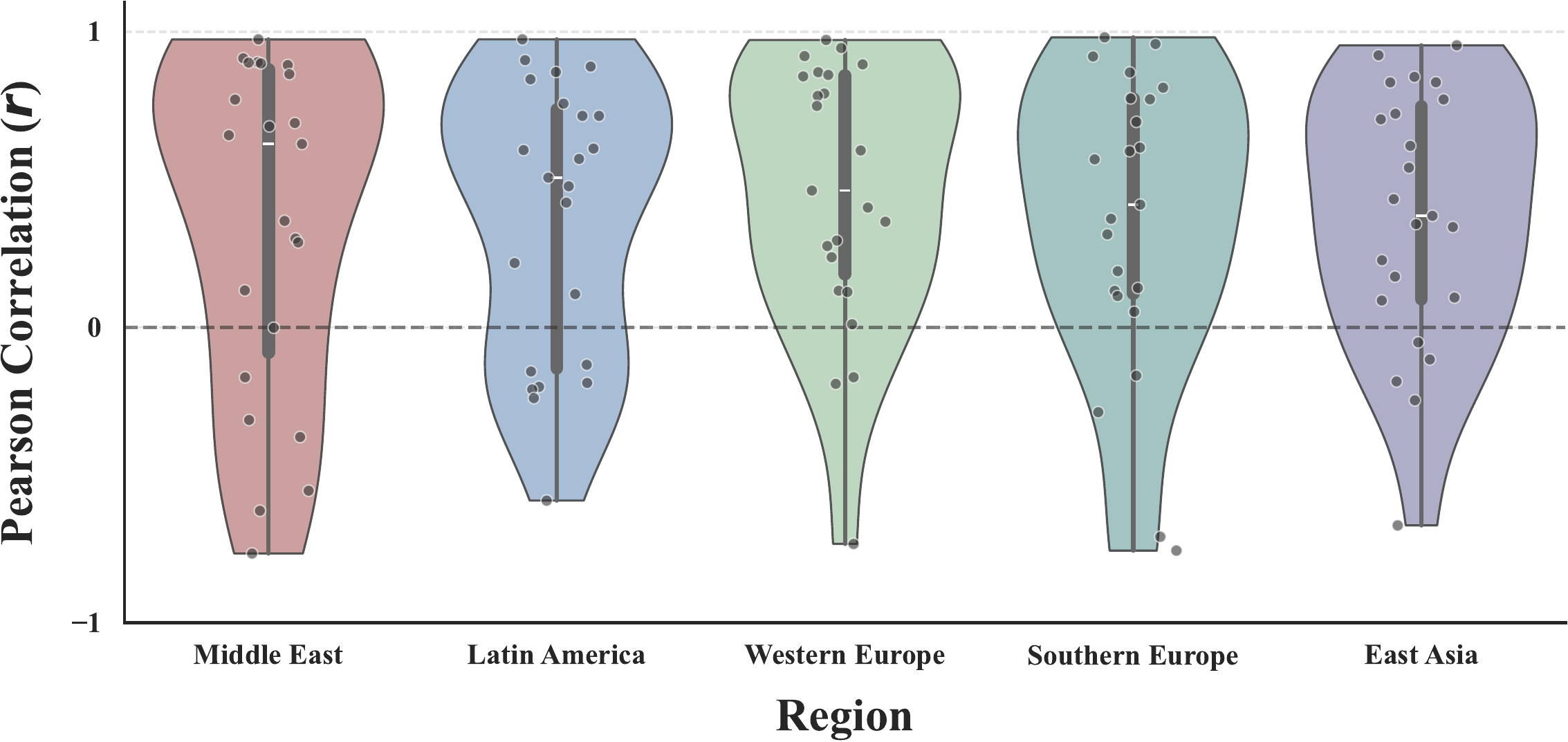}
  \caption{Distribution of Pearson correlations between human and agent group means on PISA 2022.}
  \label{fig:pisa-region-box}
\end{figure}

At the same time, PISA makes cross-regional structure more explicit. Math engagement and efficacy constructs show strong alignment in all regions (mostly $r > 0.7$), indicating that agents preserve the cohort ordering both \emph{within} regions and \emph{across} regions over the same 16 gender $\times$ academic-achievement $\times$ risk cohorts. Curiosity / growth constructs are aligned everywhere, although somewhat lower in East Asia and Southern Europe than in Latin America and the Middle East. By contrast, mental-health indices cluster around $r \approx 0$, workload and work--home balance variables are systematically negative, and some classroom-exposure constructs even flip sign across regions.

Figure~\ref{fig:pisa-region-box} summarizes these regional distributions, while full per-construct results are reported in Appendix~\ref{app:pisa-results}. Overall, the PISA 2022 analysis supports \textbf{RQ3}: agent-based personas reproduce a stable cross-regional structure, with \textbf{strong alignment on math-related and curiosity constructs}, but much weaker alignment on latent well-being, workload, and some classroom-exposure variables.

\subsection{Protocol-matched Baseline Comparison}
\label{sec:baseline-comparison}

Finally, to isolate the gains brought by the HACHIMI framework itself rather than by differences in evaluation protocol, we compare HACHIMI against a one-shot baseline under a matched protocol.

Compared with the one-shot baseline on matched 10{,}000 personas, HACHIMI \textbf{eliminates hard errors} and \textbf{improves diversity by a large margin}, as shown in Table~\ref{tab:baseline-mini}. The baseline still exhibits substantial structural instability (12.03\% hard errors; 25.33\% warnings), whereas HACHIMI reduces these to 0.00\% and 0.82\%, respectively. At the same time, HACHIMI produces more diverse outputs (higher Distinct-1/2) and removes near-duplicate profiles entirely.

These gains also extend to \textbf{survey-grounded alignment on real-student data}. On CEPS, cohort-level correlations increase for teacher attention ($\Delta\rho=+0.132$, $\Delta r=+0.139$) and help-seeking ($\Delta\rho=+0.536$, $\Delta r=+0.641$). On PISA~2022, \texttt{MATHEASE} improves by $\Delta r=+0.27$--$+0.29$ (baseline $r=0.45$--$0.63$), and \texttt{MATHEFF} improves by $\Delta r=+0.13$--$+0.14$ across regions. This indicates that HACHIMI's gains are not limited to better-formed personas in isolation, but translate into stronger external alignment under the same downstream evaluation pipeline.

\textit{Full baseline protocol and per-construct results are reported in Appendix~\ref{app:baseline}.} For reproducibility, we report the experimental settings in Appendix~\ref{app:env}.

\begin{table}[t]
  \centering
  \small
  \setlength{\tabcolsep}{4pt}
  \renewcommand{\arraystretch}{1.05}
  \caption{Protocol-matched intrinsic comparison.}
  \label{tab:baseline-mini}
  \begin{tabular}{lccc}
    \toprule
    \textbf{Metric} & \textbf{Baseline} & \textbf{HACHIMI} & \textbf{$\Delta$} \\
    \midrule
    Hard error rate ($\downarrow$) & 12.03\% & 0.00\% & $-12.03$ \\
    Warning rate ($\downarrow$)    & 25.33\% & 0.82\% & $-24.51$ \\
    Distinct-1 ($\uparrow$)        & 0.2328  & 0.3285 & $+0.0957$ \\
    Distinct-2 ($\uparrow$)        & 0.4589  & 0.7893 & $+0.3304$ \\
    Near-duplicate pairs ($\downarrow$) & 157 & 0 & $-157$ \\
    \bottomrule
  \end{tabular}
\end{table}

\section{Discussion}
% 讨论

% [第一段翻译]
% 这项工作表明，当被视为明确的目标而非提示（prompting）的副产品时，大规模、理论锚定且分布可控的学生画像生成是可行的。
% 在 TAD-PG 公式下，HACHIMI 多智能体框架结合了模块化生成、神经符号验证和分层多样性控制，以生成满足模式和配额约束且避免模式崩溃的画像集合。
% 在这个意义上，HACHIMI-1M 不仅仅是一个数据集，而是一个实例化的设计空间：它证明了教育理论、发展分类法和配额调度器可以直接嵌入到生成过程中。
This work shows that large-scale, theory-aligned and distribution-controllable student persona generation is feasible when treated as an explicit objective rather than a byproduct of prompting. Under the TAD-PG formulation, the HACHIMI multi-agent framework combines modular generation, neuro-symbolic validation and stratified diversity control to produce persona collections that satisfy schema and quota constraints while avoiding mode collapse. In this sense, HACHIMI-1M is \textbf{not just a dataset but an instantiated design space}: it demonstrates that educational theories, developmental taxonomies and quota schedulers can be integrated directly into the generation pipeline

% [第二段翻译]
% 我们的实证分析揭示了一种结构化的群体级保真度模式。
% 在 CEPS 和 PISA 上，基于画像的智能体最忠实地复现了数学相关的效能感、参与度、抱负、教师关注和好奇心/成长构念方面的群体差异，而与抑郁症状、父母严厉程度、学校归属感和工作量平衡相关的构念则难以匹配。
% 跨区域的 PISA 结果进一步显示了一种分层特征：数学和好奇心方面的一致性稳健，课堂氛围和归属感方面的一致性处于中等水平，而幸福感和工作量指标则表现为微弱甚至负相关。
% 这种\textit{保真度梯度}表明，学生体验中外显的、面向学校的方面更容易从静态画像中推断出来，而潜在的心理健康和家庭动态模式则仍然难以确定。
Our empirical analyses reveal a structured pattern of group-level fidelity. On CEPS and PISA, persona-based agents most faithfully reproduce cohort differences in math-related efficacy, engagement, aspirations, teacher attention and curiosity/growth constructs, while constructs tied to depressive symptoms, parental strictness, school bonding and workload balance are much harder to match. The cross-regional PISA results further show a tiered profile: robust alignment for math and curiosity, intermediate consistency for classroom climate and belonging, and weak or even negative correlations for well-being and workload indicators. This \textbf{\textit{fidelity gradient}} suggests that \textit{externally visible, school-facing aspects of student experience} are \textbf{more easily inferred} from static personas, whereas \textit{latent mental health and family dynamics patterns} \textbf{remain underdetermined}.

% [第三段翻译]
% 这些发现为教育 AI 带来了机遇也带来了警示。
% HACHIMI-1M 提供了一个可复用的测试平台，用于对教育大语言模型进行基准测试，审计性别-成绩-风险阶层的公平性，并在真实数据稀缺或敏感时对干预政策进行原型设计，但它不应被视为在最微妙领域中替代真实学生证据的工具。
% 一个有前景的方向是将合成画像视为一种校准后的先验：将 HACHIMI 风格的理论锚定群体与小型、精心设计的调查样本相结合，对画像的时间变化进行建模，并将其与下游的课堂模拟、教师培训场景和安全评估流程更紧密地整合。
These findings carry both opportunities and cautions for educational AI. HACHIMI-1M provides a standardized synthetic student population and a reusable testbed for cohort-matched benchmarking of educational LLMs and for social-science simulation, especially when real-student data are scarce or access-restricted; however, it should not be treated as a substitute for real-student evidence on subtle constructs such as well-being and family dynamics.

\section{Conclusion}

Theory-aligned, distribution-controllable student personas are hard to build with ad hoc \textit{act-as-a-student} prompting.
We presented HACHIMI, casting persona construction as TAD-PG with quota scheduling, multi-agent generation, neuro-symbolic validation, and diversity control.
Across intrinsic tests and CEPS/PISA~2022, HACHIMI achieves near-perfect schema validity, accurate quotas, and stronger cohort-level alignment, strongest on school-facing math and curiosity/growth and weaker on latent well-being and family dynamics.
Overall, HACHIMI positions student personas as structured, evaluable infrastructure for educational LLMs, enabling reliable agents and principled evaluation against large-scale survey data.

\section*{Limitations}

While our findings are encouraging, several limitations warrant caution. First, our external validation relies on two large-scale survey datasets---CEPS Grade~8 and PISA~2022---which, although widely used, cover specific age ranges, curricula and sociocultural contexts. HACHIMI personas and agents may behave differently when instantiated in other educational systems (e.g., vocational tracks, early childhood, or adult learners), and we do not claim cross-context generality beyond the settings examined in this work.

Second, our behavioural analysis focuses on group-level statistics over relatively short shadow surveys. We do not evaluate fine-grained, turn-by-turn classroom interactions, long-term learning trajectories, or micro-level causal effects of pedagogical interventions. A persona that reproduces cohort means on selected constructs may still exhibit unrealistic behaviour in extended dialogues, and our framework currently treats personas as static states rather than dynamically evolving learners. Third, we instantiate agents with a single family of LLM backbones and a specific prompting setup; different base models, decoding strategies, or role-playing prompts could lead to different degrees of alignment.

Finally, our approach remains constrained by the limitations of both LLMs and the underlying surveys. HACHIMI may inherit or amplify biases present in CEPS and PISA, and our theory-anchored schema necessarily simplifies complex constructs such as mental health, values, and family relations into a finite set of labels and narratives. We do not address fairness, privacy, or potential misuse in depth, and any real-world deployment of HACHIMI personas or agents should be accompanied by careful human oversight, ethical review, and, where possible, validation against up-to-date, context-specific student data.

\section*{Ethics Statement}
This paper introduces HACHIMI, a framework for generating \emph{synthetic} Grades 1--12 student personas and using persona-instantiated agents for cohort-level survey analyses.
We emphasize that all personas in HACHIMI-1M are \emph{fictional} and are not intended to represent, impersonate, or be linked to any real individual.

\paragraph{Use of real-student data.}
We use CEPS and PISA~2022 solely as \emph{evaluation references} and conduct analyses at the \emph{aggregated cohort level}.
We follow the corresponding data-use terms and do not redistribute any restricted microdata.
When constructing shadow surveys, we only report derived statistics and provide procedures such that replication relies on users' legitimate access to the original instruments and datasets.

\paragraph{Sensitive attributes and interpretation.}
Some labels (e.g., psychological-risk indicators derived from survey scales) are used only for \emph{matched cohort analysis} and should not be interpreted as clinical diagnoses or used for individual-level screening.
Similarly, M/F follows survey coding and is used only for cohort matching; it is not intended to imply that gender is binary.
More broadly, the observed ``fidelity gradient'' indicates that certain latent or private constructs (e.g., well-being or family dynamics) are harder to infer reliably from static personas; outputs should therefore not be used for high-stakes decisions.

\paragraph{Bias, representativeness, and potential misuse.}
Synthetic personas may reflect biases in model priors and in the choice of schema, prompts, and constraints, and they may underrepresent minority or non-normative experiences.
To mitigate these risks, we (i) make schema and control knobs explicit, (ii) report stratified evaluations across cohorts and regions, and (iii) recommend that future users validate conclusions against real-student evidence when feasible.
We caution against using HACHIMI-1M for surveillance, automated profiling, or decision-making about real students; it is intended as a standardized testbed for benchmarking and simulation research.

Before release, we run automated screening to remove instances that contain obvious PII patterns (e.g., phone numbers, emails, URLs) and apply toxicity/offensiveness filters; flagged samples are discarded or regenerated.

\begin{comment}
本文提出 HACHIMI 用于生成\emph{合成}K--12 学生画像，并将画像实例化为学生代理进行群体层面评测。我们强调 HACHIMI-1M 中所有画像均为虚构，不用于指代或关联任何真实个体。
真实学生数据（CEPS、PISA~2022）仅用于评测参照，分析在群体汇总层面完成，并遵循数据使用协议，不再分发受限微观数据。影子问卷只用于得到汇总统计，复现依赖研究者对原始数据/量表的合法获取。
心理风险等标签仅用于群体匹配，不构成临床诊断；M/F 仅沿用问卷编码用于群体匹配，不意味着性别二元。鉴于部分潜在构念难以从静态画像可靠推断，严禁将输出用于对真实学生的高风险决策或筛查。我们也提示合成画像可能带来偏差与代表性不足，建议使用者在可行时用真实数据进行校准与验证，避免用于监控、自动化画像或实际决策。
\end{comment}

\begin{comment}
    尽管我们的研究结果令人鼓舞，但仍存在若干局限性需谨慎对待。首先，我们的外部验证依赖于两个大规模调查数据集——CEPS八年级数据和PISA-2022数据——尽管这些数据被广泛使用，但仅覆盖特定年龄段、课程体系及社会文化背景。当HACHIMI角色与智能体应用于其他教育体系（如职业教育、学前教育或成人教育）时，其行为模式可能存在差异，我们亦不主张其结论具有超越本研究场景的跨情境普适性。

其次，行为分析聚焦于相对短期的影子调查中的群体统计数据，未评估细粒度的逐轮课堂互动、长期学习轨迹或教学干预的微观因果效应。在特定构念上复现群体均值的角色模型，在扩展对话中仍可能表现出不切实际的行为，且当前框架将角色视为静态状态而非动态演进的学习者。第三，当前代理实例仅基于单一大型语言模型（LLM）骨干架构及特定提示设置构建；不同基础模型、解码策略或角色扮演提示可能导致不同程度的匹配偏差。

最后，本方法仍受限于LLM与基础调查的双重局限。HACHIMI可能继承或放大CEPS和PISA存在的偏见，而理论锚定的框架必然将心理健康、价值观、家庭关系等复杂构念简化为有限标签与叙事集。我们未深入探讨公平性、隐私保护及潜在滥用问题，任何现实场景中部署HACHIMI人格或智能体时，均需配备严谨的人工监督机制、伦理审查流程，并在可行情况下结合最新情境化学生数据进行验证。
\end{comment}
\section*{Acknowledgments}
This work is supported by the Shanghai Municipal Education Commission's Special Fund for Educational Large Models (93600-515100-25001).
\bibliography{custom}

\clearpage
\onecolumn
\appendix
\section{Implementation and Runtime Environment}
\label{app:env}

For reproducibility, we summarise the main implementation and runtime settings below:

\begin{itemize}
    \item \textbf{Programming language.} All data processing, persona scheduling, and agent-based evaluations (including CEPS and PISA shadow surveys) are implemented in Python~3.10.
    \item \textbf{Operating systems.} Experiments were conducted on standard workstation-class machines running recent versions of Linux and Windows.
    \item \textbf{Hardware.} All pipelines run on commodity CPUs without requiring specialised local GPUs; language model inference is performed via remote APIs rather than local deployment.
    \item \textbf{Python stack.} We rely on a standard scientific Python stack, including NumPy, pandas, SciPy, scikit-learn, and tqdm, plus common utilities for JSON/CSV I/O and logging.
    \item \textbf{LLM access.} Student agents are instantiated and queried through an OpenAI-compatible Python client connecting to the \texttt{deepseek-chat} model over HTTPS, with modest request-level concurrency to respect provider rate limits.
\end{itemize}

\section{High-level Education Capacities and Persona Schema}
\label{app:persona-schema}

\subsection{High-level Education Capacities: Behavioral Definitions}

We operationalize “high-level education” along four capacities that guide both persona design and model evaluation:

\paragraph{Personalization.}
Through ongoing interaction rather than explicit labels alone, the model should implicitly construct and refine an internal student persona, dynamically diagnose the learner's knowledge state (including academic proficiency, value orientation, creativity and mental health), adapt to the student's cognitive style and pace, and provide targeted support that not only closes knowledge gaps but also promotes positive development in values, creativity and psychological well-being.

\paragraph{Value formation.}
When facing moral dilemmas, harmful content or value-laden questions, the model is expected to provide heuristic guidance, positive clarification and norm-consistent redirection, while upholding safety boundaries and avoiding value distortion.

\paragraph{Creativity support.}
The model should elicit and scaffold divergent thinking, pose Socratic questions, and promote distant associations between concepts, thereby improving the student's ability to generate, analyze and refine creative solutions.

\paragraph{Mental health support.}
The model should empathetically recognize and respond to students' emotions, provide support grounded in positive psychology, and respect safety boundaries and crisis-prevention principles, without stepping into non-professional diagnosis.

These behavioral definitions provide the target behaviors for teacher models and determine which aspects of each persona must be observable and editable for the TAD-PG task.

\subsection{Mapping Capacities to Persona Components}

The four capacities are not represented by a single field, but are distributed across five complementary persona components used in the TAD-PG task:

\begin{itemize}
    \item \textbf{Demographic \& developmental status} and \textbf{academic profile} primarily support \emph{personalization}, by encoding grade, developmental stage and subject-specific strengths and weaknesses that the model can use to tailor instruction.
    \item \textbf{Personality \& value orientation} provide the main observable substrate for \emph{value formation}, while also informing how personalized guidance and mental-health responses should be framed.
    \item \textbf{Social relations \& creativity} directly expose the student's social context and creative problem-solving patterns, enabling the model to exercise \emph{creativity support} and to reason about peer/teacher/family dynamics.
    \item \textbf{Mental health \& well-being} offers a coarse but structured description of the student's emotional functioning and support systems, forming the primary basis for \emph{mental health support} and for tracking changes in well-being.
\end{itemize}

In practical downstream applications, these component-level descriptors can serve as observable state variables for tracking whether teacher models move students toward desirable educational directions. In this paper, however, we do not evaluate such state transitions; instead, we use these components as static persona states for agent instantiation and group-level consistency analysis.

\subsection{Persona Components in HACHIMI}

HACHIMI instantiates the theory-anchored schema through five concrete components. Each component corresponds to one agent in the multi-agent architecture and is implemented as a structured set of fields.

\subsubsection{Demographic \& Developmental Status}

This component is produced by the \textsc{Scholar} agent and contains:

\begin{itemize}
    \item \textbf{Name}: a full name.
    \item \textbf{Age} an integer between 6 and 18, loosely consistent with grade level.
    \item \textbf{Gender}: categorical.
    \item \textbf{Grade}: from primary grade one to senior high three.
    \item \textbf{Developmental stages}: a structured object with three keys---Piaget cognitive stage, Erikson psychosocial stage, and Kohlberg moral development stage, aligned with classical developmental theories~\cite{kohlberg1977moral}.
    \item \textbf{Agent identifier}: a phonetic identifier following a constrained schema (1--2 syllables for surname, 1--3 for given name; each syllable is lowercase \texttt{pinyin} plus tone number; surname and given name separated by an underscore), ensuring uniqueness and downstream usability.
\end{itemize}

This component anchors the persona in a plausible age--grade band and provides the developmental frame needed to interpret value, creativity and mental-health descriptions.

\subsubsection{Academic Profile}

The \textsc{Academic} agent generates the academic profile:

\begin{itemize}
    \item \textbf{Strong subjects}: a non-empty set of subjects drawn from clusters (STEM, humanities/social sciences, arts/PE, languages/biology), influenced by the sampling constraint on preferred subject cluster.
    \item \textbf{Weak subjects}: a non-empty set disjoint from strong subjects.
    \item \textbf{Achievement level}: a four-level categorical scale with fixed textual anchors:
    \begin{enumerate}
        \item “High: top 10\% in school”,
        \item “Medium: top 10\%--30\%”,
        \item “Low: top 30\%--50\%”,
        \item “Poor: bottom 50\%”.
    \end{enumerate}
\end{itemize}

Achievement levels are used as hard anchors in quota scheduling and as conditioning signals for other agents, thus linking macro-level distribution control with micro-level content.

\subsubsection{Personality \& Value Orientation}

The \textsc{Values} agent is responsible for:

\begin{itemize}
    \item \textbf{Personality}: a short narrative description of personality traits (e.g., introversion/extraversion, conscientiousness, emotional stability), written in natural language.
    \item \textbf{Values}: a single-paragraph description that must explicitly cover seven dimensions, each with an interpretable level word, inspired by contemporary value-education frameworks:
    \begin{enumerate}
        \item moral cultivation,
        \item physical and mental health,
        \item rule-of-law awareness,
        \item social responsibility,
        \item political identity,
        \item cultural literacy,
        \item family orientation.
    \end{enumerate}
\end{itemize}

Heuristics in the prompts and post-hoc filters prevent uniformly optimistic profiles by requiring a minimum count of “medium / low” level indicators for personas anchored to lower achievement tiers, thus tying value descriptions to the overall distributional design.

\subsubsection{Social Relations \& Creativity}

The \textsc{Social-Creative} agent produces:

\begin{itemize}
    \item \textbf{Social relations}: a single paragraph (approximately 160--260 Chinese characters) describing peer, teacher and family interactions in a background--event--impact structure.
    \item \textbf{Creativity}: a single paragraph that combines eight dimension-specific judgments with a short “radar-style” summary, following creativity and CPS research~\cite{runco2004creativity,beghetto2014classroom}:
    \begin{enumerate}
        \item fluency,
        \item originality,
        \item flexibility,
        \item feasibility,
        \item problem finding,
        \item problem analysis,
        \item solution generation,
        \item solution refinement.
    \end{enumerate}
\end{itemize}

Each dimension must be associated with a level word and a brief justification. Internal consistency constraints (e.g., low feasibility cannot co-occur with very high solution generation) are enforced by the validator and light filters. As with values, creativity levels adapt to the academic-level anchor to suppress overly optimistic profiles.

\subsubsection{Mental Health \& Well-being}

The \textsc{Health} agent is in charge of the mental-health component:

\begin{itemize}
    \item \textbf{Mental health}: a single paragraph that integrates:
    \begin{enumerate}
        \item an overall summary of psychological functioning;
        \item at least two salient personality or temperament features;
        \item coarse indicators of overall mental state and subjective well-being (e.g., “overall mental status” and “happiness index”);
        \item risk descriptions for depression and anxiety (non-diagnostic, using educational language);
        \item background stressors and protective factors (e.g., family, peers, school);
        \item current supports and coping strategies.
    \end{enumerate}
\end{itemize}

Prompts explicitly require non-diagnostic language and coherence with the value dimension “physical and mental health”, while filters prevent unrealistic combinations (e.g., very low achievement anchors with uniformly low risk and very high happiness).

\subsection{Sampling Constraints and Distribution Control}

For each persona, \texttt{\_sampling-constraint} field encodes:

\begin{itemize}
    \item target grade,
    \item gender,
    \item preferred subject cluster,
    \item target achievement level.
\end{itemize}

These constraints are generated by the quota scheduler to match pre-specified cohort-level distributions and are passed to all agents through the shared whiteboard. Agents must respect these constraints when producing their fields, and the validator checks alignment. As a result, the theory-anchored persona schema functions both as a conceptual bridge to high-level education capacities and as a concrete interface for distribution-controllable persona generation in the TAD-PG task.

\section{Overview of the HACHIMI Pipeline}
\label{app:hachimi-overview}

\textbf{Where to find the end-to-end overview.}
Figure~\ref{fig:hachimi-overview} (main text, Section~\ref{sec:architecture}) gives an end-to-end overview of the HACHIMI pipeline, from distribution-aware sampling to the final HACHIMI-1M corpus for Theory-Aligned and Distribution-Controllable Persona Generation (TAD-PG).
This appendix complements the main figure with a procedural summary (Algorithm~\ref{alg:hachimi-pipeline}) and stage-wise implementation details.

% (Optional but recommended) put the algorithm block here to replace the removed figure.
\begin{algorithm}[t]
\caption{HACHIMI pipeline.}
\label{alg:hachimi-pipeline}
\begin{algorithmic}[1]
\Require Target distributions over (grade, gender, academic level); persona schema $\mathcal{S}$; hard constraints $\mathcal{C}$; revision budget $R$
\Ensure Validated persona corpus $\mathcal{D}$
\State $\mathcal{D} \gets \emptyset$
\State $\mathcal{Q} \gets \textsc{QuotaSchedule}(\text{targets})$
\ForAll{slot $s \sim \textsc{StratifiedSample}(\mathcal{Q})$}
  \State $W \gets \textsc{InitWhiteboard}(s)$
  \State $p \gets \textsc{MultiAgentGenerate}(W, \mathcal{S})$
  \For{$t=1$ to $R$}
    \State $(ok, err) \gets \textsc{SymbolicCritic}(p, \mathcal{C})$
    \If{$ok$} \State \textbf{break} \EndIf
    \State $p \gets \textsc{Revise}(p, err, W)$
  \EndFor
  \If{$ok$} \State $\mathcal{D} \gets \mathcal{D} \cup \{p\}$ \EndIf
\EndFor
\State $\mathcal{D} \gets \textsc{DiversityControl}(\mathcal{D})$
\State \Return $\mathcal{D}$
\end{algorithmic}
\end{algorithm}

The pipeline consists of five tightly coupled stages:

\paragraph{1. Quota scheduling \& stratified sampling.}
Given target distributions over \emph{grade}, \emph{gender}, and \emph{academic level}, the scheduler allocates explicit quotas for each stratum and draws stratified samples of abstract ``slots''. Each slot encodes the macro-level variables required by the TAD-PG task (e.g., Grade~8, female, low-achievement, high-risk) and serves as a conditioning anchor for all subsequent agents. This stage enforces population-level control and guarantees coverage of under-represented groups such as struggling learners.

\paragraph{2. Multi-agent cooperative persona generation.}
For each scheduled slot, a society of specialized agents jointly constructs a holistic student persona on a shared whiteboard. Different agents are responsible for the four major components of the persona schema: \emph{academic profile}, \emph{personality \& values}, \emph{social relations \& creativity}, and \emph{mental health \& well-being}. The shared whiteboard exposes the partial state (e.g., subject strengths, achievement tier, family background) so that later agents can condition on earlier decisions, preventing cross-component contradictions and mitigating intra-profile inconsistency in long-form generation.

\paragraph{3. Neuro-symbolic validation.}
The draft persona is then passed to a rule-based \emph{Symbolic Critic}, which implements the neuro-symbolic constraints defined in the main text. The critic checks hard constraints derived from educational psychology and developmental theories (e.g., consistency between age and developmental stage, coherence between academic tier and self-efficacy, admissible combinations of risk factors). Instead of treating these checks as soft preferences, the validator encodes them as logical predicates over categorical labels and textual anchors extracted from the narrative fields.

\paragraph{4. Iterative revision with structured error feedback.}
Whenever a violation is detected, the Symbolic Critic emits structured error messages that point to the offending components and the violated rules. These feedback signals are fed back to the relevant generators via the shared whiteboard, prompting targeted revision rather than unconditional regeneration. The loop continues until all hard constraints are satisfied or a small revision budget is exhausted, yielding a persona that satisfies the TAD-PG schema while preserving as much narrative richness as possible.

\paragraph{5. Diversity control \& finalization.}
In the final stage, the system applies semantic diversity control over the pool of validated personas. A semantic deduplication mechanism (e.g., SimHash-based or other locality-sensitive hashing) flags near-duplicate narratives within the same stratum, and redundant entries are pruned or rewritten. Diversity indices at both token and construct levels are monitored to avoid mode collapse towards a few generic templates. The remaining personas are serialized into the hybrid semi-structured format described in the main paper, and aggregated into the HACHIMI-1M corpus as a theory-grounded, quota-controlled resource for downstream educational AI research.

% In preamble (if not already included):
% \usepackage{tabularx}
% \usepackage{array}

\clearpage

\section{Agent Prompt Templates}
\label{app:agent-prompts}

This appendix provides the complete prompt templates for all agents in the HACHIMI multi-agent system. 
Each prompt box preserves the effective instruction template used in the implementation, while omitting low-level runtime wrappers such as API calls and logging.

\subsection{Universal Agent Preamble (Shared by All Content Agents)}

\begin{fullpromptlisting}{Universal Agent Preamble}
You are a "student profile" production member collaborating with other agents. We use a "public whiteboard" to share drafts and discussions.

Rules (must follow):
- All output must be valid JSON objects, and only contain keys you are responsible for.
- Do not quote template phrases; use natural English; avoid empty cliches; avoid contradicting whiteboard draft.
- If asked to revise, only change keys you're responsible for; leave nothing empty; ensure logical consistency with other fields.
- Names should be in English; numbers and percentiles use English context (e.g., "top 10%").
- Do not output any extra explanatory text. Only output JSON.
- If "_Sampling Constraint" exists in the whiteboard, strictly follow "Grade", "Gender", "Strong Subject Preference", "Target Academic Level" and other requirements; if conflicts occur, follow sampling constraints and maintain overall consistency.
- Important: Every student has both strengths and weaknesses. Prohibited from writing all dimensions as "very good" or "clear advantage". When target academic level is "Mid/Low/Poor", your responsible fields must explicitly write some "Mid/Average/Weaker/Relatively Low/Low" descriptions to make the profile resemble a real person.

[Universal Output Hard Constraints] (All Agents must follow):
- You can only output one JSON object, not an array, and cannot add any explanatory text outside JSON.
- Absolutely prohibited from using ```json, ```, or other Markdown code block wrappers around output.
- The outermost layer of JSON can only contain keys this Agent is responsible for, must not add "id", "Student Info", "profile", or any other keys.
- Must not nest another meaningless wrapper object (e.g., structure like {"Student Info": {...}} is prohibited).
\end{fullpromptlisting}
\clearpage
\subsection{Enrollment \& Development Agent}

\noindent
\textbf{Responsible Fields:} Name, Age, Gender, Grade, Developmental Stage, Agent Name

\noindent
\textbf{Task Mode:} propose or revise

\begin{fullpromptlisting}{Enrollment \& Development Agent Prompt}
[AGENT_PREAMBLE]
[Sampling Constraint hint if present: Grade, Gender, Target Academic Level]

You are responsible for: ["Name", "Age", "Gender", "Grade", "Developmental Stage", "Agent Name"]
Task mode: {propose|revise}
Diversity seed: {seed}

Generation and Constraints (required):
- Age 6-18; Age must be an Arabic numeral integer; Grade must match Age (allow +/-1 year for skip/retention but must be consistent with other sections);
- Developmental Stage object must contain three keys: Piaget Cognitive Development Stage, Erikson Psychosocial Development Stage, Kohlberg Moral Development Stage;
- Agent Name format (multi-syllable support): Surname 1-2 syllables, Given Name 1-3 syllables, but 2-syllable surname cannot pair with 3-syllable given name; each syllable is "lowercase pinyin+tone number(1-5)"; underscore between surname and given name, e.g., li3_gan4, jiang3_jie4shi2, ou1yang2_chen2fei1, yi4_yang2qian1xi3.

[Output Format Hard Constraints]:
- You can only output one JSON object, and top-level can only contain the following 6 keys:
  1) "Name" (string, English name)
  2) "Age" (integer, e.g., 12)
  3) "Gender" (string, can only be "Male" or "Female")
  4) "Grade" (string, e.g., "Grade 6", "Grade 7", "Grade 10")
  5) "Developmental Stage" (object, containing three subkeys)
  6) "Agent Name" (string, conforming to given regex)
- "Developmental Stage" must be an object and can only contain these three subkeys:
  - "Piaget Cognitive Development Stage"
  - "Erikson Psychosocial Development Stage"
  - "Kohlberg Moral Development Stage"
- Absolutely prohibited from adding "id", "Student Info", or other keys; no additional wrapper layer.
- Do not use ```json or ``` to wrap output.

[Qualified Example (strictly mimic structure, only change content)]:
[$case of JSON structure with Name, Age, Gender, Grade, Developmental Stage object with Piaget/Erikson/Kohlberg, Agent Name]

Please follow the above JSON structure, directly output the current student's JSON object, no explanatory text.
\end{fullpromptlisting}
\clearpage
\subsection{Academic Profile Agent}

\noindent
\textbf{Responsible Fields:} Strong Subjects, Weak Subjects, Academic Level

\noindent
\textbf{Task Mode:} propose or revise

\begin{fullpromptlisting}{Academic Profile Agent Prompt}
[AGENT_PREAMBLE]

You are responsible for: ["Strong Subjects", "Weak Subjects", "Academic Level"]
Task mode: {propose|revise}
Diversity seed: {seed}

Field meanings and content requirements (required):
- "Strong Subjects": Non-empty array, each element is a subject name (e.g., "Chinese", "Mathematics"); no duplicates within array.
- "Weak Subjects": Non-empty array, each element is a subject name; set disjoint with "Strong Subjects".
- "Academic Level": Must strictly equal one of the following four strings:
  1) "High: Top 10% school ranking"
  2) "Mid: Top 10%-30% school ranking"
  3) "Low: Top 30%-50% school ranking"
  4) "Poor: Bottom 50% school ranking"
- [Subject preference hint if present]
- [Target Academic Level constraint if present: "This sample's 'Academic Level' must strictly equal: {level}"]

[Output Format Hard Constraints]:
- You can only output one JSON object, and top-level can only contain the following 3 keys:
  1) "Strong Subjects"
  2) "Weak Subjects"
  3) "Academic Level"
- These 3 keys must all appear, cannot be missing, cannot add any other keys.
- Absolutely prohibited from using "Student Info", "id", or other extra wrapper objects.
- Do not use ```json or ``` to wrap output.

[Qualified Example]:
[$case of JSON with Strong Subjects array, Weak Subjects array, and Academic Level string]

Please follow the above JSON structure, output only the JSON object itself.
\end{fullpromptlisting}

\subsection{Personality \& Values Agent}

\noindent
\textbf{Responsible Fields:} Personality, Values

\noindent
\textbf{Task Mode:} propose or revise

\noindent
\textbf{Adaptive Constraints (triggered when Target Academic Level is Mid/Low/Poor):}

\begin{fullpromptlisting}{Personality \& Values Agent: Adaptive Constraints}
- [7-Dimension Mandatory Distribution (strongly bound to system filters, must satisfy)]
  - You must write both strengths and weaknesses across seven dimensions, cannot all be 'Relatively High/Very High'.
  - Only allowed to use these level words: High / Relatively High / Upper-Mid / Mid / Relatively Low / Low.
  - When target academic level is 'Mid': Among seven dimensions, at least 1 dimension described with 'Mid/Relatively Low/Low'.
  - When target is 'Low': Among seven dimensions, at least 2 dimensions with 'Mid/Relatively Low/Low'.
  - When target is 'Poor': Among seven dimensions, at least 3 dimensions with 'Mid/Relatively Low/Low'.
  - For each 'Relatively High/Very High/Clear Advantage' description, pair with at least one 'Mid/Relatively Low/Low/Needs Improvement' description.
  - For each dimension written as 'Mid/Relatively Low/Low', provide brief justification.
\end{fullpromptlisting}

\subsection{Personality \& Values Agent}

\noindent
\textbf{Responsible Fields:} Personality, Values

\noindent
\textbf{Task Mode:} propose or revise

\begin{fullpromptlisting}{Personality \& Values Agent Prompt}
[AGENT_PREAMBLE]

You are responsible for: ["Personality", "Values"]
Task mode: {propose|revise}
Diversity seed: {seed}

Output format requirements (required):
- "Personality": One or several sentences of natural language, summarizing core personality traits (e.g., introverted/extroverted, responsibility, openness), maintaining education-scenario friendliness.
- "Values": Single-paragraph continuous natural language, no paragraphs, no lists or numbering.
  - Need to explicitly cover seven dimensions: Moral Character, Physical-Mental Health, Rule of Law, Social Responsibility, Political Identity, Cultural Literacy, Family Values;
  - Each dimension should have a recognizable level word (e.g., High/Relatively High/Upper-Mid/Mid/Relatively Low/Low), with brief justification.

[Conditional Adaptive Constraints, appended only when Target Academic Level is Mid/Low/Poor]:
- [7-Dimension Mandatory Distribution (strongly bound to system filters, must satisfy)]
  - You must write both strengths and weaknesses across seven dimensions, cannot all be 'Relatively High/Very High'.
  - Only allowed to use these level words: High / Relatively High / Upper-Mid / Mid / Relatively Low / Low.
  - When target academic level is 'Mid': Among seven dimensions, at least 1 dimension described with 'Mid/Relatively Low/Low'.
  - When target is 'Low': Among seven dimensions, at least 2 dimensions with 'Mid/Relatively Low/Low'.
  - When target is 'Poor': Among seven dimensions, at least 3 dimensions with 'Mid/Relatively Low/Low'.
  - For each 'Relatively High/Very High/Clear Advantage' description, pair with at least one 'Mid/Relatively Low/Low/Needs Improvement' description.
  - For each dimension written as 'Mid/Relatively Low/Low', provide brief justification.

[Output Format Hard Constraints]:
- You can only output one JSON object, and top-level can only contain the following 2 keys:
  1) "Personality"
  2) "Values"
- Not allowed to have "Student Info", "id", "Evaluation", or any other top-level keys.
- "Values" must be single-paragraph text, no blank lines, no list symbols (e.g., "-", "1.", etc.) or Markdown in the middle.
- Do not use ```json or ``` to wrap output.

[Qualified Example]:
[$case of JSON with Personality description and Values single-paragraph covering 7 dimensions with level words]

Please follow the above JSON structure, output only JSON object.
\end{fullpromptlisting}

\clearpage
\subsection{Social \& Creativity Agent}

\noindent
\textbf{Responsible Fields:} Social Relationships, Creativity

\noindent
\textbf{Task Mode:} propose or revise

\begin{fullpromptlisting}{Social \& Creativity Agent Prompt}
[AGENT_PREAMBLE]

You are responsible for: ["Social Relationships", "Creativity"]
Task mode: {propose|revise}
Diversity seed: {seed}

Field and format requirements (required):
- "Social Relationships": Single paragraph (approx. 160-260 chars), narrate in "Background -> Key Event -> Impact" order, no line breaks or lists.
- "Creativity": Single-paragraph natural language, must include:
  - Evaluation of eight dimensions one by one: Fluency, Novelty, Flexibility, Feasibility, Problem Discovery, Problem Analysis, Proposing Solutions, Improving Solutions, each dimension with clear level word (High/Relatively High/Upper-Mid/Mid/Relatively Low/Low) and brief justification;
  - Ending with an overall "Radar Summary" comprehensive summary.
- If "Feasibility" relatively low or low, then "Proposing Solutions" no higher than mid-level, avoid self-contradiction.

[Conditional Adaptive Constraints, appended only when Target Academic Level is Mid/Low/Poor]:
- [8-Dimension Level Mandatory Distribution (strongly bound to filters)]
  - You need to first mentally assign a level word to each of 8 dimensions, only choose from 'High/Relatively High/Upper-Mid/Mid/Relatively Low/Low', then weave these 8 dimensions into a paragraph description.
  - Prohibited from writing all 8 dimensions as 'High/Relatively High', must have clear ups and downs.
  - When target academic level is 'Mid': Among 8 dimensions, at least 2 dimensions with 'Mid/Relatively Low/Low'.
  - When target is 'Low': Among 8 dimensions, at least 3 dimensions with 'Mid/Relatively Low/Low'.
  - When target is 'Poor': Among 8 dimensions, at least 4 dimensions with 'Mid/Relatively Low/Low'.
  - If 'Feasibility' written as 'Relatively Low/Low', then 'Proposing Solutions' level not allowed above 'Mid'.
  - The ending 'Radar Summary' must clearly point out which dimensions are strengths, which are obvious shortcomings. Must contain 'radar' or 'summary' word.

[Output Format Hard Constraints]:
- You can only output one JSON object, and top-level can only contain the following 2 keys:
  1) "Social Relationships"
  2) "Creativity"
- Not allowed to have "Student Info", "id", "Description", or any other keys.
- Both "Social Relationships" and "Creativity" must be single-paragraph text, cannot contain list symbols, numbering, Markdown, etc.
- Do not use ```json or ``` to wrap output.

[Qualified Example]:
[$case of JSON with Social Relationships paragraph and Creativity paragraph covering 8 dimensions with Radar Summary]

Please follow the above JSON structure, output only JSON object.
\end{fullpromptlisting}

\clearpage
\subsection{Mental Health Agent}

\noindent
\textbf{Responsible Fields:} Mental Health

\noindent
\textbf{Task Mode:} propose or revise

\begin{fullpromptlisting}{Mental Health Agent Prompt}
[AGENT_PREAMBLE]

You are responsible for: ["Mental Health"]
Task mode: {propose|revise}
Diversity seed: {seed}

Field and format requirements (required):
- "Mental Health": Single-paragraph natural language, no paragraphs or lists.
  Suggest naturally interspersing in the following order within paragraph:
  1) Overview of overall mental state;
  2) At least two personality traits related to psychological adaptation;
  3) Give clear level or degree descriptions for: Overall Mental State, Happiness Index, Depression Risk, Anxiety Risk;
  4) If no clear mental illness, include "Insufficient information or no significant symptoms" non-diagnostic description; if risks or tendencies exist, use "May have... tendency", "Mild... experience", "Recommend further assessment";
  5) Brief background story (e.g., academic pressure, interpersonal conflicts, family events);
  6) Current support and coping methods (family, teachers, peers, school resources).

[Conditional Adaptive Constraints, appended only when Target Academic Level is Low/Poor]:
- [Psychological Index Distribution (strongly bound to filters)]
  - Please explicitly give levels or degrees for four items in the text: Overall Mental State, Happiness Index, Depression Risk, Anxiety Risk.
  - When target academic level is 'Low/Poor':
    * Among Overall Mental State and Happiness Index, at least 1 item uses 'Mid/Average/Relatively Low/Below Average' mid-to-low descriptions, cannot both write 'Relatively High/Very High'.
    * Depression Risk and Anxiety Risk cannot both be written as 'Low/Very Low/Almost No Risk', at least 1 needs to show 'Mild/Some Possibility/Needs Attention' description.
  - Meanwhile maintain non-diagnostic tone for education scenarios: avoid directly using clinical diagnostic terms like 'Severe Depression/Bipolar/Need Hospitalization', use 'May have... tendency', 'Periodic Emotional Lows', 'Recommend Further Assessment' to describe.
  - When describing background stories and support systems, make readers feel: although some stress or distress exists currently, it can gradually improve through family, teachers, school resource cooperation.

[Output Format Hard Constraints]:
- You can only output one JSON object, and top-level can only contain 1 key:
  1) "Mental Health"
- Not allowed to have "Student Info", "id", "Evaluation", or other keys.
- "Mental Health" must be single-paragraph text, cannot contain blank lines, list symbols, Markdown code blocks.
- Do not use ```json or ``` to wrap output.

[Qualified Example]:
[$case of JSON with Mental Health single-paragraph containing overview, traits, 4 metrics, conditions, background, and support]

Please follow the above JSON structure, output only JSON object.
\end{fullpromptlisting}
\clearpage

\subsection{Validator Agent}

\noindent
\textbf{Purpose:} Comprehensive validation using all R1-R15 rules for fine-grained consistency checking

\noindent
\textbf{Input:} Current whiteboard state (draft profile)

\begin{fullpromptlisting}{Validator Prompt}
You are a 'Validator' agent. Please strictly review and provide structured revision tasks.
[AGENT_PREAMBLE]
You only output JSON, keys are issues and final_ready. Do not output extra text.

Reference Rules (required):
[Full R1-R15 rule block appended at runtime]

Output: issues: [{code, desc, owner, fields, hint}], final_ready: bool
\end{fullpromptlisting}

\noindent
\textit{Note.} In the implementation, the placeholder ``[Full R1--R15 rule block appended at runtime]'' is replaced by the complete executable rule set reported in Table~\ref{tab:validator-rules} in Appendix~\ref{app:validator-rules}.

\noindent
\textbf{Output Schema:}

\begin{fullpromptlisting}{Validator Output Schema}
{
  "issues": [
    {
      "code": "F1|F2|F3|F4",
      "desc": "Description of the issue",
      "owner": "Enrollment & Development|Academic Profile|Personality & Values|Social & Creativity|Mental Health",
      "fields": ["field_name"],
      "hint": "Suggestion for fix"
    }
  ],
  "final_ready": true|false
}
\end{fullpromptlisting}

\subsection{Field-to-Agent Responsibility Mapping}

\begin{table}[h]
\centering
\small
\renewcommand{\arraystretch}{1.08}
\setlength{\tabcolsep}{8pt}
\caption{Field-to-agent responsibility mapping in the HACHIMI multi-agent system.}
\label{tab:field-agent-mapping}
\begin{tabularx}{\textwidth}{>{\raggedright\arraybackslash}p{0.38\textwidth} >{\raggedright\arraybackslash}X}
\toprule
\textbf{Field} & \textbf{Responsible Agent} \\
\midrule
id & System \\
Name & Enrollment \& Development \\
Age & Enrollment \& Development \\
Gender & Enrollment \& Development \\
Grade & Enrollment \& Development \\
Developmental Stage & Enrollment \& Development \\
Agent Name & Enrollment \& Development \\
Strong Subjects & Academic Profile \\
Weak Subjects & Academic Profile \\
Academic Level & Academic Profile \\
Personality & Personality \& Values \\
Values & Personality \& Values \\
Social Relationships & Social \& Creativity \\
Creativity & Social \& Creativity \\
Mental Health & Mental Health \\
\bottomrule
\end{tabularx}
\end{table}

\clearpage

% If you still have later content that must return to two-column mode, uncomment:
% \clearpage

\section{Executable Validator Rules}
\label{app:validator-rules}

We implement a two-stage validator. A fast validator performs low-cost structural screening, while a deep validator applies the full executable rule set (R1--R15) below during generation-time validation and revision, as shown in Table \ref{tab:validator-rules}.

\begin{table*}[t]
\centering
\footnotesize
\renewcommand{\arraystretch}{1.05}
\setlength{\tabcolsep}{5pt}
\caption{Executable validator rules used in HACHIMI generation.}
\label{tab:validator-rules}
\begin{tabularx}{\textwidth}{>{\raggedright\arraybackslash}p{0.08\textwidth} >{\raggedright\arraybackslash}X}
\toprule
\textbf{ID} & \textbf{Rule} \\
\midrule
R1 & Age--grade norm: Grade 1--12 should approximately align with ages 6--18, allowing at most $\pm 1$ year deviation. \\
R2 & Developmental-stage plausibility: Piaget, Erikson, and Kohlberg stage labels must remain broadly compatible with the student's age band. \\
R3 & \texttt{Strong Subjects} and \texttt{Weak Subjects} must both be non-empty and mutually disjoint. \\
R4 & The eight creativity dimensions must show variation rather than identical levels; if feasibility is relatively low/low, proposing solutions cannot exceed a mid-level rating. \\
R5 & If the values field presents stable positive physical/mental health, the mental-health field must not describe severe functional impairment or heavy clinical pathology. \\
R6 & Agent-name regex: \texttt{\string^(?:[a-z]+[1-5])\{1,2\}\_(?:[a-z]+[1-5])\{1,3\}\string$}. \\
R7 & All required keys must be present and non-empty: \texttt{id}, \texttt{Name}, \texttt{Age}, \texttt{Strong Subjects}, \texttt{Weak Subjects}, \texttt{Grade}, \texttt{Personality}, \texttt{Social Relationships}, \texttt{Academic Level}, \texttt{Gender}, \texttt{Developmental Stage}, \texttt{Agent Name}, \texttt{Values}, \texttt{Creativity}, and \texttt{Mental Health}. \\
R8 & The values paragraph must cover seven dimensions (Moral Character, Physical-Mental Health, Rule of Law, Social Responsibility, Political Identity, Cultural Literacy, Family Values), each with a locatable level expression. \\
R9 & The creativity paragraph must include an overview, all eight dimensions (each with a level and brief rationale), and a radar-style summary. \\
R10 & The mental-health paragraph must include: overview; at least two personality/adaptation traits; overall mental state; happiness index; depression/anxiety risk; a non-diagnostic risk/tendency statement; background context; and support/coping. \\
R11 & Cross-field consistency: values, social, academic, and mental-health descriptions must be mutually supportive and not obviously contradictory. \\
R12 & Non-diagnostic language: avoid heavy clinical wording (e.g., severe depression, bipolar disorder, medication, hospitalization); allow mild/tendency/situational/manageable/recommend consultation wording. \\
R13 & Readability and anti-template constraint: reject mechanical repetition, laundry-list style text, or cases with missing level indicators/dimensions. \\
R14 & \texttt{Values}, \texttt{Creativity}, and \texttt{Mental Health} must each be a single continuous paragraph, without lists, bullets, numbering, or multi-paragraph breaks. \\
R15 & \texttt{Academic Level} must be one of four fixed labels only; otherwise the academic-profile owner must rewrite it using the strict four-choice format. \\
\bottomrule
\end{tabularx}
\end{table*}

% =========================================================

\section{Generation Efficiency and Key Speed-Related Settings}
\label{app:gen-eff}

All personas are generated with Qwen2.5-72B\cite{qwen2, qwen2.5}. We report compute in wall-clock time and GPU-hours.

\paragraph{Hardware and throughput.}
We run the end-to-end pipeline on 4$\times$8 NVIDIA H100 GPUs (32 H100s total). The average throughput is $\sim$10{,}000 personas/hour, and generating 1{,}000{,}000 personas takes $\sim$100 hours, corresponding to $\sim$3{,}200 GPU-hour in total.

\paragraph{Settings that affect runtime.}
We cap the Propose--Validate--Revise loop to at most 3 revision rounds. For semantic deduplication, we use SimHash with a near-duplicate threshold of Hamming distance $\leq 3$; candidates within this threshold are discarded and resampled.

\section{A Persona Sample (Original in Chinese; Author-translated to English)}
\label{app:persona-sample}

\paragraph{Note on language.}
The original persona record is written in Chinese for our internal generation and inspection workflow.
For readership, we provide an translation into English below, in Table~\ref{tab:persona-sample}.
All personas in our dataset are \emph{synthetic} and use pseudonymous identifiers.

\begin{table*}[t!]
\centering
\small
\renewcommand{\arraystretch}{1.08}
\setlength{\tabcolsep}{8pt}
\caption{One example persona (English translation from the original Chinese record).}
\label{tab:persona-sample}
\begin{tabularx}{\textwidth}{>{\raggedright\arraybackslash}p{3.2cm} >{\raggedright\arraybackslash}X}
\toprule
\textbf{Field} & \textbf{Content} \\
\midrule

Name (pseudonym) & Shihan Wang \\
Agent ID & \texttt{wang2\_shi1han2} \\
Age / Gender & 13 / Female \\
Grade & Grade 7 (Junior Year 1) \\
Academic standing & Low; bottom 50\% in school ranking \\
Strengths & Art \\
Weaknesses & Mathematics; English; Physics \\

\addlinespace[0.4em]
Developmental stage (theory-anchored) &
\textbf{Piaget:} Formal operational stage;
\textbf{Erikson:} Identity vs. role confusion;
\textbf{Kohlberg:} Conventional level \\

\addlinespace[0.4em]
Personality &
Shihan is relatively introverted. She enjoys sharing her art creations with familiar classmates and expresses emotions in a delicate, nuanced way, but she tends to be reserved in group conversations.
She can be patient when working on tasks; however, when facing academic challenges she easily falls into self-doubt.
Her initiative is moderate and she has not yet formed stable self-discipline habits. \\

\addlinespace[0.4em]
Values / civic literacy (inferred) &
Her moral conduct is moderate: she respects classmates and occasionally helps close friends.
Her physical and mental well-being is at a mid level, with occasional mood fluctuations triggered by academic stress or disappointing test results.
Her rule-of-law awareness is relatively low: she has a shallow understanding of school rules and her sense of punctuality is not strong.
Her sense of social responsibility is relatively low: she participates less in clubs and collective activities and focuses more on personal interests.
Her political identity/engagement is relatively low: she rarely follows current affairs beyond the classroom.
Her cultural literacy is moderate: she shows interest in traditional Chinese painting in art class but has limited exposure to other art forms.
Her family orientation is high: when encountering learning difficulties, she mainly relies on her parents; family support plays an important role in her emotional life.
(These judgments are inferred from her art-class performance, academic feedback, social participation, classroom behavior, and club participation.) \\

\addlinespace[0.4em]
Social relationships &
In Grade 7, Shihan maintains generally peaceful relationships with classmates.
She often discusses her artwork with a few like-minded girls during art class or breaks, but appears timid with unfamiliar peers.
After receiving recognition in a class art competition for her distinctive style, she became gradually more willing to join group work, though she still lacks a broad social circle.
Compared with collective activities, she prefers working independently.
When under pressure or facing academic difficulties, she tends to confide in family members or a small number of close friends.
In events such as class dinners or sports days, she often joins the atmosphere when encouraged by nearby classmates, but her willingness to initiate conversations remains limited.
Overall, her social support is moderate: relationships are stable but not wide-reaching. \\

\addlinespace[0.4em]
Creativity profile &
\textbf{Fluency:} medium; she can express personal themes relatively smoothly in drawing.
\textbf{Novelty:} high; she often incorporates distinctive colors and topics.
\textbf{Flexibility:} medium; she can try different art styles but rarely makes bold shifts in thinking.
\textbf{Feasibility:} low; many creative ideas are difficult to implement into complete works.
\textbf{Problem finding:} low outside art; she seldom proactively identifies problems in non-art subjects.
\textbf{Problem analysis:} medium; she can analyze art-related bottlenecks but not deeply.
\textbf{Solution proposing:} low; she lacks systematic plans when turning ideas into concrete artworks.
\textbf{Solution improvement:} low; she shows limited initiative and methods to refine existing plans.
Overall, she shows strong innovative awareness in art but weaker execution and sustained refinement; creativity is salient in one area but not well-balanced. \\

\addlinespace[0.4em]
Mental health (non-clinical, descriptive) &
Shihan is generally sensitive and tends to self-monitor, with noticeable emotional fluctuations under academic pressure and during group activities.
She is quiet and prefers using art to express inner feelings, but she appears less confident in social settings and academic tasks.
Overall, her well-being is slightly below medium; she often feels anxious about learning difficulties or poor performance, especially during exams and subject challenges.
There is no evidence of severe mental disorder; she may occasionally show brief low mood and anxiety tendencies, both at a mid risk level, and the information is insufficient to diagnose any mental illness.
Background: she has practiced art since primary school; her parents strongly support her artistic interests but sometimes place pressure on grades, making her particularly sensitive to criticism or failure.
She mainly copes with stress through family/close-friend support and occasionally through drawing for self-regulation.
Suggested support: strengthen positive family and teacher--student communication, encourage confidence building in a safe environment, gradually increase participation in group activities, help her recognize and express negative emotions, and cultivate self-acceptance. \\

\addlinespace[0.4em]
Sampling constraints (for quota control) &
Grade: Grade 7; Gender: Female; Preferred strength domain: art/music/sports; Target academic level: low (bottom 50\%). \\

\bottomrule
\end{tabularx}
\end{table*}

\FloatBarrier
\section{Offline Distribution Summaries of HACHIMI-1M}
\label{app:hachimi-dist-summary}

As a supplement to the marginal quota diagnostics in Section~\ref{sec:dataset}, we report the offline count summaries of HACHIMI-1M by grade $\times$ academic level and grade $\times$ gender in Table~\ref{tab:hachimi-grade-gender-summary}.

\begin{table}[H]
\centering
\footnotesize
\renewcommand{\arraystretch}{1.05}
\setlength{\tabcolsep}{4pt}
\caption{Offline summary of HACHIMI-1M by grade, academic level, and gender.}
\label{tab:hachimi-grade-gender-summary}
\resizebox{0.6\columnwidth}{!}{%
\begin{tabular}{lccccccc}
\toprule
& \multicolumn{4}{c}{\textbf{Academic Level counts}} & \multicolumn{2}{c}{\textbf{Gender counts}} & \\
\cmidrule(lr){2-5}\cmidrule(lr){6-7}
\textbf{Grade} & \textbf{High} & \textbf{Medium} & \textbf{Low} & \textbf{Poor} & \textbf{Male} & \textbf{Female} & \textbf{Total} \\
\midrule
1  & 21,154 & 20,942 & 21,257 & 21,182 & 42,197 & 42,338 & 84,535 \\
2  & 21,298 & 21,026 & 21,150 & 21,165 & 42,192 & 42,447 & 84,639 \\
3  & 21,116 & 21,329 & 21,111 & 21,240 & 42,306 & 42,490 & 84,796 \\
4  & 21,234 & 21,093 & 20,909 & 20,749 & 41,962 & 42,023 & 83,985 \\
5  & 21,267 & 20,632 & 21,226 & 21,249 & 42,237 & 42,137 & 84,374 \\
6  & 20,693 & 21,225 & 21,192 & 21,279 & 42,153 & 42,236 & 84,389 \\
7  & 21,327 & 21,331 & 21,111 & 21,016 & 42,248 & 42,537 & 84,785 \\
8  & 21,976 & 22,328 & 22,290 & 22,129 & 45,041 & 43,682 & 88,723 \\
9  & 21,745 & 22,071 & 21,785 & 22,062 & 44,028 & 43,635 & 87,663 \\
10 & 21,598 & 21,341 & 21,256 & 21,762 & 43,101 & 42,856 & 85,957 \\
11 & 21,237 & 21,579 & 21,558 & 21,237 & 43,108 & 42,503 & 85,611 \\
12 & 21,317 & 20,963 & 20,943 & 21,627 & 42,270 & 42,580 & 84,850 \\
\bottomrule
\end{tabular}%
}
\end{table}

\FloatBarrier

\section{Intrinsic Evaluation Details}
\label{app:intrinsic-eval}

\subsection{Overview and Data Flow}

We implement a standalone offline evaluator that operates on the merged persona corpus produced by HACHIMI (stored as a JSONL file where each line is a single persona). The evaluator does not modify or retrain any models; it only parses the generated personas and computes a set of post-hoc indicators. All reported intrinsic metrics in \S~\ref{sec:intrinsic-eval} are derived from this evaluator.
\begin{comment}
本节详细说明 HACHIMI 画像集合的内部评估流程与指标实现。我们实现了一个独立的离线评估脚本，对合并后的 JSONL 画像语料进行分析和打分；该评估完全是后验统计，不涉及任何模型更新或再训练。正文中的内在评估结果均由本附录所述方法计算得到。
\end{comment}

% Each persona record is treated as a structured object with fields corresponding to the five theory-anchored components (demographic \& developmental status, academic profile, personality \& values, social relations \& creativity, mental health \& well-being). The evaluator iterates over all records, applies a battery of rule-based checks and corpus-level statistics, and outputs: (i) a global summary of aggregate metrics, and (ii) per-persona diagnostic logs (errors, warnings, near-duplicates, suspicious cases).
\begin{comment}
在实现上，每条画像被视为一个包含五大理论锚定组件的结构化对象。评估器依次遍历所有画像，应用一系列规则检查与语料级统计，并输出两类结果：(i) 全局汇总指标（各类比例、散度、多样性等）；(ii) 逐画像诊断信息（错误与警告列表、近重复配对、可疑样本等），便于后续抽查和迭代优化。
\end{comment}

\subsection{Schema Validity and Theoretical Alignment}

To assess whether the generated personas instantiate the TAD-PG task and theoretical constraints, we define a set of structural and theory-based checks:

\begin{itemize}
  \item \textbf{Required field completeness.} We verify that all key fields are present and non-empty, covering identifiers, demographic fields (name, age, gender, grade), academic profile (achievement level, strong/weak subjects), developmental stages, agent identifier, and the three long-text components (values, creativity, mental health).
  \item \textbf{Academic profile well-formedness.} We enforce that the achievement level is one of four fixed textual anchors (high/ medium / low / poor; see \S~\ref{sec:dataset}), and that strong and weak subject sets are non-empty lists with empty intersection.
  \item \textbf{Agent identifier format.} The “agent name” must follow a constrained \texttt{pinyin+tone} schema: surname and given name separated by an underscore, each part composed of one or more concatenated syllables of the form \texttt{[a-z]+[1-5]}, with the total syllable count corresponding to a plausible Chinese full name length.
  \item \textbf{Paragraph structure.} The three narrative components (values, creativity, mental health) must each be realized as a single, well-formed paragraph (no bullet lists or double line breaks). We also monitor paragraph character lengths and flag texts that are unusually short or long relative to predefined thresholds.
  \item \textbf{Value and creativity dimensions.} The values paragraph must explicitly mention the seven value dimensions (moral cultivation, physical and mental health, rule-of-law awareness, social responsibility, political identity, cultural literacy, family orientation) and include level expressions (e.g., high, medium, low). The creativity paragraph must cover eight creativity/CPS dimensions and contain a brief “radar-style” summary, with heuristics to flag logically inconsistent combinations (e.g., very low feasibility but unusually high solution-generation).
  \item \textbf{Mental health slots and non-diagnostic language.} The mental-health text is checked for the presence of key slots (overall mental status, happiness index, risk descriptions, stressors, supports, coping strategies) and for adherence to non-diagnostic educational language. We additionally flag combinations where mental-health descriptions are incompatible with the value dimension “physical and mental health” (e.g., “excellent health” paired with “severe depression”).
  \item \textbf{Age--grade and developmental-stage consistency.} We apply coarse age--grade consistency checks using expected age ranges for each grade (primary, lower secondary, upper secondary). We also check whether Piagetian, Eriksonian and Kohlbergian stages are roughly compatible with the student's age band, and flag obvious violations (e.g., pre-operational stage assigned to a high-school student).
\end{itemize}
\begin{comment}
为评估画像是否真正落实了 TAD-PG 的模式与理论约束，我们定义了若干结构性与理论性检查：
(1) 必备字段完整性：检查标识、人口学字段（姓名、年龄、性别、年级）、学业档案（学术水平、擅长/薄弱科目）、发展阶段、代理名以及三段长文本（价值观、创造力、心理健康）是否存在且非空；
(2) 学业档案规范性：限定学术水平必须为四种固定文案之一，且擅长科目与薄弱科目均为非空列表且无交集；
(3) 代理名规范：要求姓名遵循”拼音+声调”的约束模式（姓与名用下划线分隔，每段由若干 [a-z]+[1-5] 音节串接，总音节数对应合理的中文姓名长度），便于后续引用和调用；
(4) 段落体裁：要求价值观、创造力、心理健康均为单一自然段（禁止条目式或双换行），并通过字数阈值标记过短或过长文本；
(5) 价值观与创造力维度覆盖：价值观段需显式覆盖七个价值维度并给出等级词，创造力段需覆盖八个创造/问题解决维度并包含”雷达式”总结，同时用启发式规则标记明显不一致组合；
(6) 心理健康槽位与非诊断化：心理健康文本需包含整体状况、幸福指数、风险描述、压力与保护因素、支持与应对等关键槽位，并检测其与价值观中”身心健康”维度的一致性，避免出现”身心极佳”与”重度抑郁”并存；
(7) 年龄–年级与发展阶段一致性：依据常识年龄段对年级和皮亚杰/埃里克森/科尔伯格阶段进行粗略匹配，标记明显越界情形。
\end{comment}

For each persona, the evaluator records violations under a concrete taxonomy of \emph{errors} (hard schema violations) and \emph{warnings} (soft theoretical mismatches):

\begin{itemize}
  \item \textbf{Errors (hard schema violations).} A persona is marked as having an error if any of the following holds:
  \begin{itemize}
    \item \emph{Missing or empty required field:} any of the core slots is absent or empty (identifiers, age, gender, grade, academic level, strong/weak subject lists, developmental stages, agent identifier, or any of the three long-text components).
    \item \emph{Ill-formed academic profile:} the achievement level does not match one of the four allowed anchors (high / medium / low / poor), or the strong/weak subject lists are empty or have non-empty intersection.
    \item \emph{Malformed agent identifier:} the agent name violates the constrained \texttt{pinyin+tone} pattern (surname\_givenname, with syllables of the form \texttt{[a-z]+[1-5]} and a plausible total syllable count).
    \item \emph{Invalid paragraph format:} any of the values, creativity, or mental-health components is not realized as a single paragraph (e.g., list-like formatting, multiple blank lines) and thus does not satisfy the schema.
    \item \emph{Missing theory-mandated dimensions:} the values text fails to cover all seven value dimensions with explicit level words, the creativity text omits one or more of the eight CPS dimensions, or the mental-health text lacks core slots such as overall status, risk/stressors, supports, or coping strategies.
    \item \emph{Age--grade / stage inconsistency:} the age–grade combination falls outside the allowed ranges for compulsory schooling, or the Piagetian / Eriksonian / Kohlbergian stages are incompatible with the age band in a way that cannot be justified as borderline (e.g., pre-operational stage assigned to an upper-secondary student).
  \end{itemize}

  \item \textbf{Warnings (soft theoretical mismatches).} A persona is marked as having a warning if it passes the hard schema checks but triggers any of the following softer heuristics:
  \begin{itemize}
    \item \emph{Atypical paragraph length:} values, creativity, or mental-health paragraphs are unusually short or long relative to target ranges (flagging potential under-specification or verbosity).
    \item \emph{Partial dimension coverage:} some but not all value or creativity dimensions are mentioned, or some level words are vague or missing, while the overall structure remains acceptable.
    \item \emph{Creativity profile inconsistencies:} “radar-style” summaries contain mild contradictions (e.g., very low feasibility paired with extremely high overall CPS score) that are not severe enough to reject the persona.
    \item \emph{Mental-health / value mismatches:} the mental-health text uses educationally undesirable diagnostic labels (e.g., clinical disorder names) or shows tension with the “physical and mental health” value dimension (e.g., “excellent health” co-occurring with strong negative symptom descriptions).
    \item \emph{Borderline developmental mismatch:} the age band and developmental stages are slightly misaligned (e.g., a younger-than-typical student assigned to a later stage), but still within a borderline interpretable range.
  \end{itemize}
\end{itemize}

At the corpus level, we report the proportion of personas with at least one error or warning, as well as summary statistics such as paragraph-length distributions. In addition, we construct simple text-derived “alignment scores” by mapping level words in the values, creativity, and mental-health texts to polarity weights and computing Pearson correlations between these scores and the four-level academic-achievement anchor; these correlations quantify how consistently textual descriptions track the intended academic tiers.

\begin{comment}
对每条画像，评估器会按照一个具体的分类方案，将违规情况标记为”错误”（硬模式违规）或”警告”（软理论不一致）：

(1) 错误：只要出现以下任一情况即视为模式错误：
  - 必备字段缺失或为空：标识、年龄、性别、年级、学业水平、擅长/薄弱科目列表、发展阶段、代理名或三段长文本任一缺失；
  - 学业档案不规范：学业水平不在四个固定锚点（高/中/低/差）中，或擅长/薄弱科目列表为空、存在交集；
  - 代理名格式错误：不符合 ”姓\_名 + 拼音+声调” 规则（[a-z]+[1--5] 串、音节数不合理等）；
  - 段落体裁违规：价值观、创造力或心理健康部分未以单一自然段呈现（如列表、双换行等）；
  - 理论必需维度缺失：价值观未覆盖七个价值维度或缺少等级词；创造力未覆盖八个创造/CPS 维度；心理健康缺少整体状况、风险/压力、支持或应对等关键槽位；
  - 年龄–年级/发展阶段严重不符：年龄与年级超出义务教育合理范围，或皮亚杰/埃里克森/科尔伯格阶段与年龄段明显不兼容（如高中生被标为前运算阶段）。

(2) 警告：若通过硬模式检查，但触发以下任一软性规则，则标为警告：
  - 段落长度异常：价值观、创造力或心理健康过短或过长，提示可能欠描写或过度冗长；
  - 维度覆盖不完全：只覆盖了部分价值观或创造力维度，或等级用语含糊，但整体结构尚可接受；
  - 创造力雷达轻微矛盾：雷达式总结中存在轻度的不一致（如可行性极低但总体创造力极高），但不足以拒绝画像；
  - 心理健康与价值维度张力：心理健康段使用了过于诊断化的标签，或与价值观中“身心健康”维度存在张力（如”身心极佳”同时伴随强烈症状描述）；
  - 发展阶段边界模糊：年龄与阶段略有错位，但仍可被解释为边界情形。

在语料级，我们统计至少含有一个错误或一个警告的画像比例，并报告段落长度等分布统计。同时，我们通过将价值观、创造力与心理健康文本中的等级词映射为权重，构造简单的文本打分，并与四级学业水平锚点做皮尔逊相关，以衡量长文本描述是否随学业层级按预期变化。
\end{comment}

\subsection{Distributional Alignment with Target Quotas}

To evaluate the “distribution-controllable” aspect of TAD-PG, we compare the empirical distributions of key stratification variables against target quotas or generation plans:

\begin{itemize}
  \item \textbf{Marginal distributions.} We compute empirical frequency distributions over grade, gender and academic-achievement level for the entire corpus and treat them as empirical distributions $P(\cdot)$. We compare these to pre-specified target distributions $Q(\cdot)$ (e.g., uniform over grades, balanced gender, uniform over the four academic tiers) using absolute deviation and Kullback--Leibler (KL) divergence $\mathrm{KL}(P \Vert Q)$ with simple Laplace smoothing.
  \item \textbf{Plan-based alignment (optional).} When a generation schedule is available (e.g., a JSON plan recording the intended grade, gender and academic level for each scheduled persona), we compute per-field \emph{match rates} between generated and planned labels, as well as KL divergence between the empirical corpus distributions and the plan-induced distributions. This directly measures how tightly HACHIMI follows a concrete quota schedule when one is used.
\end{itemize}
\begin{comment}
为验证 TAD-PG 在”分布可控”维度上的实现，我们将画像语料中的经验分布与先验或生成计划进行对齐度分析：
(1) 边缘分布：对年级、性别、学术水平等变量统计经验频率分布 $P(\cdot)$，并与预设的目标分布 $Q(\cdot)$（如年级均匀、性别接近 1:1、学业四档均匀）进行比较，采用绝对偏差和 KL 散度 $\mathrm{KL}(P \Vert Q)$（带拉普拉斯平滑）作为量化指标；
(2) 计划对齐（可选）：当存在生成计划（如 schedule.json 记录了每条画像目标的年级/性别/学术水平）时，我们进一步计算生成结果与计划标签的一致率，以及语料经验分布相对于计划分布的 KL 散度，用于直接衡量 HACHIMI 在实际运行中对配额调度的遵从程度。
\end{comment}

These distributional diagnostics are summarized in the evaluation report as overall KL divergence values for each variable and, when applicable, as plan-alignment metrics. They provide a corpus-level check that HACHIMI does not drift away from the intended grade, gender and achievement landscape when scaling up to large persona populations.
\begin{comment}
上述分布诊断会在评估汇总中以每个变量的 KL 散度和（如适用）计划对齐指标的形式呈现，为 HACHIMI 在大规模生成时是否偏离预期的年级–性别–学业结构提供了一个直观的全局检查。
\end{comment}

\subsection{Diversity and Redundancy Control}

To ensure that the persona collection covers a wide variety of student types rather than collapsing into a few templates, we compute several diversity and redundancy indicators on concatenated persona texts:

\begin{itemize}
  \item \textbf{Distinct-$n$}. We compute character-level Distinct-1 and Distinct-2 at the per-sample level (ratio of unique $n$-grams to all $n$-grams, averaged across personas) over a concatenation of key narrative fields. Higher Distinct-$n$ values indicate richer lexical and phrasal variety.

  \item \textbf{SimHash near-duplicate detection.} For each persona, we compute a 64-bit SimHash over character trigrams of the concatenated long texts. We then approximate the distribution of pairwise Hamming distances via random sampling and record the number of persona pairs whose distance falls below a small threshold (i.e., near-duplicates). This provides both a global sense of redundancy and a concrete list of highly similar profile pairs.
  \item \textbf{Cross-component similarity.} We compute character-level Jaccard similarity between the values, creativity and mental-health paragraphs for each persona and flag cases where any pair exceeds a high threshold (e.g., $0.8$). These “template-like” samples are counted as potential overuse of shared boilerplate across components.
\end{itemize}
\begin{comment}
为验证画像集合在语义层面是否具备足够多样性，而非仅是少数模板的微调版本，我们在关键文本字段上计算多种多样性与冗余指标：
(1) Distinct-$n$：基于字符级 $n$-gram，在单条画像上计算 Distinct-1/2（唯一 $n$-gram 数除以总 $n$-gram 数），再在样本上取平均；数值越高通常表示词汇与短语层面的多样性越好；

(3) SimHash 近重复检测：对合并长文本计算 64 位 SimHash，并通过随机对采样估计汉明距离分布，同时统计距离低于阈值的”近重复”画像对数量，为冗余程度提供定量信号与具体样本；
(4) 组件间相似度：对每条画像的价值观、创造力、心理健康三段文本逐对计算字符级 Jaccard 相似度，并对超过高阈值（如 0.8）的样本进行计数，作为多组件共用模版的潜在信号。
\end{comment}

The evaluator reports corpus-level Distinct-$n$ values, summary statistics of SimHash Hamming distances, the count of near-duplicate pairs, and the number of template-like samples with excessive cross-component similarity. Collectively, these metrics characterize how well HACHIMI balances structural constraints with semantic variety.
\begin{comment}
评估输出中会给出 Distinct-$n$ 的全局数值、SimHash 汉明距离的均值与方差、近重复画像对数量以及”疑似模板化”样本计数等，从而整体刻画 HACHIMI 在满足结构约束的同时，是否仍然保持了足够丰富的语义差异。
\end{comment}

\subsection{Local Coherence and Contradiction Checks}

Finally, we perform basic local-coherence checks to guard against salient contradictions within individual personas. These checks are primarily rule-based:

\begin{itemize}
  \item \textbf{Hard contradictions.} We search for patterns where different fields convey mutually incompatible information, such as: extremely high academic achievement paired with long-term, severe perceived difficulty in the same subject; very high physical and mental health ratings co-occurring with explicit descriptions of extreme psychological distress; or age--grade combinations that are strongly implausible.
  \item \textbf{Slot-level consistency.} Within the mental-health component, we cross-check the overall mental status, happiness index and risk descriptions for depression/anxiety to ensure that severity terms (e.g., mild, moderate, severe) are not obviously inconsistent with the qualitative description of functioning and support.
  \item \textbf{Subject-mention coverage.} As a weak form of coherence between structured and narrative fields, we measure the proportion of personas for which strong/weak subjects explicitly appear in the long-text descriptions (values, social relations, creativity, mental health), indicating that narrative content meaningfully reflects the structured academic profile.
\end{itemize}
\begin{comment}
我们还对单条画像执行若干局部自洽性检查，以识别明显的内部矛盾。这些检查主要基于启发式规则：
(1) 硬矛盾：搜索字段之间互相否定的模式，例如”学业成绩全校前 10\%”却长期自述”在同一学科上极度吃力”；”身心健康优秀”同时伴随大量极端心理痛苦描述；或极不合理的年龄–年级组合等；
(2) 槽位一致性：在心理健康文本内部交叉检查整体心理状况、幸福指数与抑郁/焦虑风险等级，避免严重程度用语与功能和支持描述明显不符；
(3) 学科提及覆盖度：作为结构字段与叙事文本之间一致性的弱信号，统计擅长/薄弱科目在长文本中被明确提及的比例，检验学业档案是否在叙事层面得到体现。
\end{comment}

At the corpus level, we summarize the fraction of personas exhibiting any hard contradiction, the subject-mention coverage rates, and the distribution of errors and warnings per persona. We also identify the top-\,$k$ most problematic personas (with the largest number of errors/warnings) to facilitate manual inspection. These diagnostics serve as a sanity check that TAD-PG constraints are not only satisfied at the field level, but also reflected in self-consistent narratives.
\begin{comment}
在语料级，我们报告存在硬矛盾的画像比例、学科提及覆盖率以及每条画像的错误/警告数量分布，并列出若干”最可疑”样本以便人工抽查。通过这些诊断，我们可以验证 TAD-PG 约束不仅在字段层面被满足，而且在叙事层面也基本保持了自洽。
\end{comment}

\section{Details of the CEPS-based Consistency Evaluation}
\label{app:ceps-pipeline}

\subsection{Data Preprocessing and Cohort Construction}
\label{app:ceps-pipeline-preprocessing}

This section documents how we preprocess CEPS Grade~8 data and construct the stratified cohorts used as human baselines in \S~\ref{sec:ceps-consistency}.
\begin{comment}
本小节补充说明 CEPS 八年级数据的预处理流程，以及正文第~\ref{sec:ceps-consistency} 节中所用分层群体是如何构建的，作为真实学生端的基准。
\end{comment}

\paragraph{Raw data and basic filters.}
We use the 2014--2015 wave of the China Education Panel Survey (CEPS) student questionnaire and retain only Grade~8 students.\footnote{We rely on the official grade indicator provided in the CEPS documentation; implementation-wise this is a simple filter on the grade field.} We drop records with missing values in any of the key fields needed for cohort construction (core subject scores, gender and depression scale items), resulting in $N_{\mathrm{CEPS}}$ valid Grade~8 cases.
\begin{comment}
原始数据来自 CEPS 2014--2015 学年的学生问卷，我们仅保留八年级学生（通过官方年级字段筛选）。若在构建分层群体所需的关键变量（核心学科成绩、性别、抑郁量表条目）存在缺失，则剔除该样本，最终得到 $N_{\mathrm{CEPS}}$ 份有效的八年级问卷记录。
\end{comment}

\paragraph{Gender and gender imputation.}
For gender, we use the official CEPS gender variable and treat it as binary (M/F). A small fraction of cases lack this field; for these, we perform gender imputation based on mutually exclusive puberty indicators (e.g., age at first menstruation vs.\ age at first nocturnal emission). Concretely, we assign ``F” when only the menarche item is non-missing and ``M” when only the nocturnal-emission item is non-missing; records with ambiguous or missing puberty information are dropped.
\begin{comment}
性别首先直接使用 CEPS 提供的性别变量，并视为二元（男/女）。对少量性别缺失的样本，我们利用青春期发育条目进行性别插补：若仅初潮年龄有值，则判定为女；若仅遗精年龄有值，则判定为男；若两者均缺失或信息矛盾，则剔除该样本。
\end{comment}

\paragraph{Academic achievement level.}
We use the raw scores of Chinese, mathematics and English in CEPS Grade~8 (denoted $s_{\mathrm{chn}}$, $s_{\mathrm{mat}}$ and $s_{\mathrm{eng}}$). Each subject score is first standardised within the Grade~8 population,
\begin{equation}
z_{\mathrm{chn}} = \frac{s_{\mathrm{chn}} - \mu_{\mathrm{chn}}}{\sigma_{\mathrm{chn}}}
\end{equation}
\begin{equation}
z_{\mathrm{mat}} = \frac{s_{\mathrm{mat}} - \mu_{\mathrm{mat}}}{\sigma_{\mathrm{mat}}}
\end{equation}
\begin{equation}
z_{\mathrm{eng}} = \frac{s_{\mathrm{eng}} - \mu_{\mathrm{eng}}}{\sigma_{\mathrm{eng}}}
\end{equation}
We then form a total achievement index
\begin{equation}
z_{\mathrm{total}} = z_{\mathrm{chn}} + z_{\mathrm{mat}} + z_{\mathrm{eng}}.
\end{equation}
The distribution of $z_{\mathrm{total}}$ is sliced into four tiers by empirical percentiles:
\begin{itemize}
  \item \emph{High}: top $10\%$ of $z_{\mathrm{total}}$;
  \item \emph{Meduim}: $10$--$30\%$;
  \item \emph{Low}: $30$--$50\%$;
  \item \emph{Poor}: bottom $50\%$.
\end{itemize}
These four tiers are used as the academic-level labels in \S~\ref{sec:ceps-consistency}.
\begin{comment}
学业水平基于语文、数学和英语三科原始分数。首先在八年级总体内对每科成绩做标准化，得到 $z_{\mathrm{chn}}$、$z_{\mathrm{mat}}$、$z_{\mathrm{eng}}$，再将三科 $z$ 分数求和得到总体学业指标 $z_{\mathrm{total}}$。随后按 $z_{\mathrm{total}}$ 的经验分布划分四档：前 10\% 为 High，10--30\% 为 Mid，30--50\% 为 Low，后 50\% 为 Poor，这四档即正文所述的学业分层。
\end{comment}

\paragraph{Psychological risk from CES-D.}
To characterise psychological risk, we use the CES-D short depression scale embedded in CEPS. Let $d_1,\dots,d_{10}$ denote the ten item responses (coded on the original CEPS Likert scale). We compute a simple sum score
\begin{equation}
\mathrm{CESD} = \sum_{k=1}^{10} d_k.
\end{equation}
We then define ``high” vs.\ ``low” psychological risk via the empirical $75$th percentile of $\mathrm{CESD}$ within Grade~8:
\begin{itemize}
  \item \emph{High-risk}: $\mathrm{CESD} \geq$ 75th percentile;
  \item \emph{Low-risk}: $\mathrm{CESD} <$ 75th percentile.
\end{itemize}
We do not apply any further weighting or clinical cutoffs, as our goal is to obtain a stable rank-based risk stratification rather than a diagnostic label.
\begin{comment}
心理风险使用 CEPS 中嵌入的 CES-D 抑郁量表（10 题版）。记 $d_1,\dots,d_{10}$ 为各条目的原始量表得分，我们计算总分
\begin{equation}
\mathrm{CESD} = \sum_{k=1}^{10} d_k.
\end{equation}
并以年级内 CESD 的 75\% 分位数作为阈值：总分位于前 25\% 的学生标记为高风险，其余为低风险。这里不尝试给出临床诊断，而是构建一个稳定的风险分层。
\end{comment}

\paragraph{Resulting stratified cohorts.}
Combining the four academic levels, two gender categories and two psychological risk levels yields $4 \times 2 \times 2 = 16$ mutually exclusive cohorts. Each Grade~8 student is assigned a unique triplet (academic level, gender, risk), and group-wise statistics (e.g., item means) are computed by taking unweighted averages within each cohort.
\begin{comment}
将四档学业水平、二元性别和高/低心理风险组合后，可以得到 $4 \times 2 \times 2 = 16$ 个互斥群体。每名八年级学生被唯一分配到其中一个组合（学业水平 × 性别 × 心理风险），随后在每个群体内对各问卷条目做简单算术平均，从而获得群体均值等统计量。
\end{comment}

\subsection{Consistency Metrics and Implementation Details}
\label{app:ceps-pipeline-CMdetails}

This section details how we construct the human and agent cohort-level statistics and compute the consistency metrics reported in \S~\ref{sec:ceps-consistency}.
\begin{comment}
本小节补充说明真实学生和 Agent 端的群体统计量是如何计算的，以及正文第~\ref{sec:ceps-consistency} 节所用的一致性指标在实现层面的细节。
\end{comment}

\paragraph{Item selection and coding.}
From the CEPS Grade~8 student questionnaire we select a subset of items that satisfy two criteria: (i) they reflect perceptions, attitudes or self-reported behaviours that can reasonably be inferred from a textual persona (e.g., perceived difficulty in mathematics, perceived parental expectations, class climate, prosocial or problem behaviours); and (ii) they have well-defined discrete response options with stable coding schemes. Purely factual items that cannot be inferred from the persona (such as height or weight) are excluded. We keep the original item IDs and numeric codes, and re-use them on the agent side so that human and agent responses are aligned at the code level.
\begin{comment}
在 CEPS 八年级学生问卷中，我们筛选用于一致性分析的条目，标准为：（i）题目反映的是可从画像合理推断的感知、态度或自陈行为（如数学吃力感、父母成绩期望、班级氛围、助人/攻击行为等）；（ii）具备明确的有限选项及稳定的编码规则。无法从画像推断的纯事实题（如身高、体重）被剔除。我们保留原始的题目编号和数值编码，并在 Agent 端沿用这些编码，以保证人类与 Agent 的作答在编码层面一一对应。
\end{comment}

\paragraph{Human cohort means.}
For each selected CEPS item $j$ and each of the 16 cohorts $g \in \{1,\dots,16\}$, we compute the unweighted mean
\begin{equation}
\mu^{(H)}_{g,j} = \frac{1}{\lvert S_{g,j} \rvert} \sum_{i \in S_{g,j}} x_{i,j}
\end{equation}
where $x_{i,j}$ is the numeric response of student $i$ to item $j$, and $S_{g,j}$ is the set of students in cohort $g$ with non-missing responses on $j$. These 16-dimensional vectors
\begin{equation}
\mathbf{\mu}^{(H)}_j = \big(\mu^{(H)}_{1,j}, \dots, \mu^{(H)}_{16,j}\big)
\end{equation}
are treated as the human reference patterns for item $j$.
\begin{comment}
在真实学生端，对于每一道被选中的题目 $j$，以及 16 个群体中的任一群体 $g$，我们在该群体内对非缺失样本的作答取简单平均，得到
\begin{equation}
\mu^{(H)}_{g,j} = \frac{1}{\lvert S_{g,j} \rvert} \sum_{i \in S_{g,j}} x_{i,j}.
\end{equation}
由此形成长度为 16 的向量
\begin{equation}
\mathbf{\mu}^{(H)}_j = \big(\mu^{(H)}_{1,j}, \dots, \mu^{(H)}_{16,j}\big),
\end{equation}
作为该题在真实学生端的群体模式。
\end{comment}

\paragraph{Agent cohort means.}
On the agent side, each sampled persona is instantiated as a student agent and prompted with the CEPS-based shadow survey (Figure~\ref{fig:prompt_template}). The LLM outputs are constrained to a JSON object whose keys are CEPS item IDs and whose values are the chosen option codes. After basic validation, we assign each agent to one of the 16 cohorts (using the academic level, gender and psychological risk labels embedded in the persona) and compute cohort means in exactly the same way as for humans:
\begin{equation}
\mu^{(A)}_{g,j} = \frac{1}{\lvert T_{g,j} \rvert} \sum_{a \in T_{g,j}} y_{a,j}
\end{equation}
where $y_{a,j}$ is the numeric response of agent $a$ to item $j$, and $T_{g,j}$ is the set of agents in cohort $g$ with valid responses on $j$. This yields agent-side vectors
\begin{equation}
\mathbf{\mu}^{(A)}_j = \big(\mu^{(A)}_{1,j}, \dots, \mu^{(A)}_{16,j}\big)
\end{equation}
for each item $j$.
\begin{comment}
在 Agent 端，我们将每个被抽样的画像实例化为学生代理，并使用图~\ref{fig:prompt_template} 所示的提示词让其完成 CEPS 影子问卷。模型输出被约束为一个 JSON 对象，键为 CEPS 题目编号，值为所选选项的编码。通过基础合法性检查后，我们依据画像内的学业水平、性别和心理风险标签将每个 Agent 映射到 16 个群体之一，并与真实学生同样地计算群体均值：
\begin{equation}
\mu^{(A)}_{g,j} = \frac{1}{\lvert T_{g,j} \rvert} \sum_{a \in T_{g,j}} y_{a,j},
\end{equation}
由此得到 Agent 端对每题 $j$ 的 16 维群体均值向量
\begin{equation}
\mathbf{\mu}^{(A)}_j = \big(\mu^{(A)}_{1,j}, \dots, \mu^{(A)}_{16,j}\big)。
\end{equation}
\end{comment}

\paragraph{Pearson and Spearman correlations.}
For each item $j$, we compute two scalar consistency metrics between humans and agents:
\begin{itemize}
  \item \textbf{Pearson correlation} $r_j$ between $\mathbf{\mu}^{(H)}_j$ and $\mathbf{\mu}^{(A)}_j$, capturing \emph{linear trend consistency} in absolute cohort means;
  \item \textbf{Spearman rank correlation} $\rho_j$ between the same vectors, capturing \emph{rank-order consistency} in the relative ordering of cohorts.
\end{itemize}
Ties in cohort means are handled using average ranks in the Spearman computation. Items for which fewer than two cohorts have non-missing means on either side are excluded from correlation analysis.
\begin{comment}
对于每一道题目 $j$，我们在真实学生向量 $\mathbf{\mu}^{(H)}_j$ 与 Agent 向量 $\mathbf{\mu}^{(A)}_j$ 之间计算两个标量指标：（1）皮尔逊相关系数 $r_j$，反映绝对群体均值的线性趋势是否一致；（2）斯皮尔曼等级相关系数 $\rho_j$，反映群体相对排序是否一致。计算 $\rho_j$ 时，若群体均值存在并列，则采用平均秩处理。若某题在任一端可用群体数少于 2，则从相关性分析中剔除。
\end{comment}

\paragraph{Summary statistics and visualisation.}
In the main text we summarise the distribution of $\{r_j\}$ and $\{\rho_j\}$ across all evaluated items (e.g., mean, standard deviation and selected quantiles), and provide representative visualisations: scatter plots of human vs.\ agent cohort means for individual items, and grouped bar charts comparing the 16 cohort means for selected items that illustrate typical agreement or disagreement patterns. All statistics are computed in Python using standard scientific libraries (e.g., \texttt{pandas}, \texttt{numpy} and \texttt{scipy}).
\begin{comment}
在正文中，我们对所有题目的 $\{r_j\}$ 和 $\{\rho_j\}$ 分布给出汇总统计（如均值、标准差和若干分位数），并展示典型可视化：包括单题的人类均值 vs. Agent 均值散点图，以及在若干关键题目上 16 个群体均值的分组柱状图，用以直观呈现一致或不一致的模式。所有统计量均使用 Python 科学计算库（如 \texttt{pandas}、\texttt{numpy}、\texttt{scipy}）实现。
\end{comment}
\subsection{CEPS Constructs and Aggregation Rules}
\label{app:ceps-constructs}

For the CEPS-based consistency analysis in \S~\ref{sec:rq2-results}, we select a focused subset of Grade~8 questionnaire items and group them into a small number of higher-level constructs.
All raw items follow the original CEPS coding, and we use simple additive or averaging rules to obtain construct scores at the individual level before computing cohort means.
For compactness, we denote the construct scores by
$D(i)$,
$S_{\text{parent}}(i)$,
$T_{\text{att}}(i)$,
$M(i)$,
$P_{\text{prosocial}}(i)$
and
$B_{\text{school}}(i)$
for the six constructs below.
\begin{comment}
为便于排版，下文在公式中分别用
$D(i)$（抑郁症状）、
$S_{\text{parent}}(i)$（父母管教严厉度）、
$T_{\text{att}}(i)$（教师关注度）、
$M(i)$（不良行为频率）、
$P_{\text{prosocial}}(i)$（亲社会行为）、
$B_{\text{school}}(i)$（学校归属感）
来表示六个构念的个体得分。
\end{comment}

\paragraph{Depressive symptoms.}
The construct \texttt{Construct\_Depression} summarises psychological distress based on the ten C25 items (\texttt{w2c25{01--10}}), which ask how often students experience symptoms such as sadness, worry, sleep problems or loss of appetite.
Each item is coded on a 1--5 frequency scale (``never'' to ``always'').
We compute the construct score as the row-wise sum:
\begin{equation}
  D(i) = \sum_{j=1}^{10} \texttt{w2c25}j(i),
\end{equation}
with higher values indicating higher psychological risk.
\begin{comment}
抑郁/焦虑症状构念 \texttt{Construct\_Depression} 基于 C25 表中的 10 个条目（\texttt{w2c25{01--10}}），每题 1–5 级频率量表，我们在个体层面对 10 题求和得到
$D(i)$，得分越高代表心理风险越高。
\end{comment}

\paragraph{Parental strictness.}
The construct \texttt{Construct\_Parental\_Strictness} captures how strictly parents regulate academic work and media use, using A20-1 (\texttt{w2a2001}, homework supervision) and A20-5 (\texttt{w2a2005}, internet supervision), both on a 1--3 scale.
We define
\begin{equation}
  S_{\text{parent}}(i)
  = \frac{\texttt{w2a2001}(i) + \texttt{w2a2005}(i)}{2}.
\end{equation}
\begin{comment}
父母管教严厉度构念基于 A20-1 与 A20-5 两题，均为 1–3 级量表；
我们对两题取平均得到
$S_{\text{parent}}(i)$。
\end{comment}

\paragraph{Teacher attention.}
The construct \texttt{Construct\_Teacher\_Attention\_Avg} aggregates perceived teacher attention in mathematics, Chinese and English (B5-1 / \texttt{w2b0501}, B5-2 / \texttt{w2b0502}, B5-3 / \texttt{w2b0503}), each coded 1--4.
We compute

\begin{equation}
T_{\text{att}}(i)
= \frac{\texttt{w2b0501}(i) + \texttt{w2b0502}(i) + \texttt{w2b0503}(i)}{3}.
\end{equation}
\begin{comment}
教师关注度构念基于 B5-1/2/3 三题（数学、语文、英语老师关注度），每题 1–4 级量表，我们对三题取平均得到
$T_{\text{att}}(i)$。
\end{comment}

\paragraph{Misbehaviour.}
The construct \texttt{Construct\_Misbehavior} captures externalising problem behaviours using D2-1 (\texttt{w2d0201}, swearing) and D2-3 (\texttt{w2d0203}, fighting), both coded 1--5.
We define
\begin{equation}
  M(i)
  = \frac{\texttt{w2d0201}(i) + \texttt{w2d0203}(i)}{2},
\end{equation}
with higher values indicating more frequent misbehaviour.
\begin{comment}
不良行为频率构念基于 D2-1（骂人说脏话）与 D2-3（打架）两题，1–5 级频率量表，对两题取平均得到
$M(i)$，得分越高表示不良行为越频繁。
\end{comment}

\paragraph{Prosocial behaviour.}
The construct \texttt{Construct\_Prosocial} reflects prosocial tendencies using D1-1 (\texttt{w2d0101}, helping others) and D1-2 (\texttt{w2d0102}, following rules), again on a 1--5 scale:
\begin{equation}
  P_{\text{prosocial}}(i)
  = \frac{\texttt{w2d0101}(i) + \texttt{w2d0102}(i)}{2}.
\end{equation}
\begin{comment}
亲社会行为构念基于 D1-1（助人）与 D1-2（守序）两题，1–5 级频率量表，对两题取平均得到
$P_{\text{prosocial}}(i)$，得分越高表示更频繁的助人和守规则行为。
\end{comment}

\paragraph{School bonding.}
The construct \texttt{Construct\_School\_Bonding} captures students' sense of connection to their class and school via B6-6 (\texttt{w2b0606}, class climate) and B6-7 (\texttt{w2b0607}, participation in activities), both coded 1--4.
We compute
\begin{equation}
  B_{\text{school}}(i)
  = \frac{\texttt{w2b0606}(i) + \texttt{w2b0607}(i)}{2}.
\end{equation}
\begin{comment}
学校归属感构念基于 B6-6（班风）与 B6-7（活动参与）两题，1–4 级同意度量表，对两题取平均得到
$B_{\text{school}}(i)$，得分越高表示对班级/学校的连结感更强。
\end{comment}

\section{Details of the PISA 2022-based Consistency Evaluation}
\label{app:pisa-pipeline}

\subsection{PISA 2022 Variable Selection and Preprocessing}
\label{app:pisa-vars-preprocessing}

We base our PISA-side analysis on the public PISA~2022 student questionnaire data~\cite{oecd2023pisa2022db}. To obtain construct scores that are comparable to our HACHIMI agents, we proceed in three steps: variable selection, basic filtering, and standardisation.

\paragraph{Construct selection.}
We focus on a set of psycho-social and learning-related OECD student-questionnaire indices (WLE) that reflect students' motivation, affect, well-being, and creativity. Concretely, we include:
\begin{itemize}
  \item mathematics self-efficacy / efficacy indices (e.g., \texttt{MATHEFF}, \texttt{MATHEF21});
  \item mathematics anxiety (e.g., \texttt{ANXMAT});
  \item sense of belonging at school (e.g., \texttt{BELONG});
  \item life satisfaction (e.g., \texttt{LIFESAT});
  \item psychosomatic symptoms / psychological distress (e.g., \texttt{PSYCHSYM});
  \item creative self-efficacy and creativity/openness to intellect (e.g., \texttt{CREATEFF}, \texttt{CREATOP});
  \item social connections (ease of communication about worries and concerns) (e.g., \texttt{SOCCON}).
\end{itemize}
All selected variables are official PISA indices constructed by the OECD from multiple questionnaire items (reported as WLE scale scores). We use these released index scores directly (without re-deriving them from item-level responses) and only apply basic filtering and standardisation to align their scale and comparability with the agent-side construct scores.

For the purpose of defining achievement-based cohorts (Appendix~\ref{app:pisa-cohorts}), we additionally use the full set of plausible values for mathematics, reading, science, and mathematics content/process subscales (e.g., \texttt{PV1MATH--PV10MATH}, \texttt{PV1READ--PV10READ}, etc.). We aggregate these into a composite achievement score as described below.

\paragraph{Sample filtering.}
We retain only students with valid values on:
(i) the selected psycho-social constructs,
(ii) the achievement plausible-value variables used to construct the composite score,
and (iii) basic demographics (country, gender).
Students with missing or invalid codes on any of these fields are dropped.
No survey weights are applied, as our goal is to compare cohort-level patterns rather than to produce nationally representative estimates.

\paragraph{Normalisation.}
For descriptive reporting and cross-region comparability, we z-normalise each psycho-social construct separately within the pooled PISA sample:
\begin{equation}
  z_{i,c} = \frac{x_{i,c} - \mu_c}{\sigma_c},
\end{equation}
where $x_{i,c}$ is the raw score of student $i$ on construct $c$, and $\mu_c$, $\sigma_c$ are the mean and standard deviation of $c$ across all retained students.
For grouping and correlation analyses, we use these normalised scores but keep the original directionality (i.e., higher values always mean more of the underlying construct; reverse-coded scales are flipped where necessary).
\begin{comment}
本节说明 PISA 侧变量如何选取与预处理。我们基于 OECD 公布的 PISA 2022 学生问卷数据，从官方指数中选取若干与动机、情绪、幸福感和创造力相关的构念（如 MATHEFF, ANXMAT, BELONG, LIFESAT, PSYCHSYM, CREATEFF, CREATOP, SOCCON 等），直接使用 OECD 已聚合好的量表分数。为构建学业分层，我们进一步使用数学、阅读、科学及数学各内容/过程子量表的全部 plausible values（如 PV1MATH--PV10MATH, PV1READ--PV10READ 等），在后文汇总为一个综合学业得分。样本过滤时，仅保留这些构念、成就相关 PV 以及基础人口学信息（国家、性别）无缺失的学生，不使用权重，因为我们的目标是对比群体模式而非估计全国代表性数值。为了跨地区可比，我们对每个心理构念做标准化，得到 $z$ 分数，并确保量表方向一致（必要时对反向题进行翻转）。
\end{comment}

\subsection{Region and Cohort Construction}
\label{app:pisa-cohorts}

We next define the cross-regional cohorts used in \S~\ref{sec:pisa-consistency}. The construction parallels the CEPS cohorts but operates on PISA countries and constructs.

\paragraph{Macro-region mapping.}
Each participating country or economy is mapped to one of a small number of macro-regions based on geographical and cultural proximity (e.g., East Asia, Western Europe, Southern Europe, Latin America, Middle East).
Let $\mathcal{R}$ denote the set of regions, and let $\text{region}(i) \in \mathcal{R}$ be the region of student $i$.
The exact country--region mapping table is provided in the project repository and omitted here for space.

\paragraph{Composite achievement quartiles.}
Let $\mathcal{V}_{\text{ach}}$ denote the set of plausible-value columns we use to characterise overall academic achievement (mathematics, reading, science, and mathematics content/process subscales; see Appendix~\ref{app:pisa-vars-preprocessing} for a description of the domains covered).
For each student $i$, we first compute a composite achievement score by averaging all plausible values in this set:
\begin{equation}
  \text{ACHV\_TOTAL}(i)
  = \frac{1}{|\mathcal{V}_{\text{ach}}|}
    \sum_{v \in \mathcal{V}_{\text{ach}}} \text{PV}_{i,v}.
\end{equation}
Within each region $r \in \mathcal{R}$, we then take region-specific quartiles of $\text{ACHV\_TOTAL}$ and assign each student to one of four achievement levels:
\begin{equation}
  \text{ach\_level}(i) \in \{\text{high}, \text{medium}, \text{low}, \text{poor}\}.
\end{equation}
This yields a balanced representation of students at different overall achievement levels within each region, while avoiding domination by a few high-performing systems.
\begin{comment}
这里不再使用单一的 PV1MATH，而是将数学、阅读、科学以及各数学内容/过程子量表的全部 plausible values
（记为集合 $\mathcal{V}_{\text{ach}}$）求平均，得到每个学生的综合学业得分 ACHV\_TOTAL。然后在每个地区内基于 ACHV\_TOTAL 计算四分位，将学生划分为 high / mid-high / mid-low / low 四档，从而在每个地区内部获得相对平衡的学业分布，避免由少数高绩效体系主导。
\end{comment}

\paragraph{Psychological risk.}
To construct a binary psychological-risk indicator, we use the normalised psychological-symptoms index (e.g., \texttt{PSYCHSYM}; see Appendix~\ref{app:pisa-vars-preprocessing}).
Within each region, we define high-risk students as those whose $z$-score on \texttt{PSYCHSYM} lies in the top quartile, and low-risk students as all others:
\begin{equation}
  \text{risk}(i) =
  \begin{cases}
    \text{high}, & z_{i,\text{PSYCHSYM}} \geq Q_{0.75}(r),\\
    \text{low}, & \text{otherwise},
  \end{cases}
\end{equation}
where $Q_{0.75}(r)$ is the 75th percentile of \texttt{PSYCHSYM} $z$-scores in region $r$.
\begin{comment}
心理风险通过标准化后的心理症状指数（如 PSYCHSYM）构建：在每个地区内取该指数 $z$ 分数的上四分位作为高风险阈值 $Q_{0.75}(r)$，将高于该阈值的学生标记为 high risk，其余标记为 low risk。
\end{comment}

\paragraph{Cohort definition.}
Combining achievement level, gender, and psychological risk yields $4 \times 2 \times 2 = 16$ mutually exclusive cohorts per region:
\begin{equation}
  \text{cohort}(i) = \bigl(\text{ach\_level}(i),\ \text{gender}(i),\ \text{risk}(i)\bigr).
\end{equation}
For each region $r$, construct $c$, and cohort $k$, we compute the unweighted mean
\begin{equation}
  \bar{z}_{r,c,k} = \frac{1}{|S_{r,k}|} \sum_{i \in S_{r,k}} z_{i,c},
\end{equation}
where $S_{r,k}$ is the set of students in region $r$ and cohort $k$.
The 16-dimensional vector $(\bar{z}_{r,c,k})_{k=1}^{16}$ is used as the real-student reference pattern for construct $c$ in region $r$.
\begin{comment}
本节定义 PISA 的跨地区群体构建方式，其思想与 CEPS 群体类似，但在国家/地区维度上展开。
首先，将参加 PISA 的国家/地区映射到少量宏观区域（东亚、西欧、南欧、拉美、中东等）。
然后在每个区域内，利用综合学业得分 ACHV\_TOTAL 计算四分位，将学生划分为四档学业水平（high / mid-high / mid-low / low），再结合二元性别与基于 PSYCHSYM 的高/低心理风险，得到每个区域 $4 \times 2 \times 2 = 16$ 个互斥群体。
最后，对每个区域–构念–群体计算非加权均值 $\bar{z}_{r,c,k}$，构成 16 维群体均值向量，作为该地区下的真实学生参考模式。
\end{comment}

\subsection{PISA-based Shadow Survey and Aggregation}
\label{app:pisa-shadow}

The PISA-side shadow survey largely mirrors the CEPS design described in Appendix~\ref{app:ceps-pipeline-preprocessing}, but is tailored to PISA constructs.

\paragraph{Item selection and translation.}
For each selected construct (Appendix~\ref{app:pisa-vars-preprocessing}), we manually pick one to three representative items from the PISA student questionnaire that load strongly on that construct according to OECD documentation.
Items are translated into Chinese while keeping the original response wording and Likert-type scales as close as possible.
Negatively oriented items are reverse-coded so that higher numeric scores consistently indicate more of the underlying construct (e.g., higher anxiety or stronger self-efficacy).

\paragraph{Agent prompting and coding.}
On the agent side, we reuse the immersive role-playing prompt template in Figure~\ref{fig:prompt_template}.
For each region and cohort, we sample a fixed number of personas, instantiate them as DeepSeek-V3.2-based student agents, and ask them to complete the PISA shadow survey. 
Free-text responses are post-processed with a simple rule-based mapper that aligns each answer to one of the discrete PISA response categories, which are then mapped to the official numeric codes used in the PISA scales.

\paragraph{Cohort-level aggregation and metrics.}
For each region, construct, and cohort, we compute the mean coded response across all sampled agents, yielding an agent-side vector of 16 cohort means that is directly comparable to the PISA vector $(\bar{z}_{r,c,k})_{k=1}^{16}$ defined in Appendix~\ref{app:pisa-cohorts}.
We then compute Pearson and Spearman correlations between the human and agent vectors exactly as in the CEPS setting (Appendix~\ref{app:ceps-pipeline-CMdetails}), and report their distributions across regions and constructs in \S~\ref{sec:rq3-pisa}.
\begin{comment}
本节说明 PISA 影子问卷与 Agent 聚合的细节。
对每个构念，我们从 PISA 学生问卷中手工挑选 1–3 个在该构念上负荷较高的条目，翻译为中文并尽量保留原始量表与选项形式；对反向表述条目进行反向编码，使数值含义一致。
Agent 端复用图~\ref{fig:prompt_template} 所示的沉浸式角色扮演提示，在每个地区、每个群体中定额抽取若干画像，实例化为 DeepSeek-V3.2 学生代理，完成 PISA 影子问卷作答。
自由文本回答通过规则映射到离散选项，再转成 PISA 官方的数值编码。
随后，我们在每个地区–构念–群体上取均值，构成 Agent 端 16 维群体均值向量，并与附录~\ref{app:pisa-cohorts} 定义的真实学生向量做皮尔逊和斯皮尔曼相关（公式与 CEPS 部分相同），其分布在第~\ref{sec:results} 节中汇报。

\end{comment}

\subsection{PISA 2022 Constructs and Questionnaire Scales}
\label{app:pisa-constructs}

Table~\ref{tab:pisa-constructs} lists all PISA 2022 student questionnaire scales used in our cross-regional analysis, together with the latent families that we refer to in the main text (math engagement/efficacy, curiosity/growth, classroom climate/belonging, well-being, and workload/work--home balance). Throughout the paper we use the PISA short codes (e.g., \texttt{MATHEFF}, \texttt{CURIOAGR}) to denote these scales; this table provides the full mapping for readers who wish to inspect the underlying constructs in more detail.

\begin{comment}
本小节给出本文在 PISA 2022 数据上使用到的所有学生问卷量表（PISA 官方的短代码），以及它们在正文中被归入的潜在构念家族（如“数学投入/效能”“好奇心/成长性”“课堂氛围/归属感”“幸福感”“负荷与作业–家庭平衡”等）。正文中我们只使用 \texttt{MATHEFF}、\texttt{CURIOAGR} 等短代码指代这些量表，读者如需进一步了解每个量表的具体内涵，可结合本表与 PISA 官方 codebook 查阅。
\end{comment}
\begin{sidewaystable*}[p]
  \centering
  \small
  \caption{PISA 2022 constructs used in the agent--human alignment analysis. Each construct is measured as an OECD-provided index in the student questionnaire; the same construct name is used for human data (e.g., \texttt{MATHEFF}) and its agent counterpart (prefixed by \texttt{K\_}, e.g., \texttt{K\_MATHEFF}). Brief descriptions follow official PISA 2022 variable definitions.}
  \label{tab:pisa-constructs}
  \begin{tabular}{lllp{12cm}}
    \toprule
    Category & Construct & Variable name(s) & Brief description \\
    \midrule

    Workload / practice
      & EXERPRAC  & Human: \texttt{EXERPRAC}; Agent: \texttt{K\_EXERPRAC}
      & Exercise or practice a sport before or after school \\
      & STUDYHMW  & Human: \texttt{STUDYHMW}; Agent: \texttt{K\_STUDYHMW}
      & Studying for school or homework before or after school \\
      & WORKHOME  & Human: \texttt{WORKHOME}; Agent: \texttt{K\_WORKHOME}
      & Working in household/take care of family members before or after school \\
      & WORKPAY   & Human: \texttt{WORKPAY}; Agent: \texttt{K\_WORKPAY}
      & Working for pay before or after school \\
    \midrule

    Math engagement \& efficacy
      & MATHPREF  & Human: \texttt{MATHPREF}; Agent: \texttt{K\_MATHPREF}
      & Preference of Math over other core subjects \\
      & MATHEASE  & Human: \texttt{MATHEASE}; Agent: \texttt{K\_MATHEASE}
      & Perception of Mathematics as easier than other subjects \\
      & MATHMOT   & Human: \texttt{MATHMOT}; Agent: \texttt{K\_MATHMOT}
      & Motivation to do well in mathematics \\
      & MATHEFF   & Human: \texttt{MATHEFF}; Agent: \texttt{K\_MATHEFF}
      & Mathematics self-efficacy: formal and applied mathematics - response options reversed in 2022 (WLE) \\
      & MATHEF21  & Human: \texttt{MATHEF21}; Agent: \texttt{K\_MATHEF21}
      & Mathematics self-efficacy: mathematical reasoning and 21st century skills (WLE) \\
      & MATHPERS  & Human: \texttt{MATHPERS}; Agent: \texttt{K\_MATHPERS}
      & Effort and Persistence in Mathematics (WLE) \\
      & ANXMAT    & Human: \texttt{ANXMAT}; Agent: \texttt{K\_ANXMAT}
      & Mathematics Anxiety (WLE) \\
    \midrule

    Classroom exposure
      & EXPOFA    & Human: \texttt{EXPOFA}; Agent: \texttt{K\_EXPOFA}
      & Exposure to Formal and Applied Mathematics Tasks (WLE) \\
      & EXPO21ST  & Human: \texttt{EXPO21ST}; Agent: \texttt{K\_EXPO21ST}
      & Exposure to Mathematical Reasoning and 21st century mathematics tasks (WLE) \\
    \midrule

    Classroom climate \& belonging
      & RELATST   & Human: \texttt{RELATST}; Agent: \texttt{K\_RELATST}
      & Quality of student-teacher relationships (WLE) \\
      & BELONG    & Human: \texttt{BELONG}; Agent: \texttt{K\_BELONG}
      & Sense of belonging (WLE) \\
      & BULLIED   & Human: \texttt{BULLIED}; Agent: \texttt{K\_BULLIED}
      & Being bullied (WLE) \\
      & SOCCON    & Human: \texttt{SOCCON}; Agent: \texttt{K\_SOCCON}
      & Social Connections: Ease of Communication About Worries and Concerns (WLE) \\
    \midrule

    Curiosity \& growth
      & CURIOAGR  & Human: \texttt{CURIOAGR}; Agent: \texttt{K\_CURIOAGR}
      & Curiosity (agreement) (WLE) \\
      & EMOCOAGR  & Human: \texttt{EMOCOAGR}; Agent: \texttt{K\_EMOCOAGR}
      & Emotional control (agreement) (WLE) \\
      & GROSAGR   & Human: \texttt{GROSAGR}; Agent: \texttt{K\_GROSAGR}
      & Growth Mindset (WLE) \\
    \midrule

    Creativity \& self-efficacy
      & CREATEFF  & Human: \texttt{CREATEFF}; Agent: \texttt{K\_CREATEFF}
      & Creative self-efficacy (WLE) \\
      & CREATOP   & Human: \texttt{CREATOP}; Agent: \texttt{K\_CREATOP}
      & Creativity and Openness to Intellect TBD (WLE) \\
    \midrule

    Well-being / mental health
      & LIFESAT   & Human: \texttt{LIFESAT}; Agent: \texttt{K\_LIFESAT}
      & Students' Life Satisfaction across Domains (WLE) \\
      & PSYCHSYM  & Human: \texttt{PSYCHSYM}; Agent: \texttt{K\_PSYCHSYM}
      & Psychosomatic Symptoms (WLE) \\
    \bottomrule
  \end{tabular}
\end{sidewaystable*}

\section{Additional Evaluation Results}
\label{app:eval-results}

\subsection{Intrinsic Evaluation Results for TAD-PG Persona Collections}
\label{app:intrinsic-eval-results}

\paragraph{Corpus-level summary.}
Table~\ref{tab:intrinsic-summary} reports corpus-level intrinsic metrics for a sample of 151{,}426 HACHIMI personas used to answer RQ1. No persona triggered any hard schema error, and only a very small fraction ($0.06\%$) received soft theoretical warnings.

% 确保你的导言区包含: \usepackage{booktabs} 和 \usepackage{graphicx}

\begin{table}[H]
\centering
% --- 复刻 HEXACO 格式设置 ---
\renewcommand{\arraystretch}{1} % 增加行高，让表格更透气
\setlength{\tabcolsep}{10pt}      % 调整列间距
% ---------------------------

\caption{Summary of intrinsic metrics on 151{,}426 HACHIMI personas.}
\label{tab:intrinsic-summary}

% 使用 resizebox 强制表格宽度适应页面（跨栏宽度）
\resizebox{0.99\textwidth}{!}{
\begin{tabular}{p{5cm} c p{8cm}} % 定义列宽：左侧固定宽，数值居中，右侧注释固定宽
\toprule
\textbf{Metric} & \textbf{Value} & \textbf{Notes} \\
\midrule
\# personas & 151{,}426 & merged corpus size \\
Error rate & 0.0000 & fraction with $\geq 1$ hard error \\
Warning rate & 0.0006 & fraction with $\geq 1$ warning \\
Distinct-1 & 0.4044 & over values/creativity/psychology \\
Distinct-2 & 0.8296 & over values/creativity/psychology \\
Near-duplicate pairs & 0 & SimHash, Hamming $\leq T$ \\
SimHash Hamming (mean) & 27.48 & raw Hamming distance \\
SimHash Hamming (std) & 4.10 & across sampled pairs \\
\bottomrule
\end{tabular}
}
\end{table}

\begin{comment}
本节给出针对 RQ1 的内部评估结果。表~\ref{tab:intrinsic-summary} 汇总了对 151{,}426 条 HACHIMI 画像运行评估器所得的语料级指标。结果显示：不存在任何硬性模式错误（error rate = 0），仅约 0.06\% 的画像被标记了至少一个软性警告。distinct-1/2 分别约为 0.40/0.83，且 SimHash 未发现近重复对，说明整体不存在严重模式坍塌。
\end{comment}

\paragraph{Distributional alignment with quotas.}
To quantify how well the empirical distributions follow the scheduler-specified quotas, we compute KL divergence between the target and empirical distributions for each key categorical variable. On this corpus, the KL values are effectively zero for gender and academic level, and close to zero for grade (Table~\ref{tab:quota-kl}), indicating that the stratified quota mechanism realizes the intended marginal distributions up to negligible sampling noise.

\begin{table}[H]
\centering
% --- 样式设置 ---
\renewcommand{\arraystretch}{1} % 增加行高
\setlength{\tabcolsep}{12pt}      % 增加列间距，让单栏表格不显得太挤
% ---------------

\caption{KL divergence between target quotas and empirical distributions.}
\label{tab:quota-kl}

% \resizebox{\linewidth}{!}{ % 强制适应单栏宽度
\begin{tabular}{l c} 
\toprule
\textbf{Variable} & \textbf{KL Divergence} \\
\midrule
Grade & 0.0001 \\
Gender & 0.0000 \\
Academic level & 0.0000 \\
\bottomrule
\end{tabular}
%}
\end{table}

\begin{comment}
为评估经验分布是否忠实执行了调度器设定的配额，我们对年级、性别与学业水平分别计算目标分布与经验分布之间的 KL 散度。结果几乎为 0（见表~\ref{tab:quota-kl}），说明分层配额采样在这些维度上实现了非常精确的控制。
\end{comment}

\begin{figure}[H]
    \centering
    \includegraphics[width=0.98\textwidth]{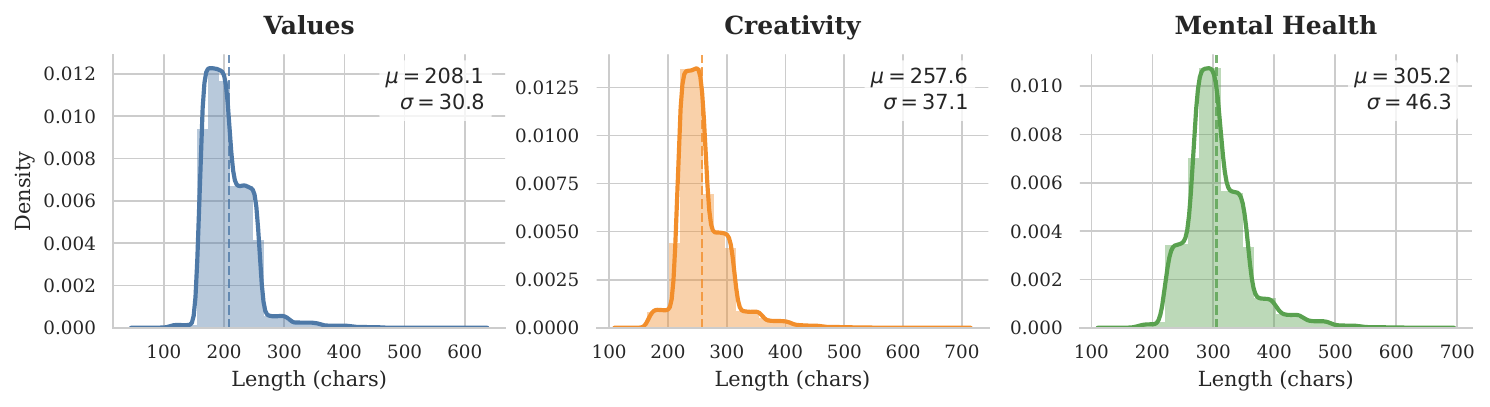}
    \caption{
        Distribution of paragraph lengths (in characters) across the three long-text components. 
        The solid curves represent kernel density estimates (KDE), and the dashed lines indicate the means. 
        The distributions show that the generation process exercises tight control over output lengths, centered around the target specifications.
    }
    \label{fig:para-len-hist}
\end{figure}

\paragraph{Semantic diversity and redundancy.}
Distinct-$n$ statistics in Table~\ref{tab:intrinsic-summary} indicate substantial lexical and phrasal diversity across the three long-text components. The SimHash-based near-duplicate detector did not find any pairs below the Hamming threshold, and the average Hamming distance is roughly 27.5 bits with a standard deviation of 4.1, suggesting that profiles occupy a wide region of the semantic space rather than clustering around a few templates. Paragraph-length summaries for each component (values, creativity, mental health) further show tight control around target ranges with negligible proportions of overly short or long texts (see Figure~\ref{fig:para-len-hist}).

\begin{comment}
语义多样性方面，distinct-$n$ 显示三段长文本在词汇与短语层面具有较高的异质性。SimHash 近重复检测未发现任一低于阈值的画像对，平均汉明距离约为 27.5，比特标准差约为 4.1，表明画像语义空间分布较为分散，并未坍塌到少数模板附近。各组件的段落长度分布（图~\ref{fig:para-len-hist}）则显示文本长度集中在设计区间内，过短或过长的比例可以忽略。
\end{comment}
\paragraph{Anchor alignment with academic level.}
Finally, we summarize the text-derived alignment scores between academic achievement and the three key components (values, creativity, mental health). For each persona, level words are mapped to numeric polarity scores and averaged within each component, then correlated with the four-level academic-achievement anchor. As shown in Table~\ref{tab:anchor-alignment}, all three correlations are positive and substantial, indicating that textual descriptions of high-achieving students tend to emphasize adaptive values, stronger creative profiles, and more favorable mental-health status in line with the TAD-PG design.

\begin{table}[H]
\centering
% --- 样式设置 ---
\renewcommand{\arraystretch}{1}
\setlength{\tabcolsep}{8pt} % 根据内容调整间距
% ---------------

\caption{Text-derived alignment between academic level and component scores.}
\label{tab:anchor-alignment}

%\resizebox{\linewidth}{!}{ % 强制适应单栏宽度
\begin{tabular}{l c}
\toprule
\textbf{Component} & \textbf{Pearson $r$} \\
\midrule
Values & 0.7671 \\
Creativity & 0.8997 \\
Mental health & 0.8380 \\
\bottomrule
\end{tabular}
%}
\end{table}

\begin{comment}
最后，我们报告学业水平与三大长文本组件之间的”锚定对齐”相关系数。对每条画像，我们将价值观、创造力与心理健康段落中的等级用语映射为数值分数，在组件内部求平均后，与四级学业水平锚点计算皮尔逊相关。表~\ref{tab:anchor-alignment} 显示三者相关系数均为显著正向（约 0.77, 0.90, 0.84），说明在整体上，高学业层级的学生会在文本中呈现出更积极/更高水平的价值观、创造力与心理健康画像，符合 TAD-PG 的教育学预期共变关系。
\end{comment}
\subsection{Additional Cross-regional Results on PISA 2022}
\label{app:pisa-results}

To complement the summary in Section~\ref{sec:rq3-pisa}, this appendix provides additional detail on the cross-regional consistency analysis based on PISA 2022. We report results at the level of individual constructs (Table~\ref{tab:pisa-constructs}) and region--construct pairs, and include per-construct plots to allow readers to inspect where the agent-based personas align more or less closely with human group profiles.

Overall distribution across constructs and families. 
Within each region, we computed Pearson $r$ and Spearman $\rho$ between human and agent group means for every construct listed in Table~\ref{tab:pisa-constructs}, using the same 16 gender $\times$ achievement $\times$ risk groups as in the CEPS analysis. 
Figure~\ref{fig:pisa-region-box} in the main text already summarizes these values as distributions by region. 
In the appendix, Figure~\ref{fig:pisa-family-box} further breaks these distributions down by latent family, showing that math engagement/efficacy and curiosity/growth constructs systematically occupy the upper part of the correlation spectrum, whereas classroom climate/belonging constructs are more dispersed and well-being and workload constructs cluster near zero or negative values. 
Across regions, math and curiosity-related scales not only have the highest median correlations but also the smallest interquartile ranges, indicating that the strong alignment observed for \texttt{MATHEFF} and \texttt{CURIOAGR} is representative of a broader family-level pattern rather than being driven by a single favorable scale.

\begin{figure}[t]
  \centering
  % 使用 resizebox 强制将图片宽度缩放至全页文本宽度 (\textwidth)
  \resizebox{0.85\textwidth}{!}{%
    \includegraphics{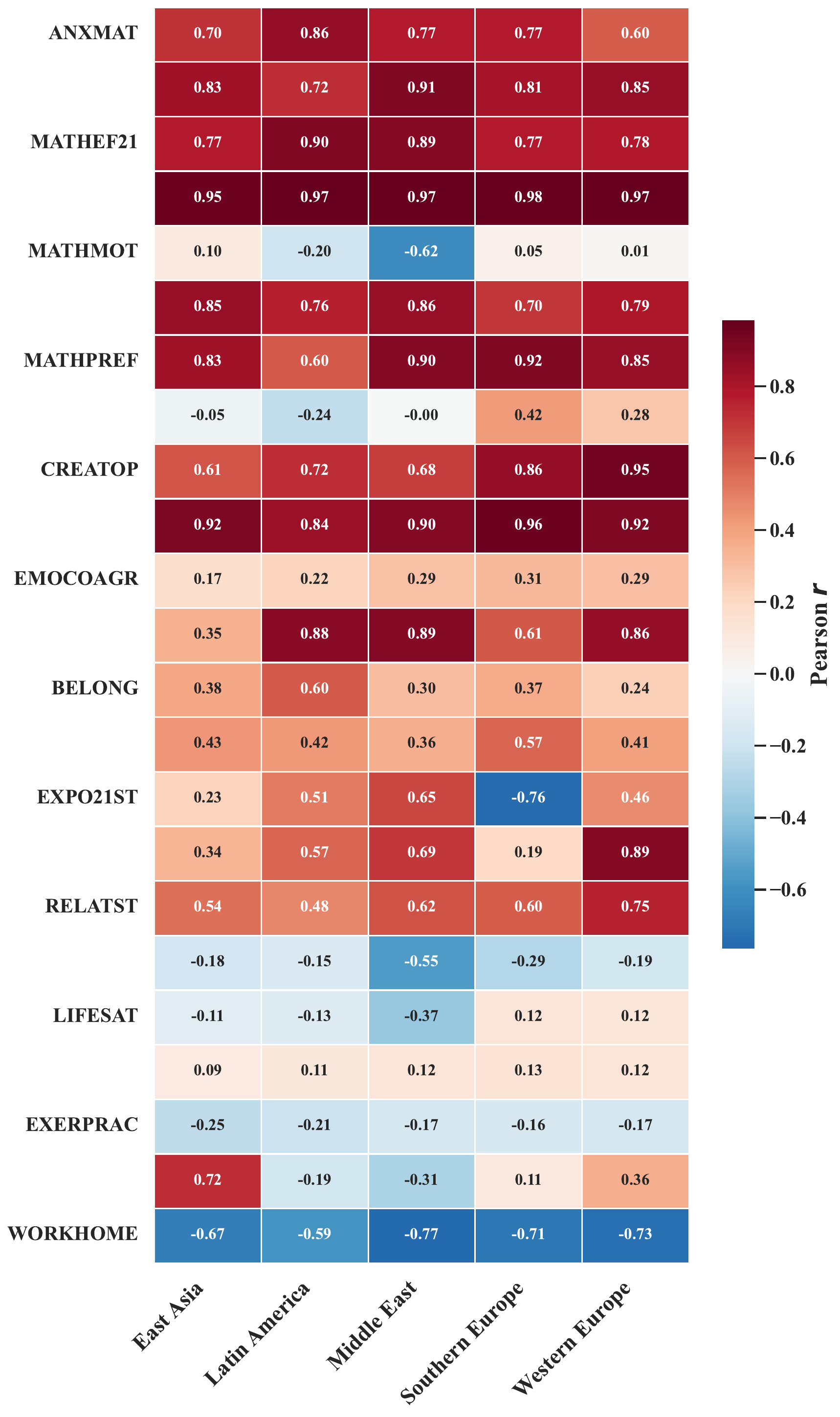}%
  }
  \caption{Heatmap of Pearson correlations between human and agent group means on PISA 2022. Constructs are ordered by latent family, highlighting the high-consistency math/curiosity block and the lower-consistency well-being and workload block.}
  \label{fig:pisa-region-heatmap}
\end{figure}

Regional variation at the construct level. 
Figure~\ref{fig:pisa-region-heatmap} presents a region $\times$ construct heatmap of Pearson correlations, making the construct-level regional differences discussed in Section~\ref{sec:rq3-pisa} more explicit. 
Math engagement and efficacy constructs (\texttt{MATHEFF}, \texttt{MATHEASE}, \texttt{MATHEF21}, \texttt{MATHPERS}, \texttt{MATHPREF}) form a consistently high-correlation block across all five regions, with no region showing systematic degradation relative to the others. 
Curiosity/growth constructs (\texttt{CURIOAGR}, \texttt{GROSAGR}, \texttt{CREATOP}) are positively aligned everywhere but show slightly lower correlations in East Asia and Southern Europe than in Latin America and the Middle East, in line with the regional pattern described in the main text. 
By contrast, mental-health and well-being scales (\texttt{PSYCHSYM}, \texttt{LIFESAT}) tend to hover around $r \approx 0$ in every region, and workload/work--home balance scales (\texttt{WORKHOME}, \texttt{STUDYHMW}, \texttt{EXERPRAC}) often fall into the negative range, indicating that agents systematically under- or over-estimate load for certain human groups. 
Some classroom-exposure variables, such as \texttt{EXPO21ST}, even flip sign between Southern Europe and other regions, illustrating that the agent-based personas can reproduce regional rank orders for academic constructs while remaining less reliable for specific aspects of classroom experience and well-being.

Finally, Table~\ref{tab:pisa-matrix} reports the full set of Pearson and Spearman correlations for all region--construct pairs used in the analysis. 
Taken together with the family- and region-level visualizations, these detailed statistics reinforce the main conclusion in Section~\ref{sec:rq3-pisa}: agent-based personas reproduce a stable cross-regional structure for academic and curiosity-related constructs on PISA 2022, whereas well-being, workload, and certain classroom-exposure constructs exhibit weaker and more region-dependent alignment.

\begin{figure}[t]
  \centering
  % 使用 resizebox 强制将图片宽度缩放至当前栏宽 (\linewidth)
  % 第二个参数 "!" 表示保持长宽比自动计算高度
  \resizebox{0.6\linewidth}{!}{%
    \includegraphics{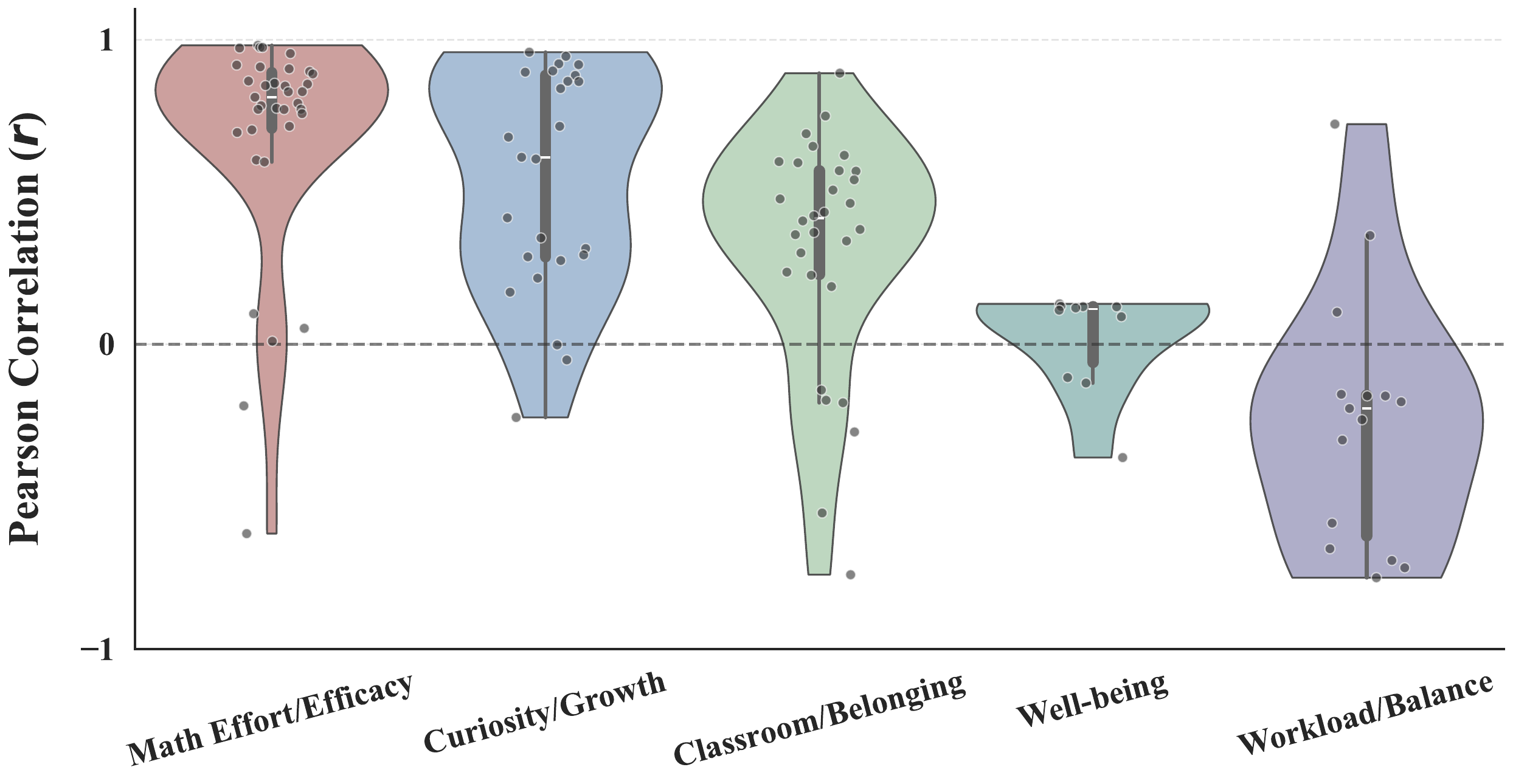}%
  }
  \caption{Distribution of Pearson correlations between human and agent group means on PISA 2022, summarized by construct family (each violin aggregates all constructs within a family across the five regions; individual construct points are jittered).}
  \label{fig:pisa-family-box}
\end{figure}

% 确保你的导言区有这两行：
% \usepackage{booktabs}
% \usepackage{graphicx}
% \usepackage{xcolor}

\begin{table}[t]
  \centering
  \small
  % Define colors to avoid compilation errors
  \definecolor{negc}{gray}{0.6}
  
  \caption{Cross-regional consistency of Pearson correlations ($r$) between human and agent group means on PISA 2022. Constructs are grouped by family. \textbf{Bold} indicates strong alignment ($r \ge 0.80$); \textcolor{negc}{gray} indicates negative alignment. Descriptions follow official PISA 2022 variable definitions.}
  \label{tab:pisa-matrix}
  
  \resizebox{\textwidth}{!}{%
    \begin{tabular}{l p{7cm} ccccc}
      \toprule
      \textbf{Construct} & \textbf{Description} & \textbf{East Asia} & \textbf{S. Europe} & \textbf{Lat. Am.} & \textbf{Mid. East} & \textbf{W. Europe} \\
      \midrule
      
      % --- Math Effort & Efficacy ---
      \multicolumn{7}{l}{\textit{Math Effort \& Efficacy}} \\
      \texttt{MATHEFF}  & Mathematics self-efficacy: formal and applied mathematics - response options reversed in 2022 (WLE) & \textbf{0.95} & \textbf{0.98} & \textbf{0.97} & \textbf{0.97} & \textbf{0.97} \\
      \texttt{MATHEASE} & Perception of Mathematics as easier than other subjects & \textbf{0.83} & \textbf{0.81} & 0.72 & \textbf{0.91} & \textbf{0.85} \\
      \texttt{MATHEF21} & Mathematics self-efficacy: mathematical reasoning and 21st century skills (WLE) & 0.77 & 0.77 & \textbf{0.90} & \textbf{0.89} & 0.78 \\
      \texttt{MATHPREF} & Preference of Math over other core subjects & \textbf{0.83} & \textbf{0.92} & 0.60 & \textbf{0.90} & \textbf{0.85} \\
      \texttt{MATHPERS} & Effort and Persistence in Mathematics  (WLE) & \textbf{0.85} & 0.70 & 0.76 & \textbf{0.86} & 0.79 \\
      \texttt{ANXMAT}   & Mathematics Anxiety (WLE) & 0.70 & 0.77 & \textbf{0.86} & 0.77 & 0.60 \\
      \texttt{MATHMOT}  & Motivation to do well in mathematics & 0.10 & 0.05 & \textcolor{negc}{-0.20} & \textcolor{negc}{-0.62} & 0.01 \\
      \addlinespace[0.5em]
      
      % --- Curiosity & Growth ---
      \multicolumn{7}{l}{\textit{Curiosity \& Growth}} \\
      \texttt{CURIOAGR} & Curiosity (agreement) (WLE) & \textbf{0.92} & \textbf{0.96} & \textbf{0.84} & \textbf{0.90} & \textbf{0.92} \\
      \texttt{EMOCOAGR} & Emotional control (agreement) (WLE) & 0.17 & 0.31 & 0.22 & 0.29 & 0.29 \\
      \texttt{GROSAGR}  & Growth Mindset (WLE) & 0.35 & 0.61 & \textbf{0.88} & \textbf{0.89} & \textbf{0.86} \\
      \texttt{CREATEFF} & Creative self-efficacy (WLE) & \textcolor{negc}{-0.05} & 0.42 & \textcolor{negc}{-0.24} & \textcolor{negc}{-0.00} & 0.28 \\
      \texttt{CREATOP}  & Creativity and Openness to Intellect TBD (WLE) & 0.61 & \textbf{0.86} & 0.72 & 0.68 & \textbf{0.95} \\
      \addlinespace[0.5em]
      
      % --- Classroom & Belonging ---
      \multicolumn{7}{l}{\textit{Classroom \& Belonging}} \\
      \texttt{RELATST}  & Quality of student-teacher relationships (WLE) & 0.54 & 0.60 & 0.48 & 0.62 & 0.75 \\
      \texttt{BELONG}   & Sense of belonging (WLE) & 0.38 & 0.37 & 0.60 & 0.30 & 0.24 \\
      \texttt{BULLIED}  & Being bullied (WLE) & 0.43 & 0.57 & 0.42 &  & 0.41 \\
      \texttt{EXPOFA}   & Exposure to Formal and Applied Mathematics Tasks (WLE) & 0.34 & 0.19 & 0.57 & 0.69 & \textbf{0.89} \\
      \texttt{EXPO21ST} & Exposure to Mathematical Reasoning and 21st century mathematics tasks (WLE) & 0.23 & \textcolor{negc}{-0.76} & 0.51 & 0.65 & 0.46 \\
      \texttt{SOCCON}   & Social Connections: Ease of Communication About Worries and Concerns (WLE) & \textcolor{negc}{-0.18} & \textcolor{negc}{-0.29} & \textcolor{negc}{-0.15} & \textcolor{negc}{-0.55} & \textcolor{negc}{-0.19} \\
      \addlinespace[0.5em]
      
      % --- Well-being ---
      \multicolumn{7}{l}{\textit{Well-being}} \\
      \texttt{PSYCHSYM} & Psychosomatic Symptoms (WLE) & 0.09 & 0.13 & 0.11 & 0.12 & 0.12 \\
      \texttt{LIFESAT}  & Students' Life Satisfaction across Domains (WLE) & \textcolor{negc}{-0.11} & 0.12 & \textcolor{negc}{-0.13} & \textcolor{negc}{-0.37} & 0.12 \\
      \addlinespace[0.5em]
      
      % --- Workload & Balance ---
      \multicolumn{7}{l}{\textit{Workload \& Balance}} \\
      \texttt{STUDYHMW} & Studying for school or homework before or after school & 0.72 & 0.11 & \textcolor{negc}{-0.19} & \textcolor{negc}{-0.31} & 0.36 \\
      \texttt{EXERPRAC} & Exercise or practice a sport before or after school & \textcolor{negc}{-0.25} & \textcolor{negc}{-0.16} & \textcolor{negc}{-0.21} & \textcolor{negc}{-0.17} & \textcolor{negc}{-0.17} \\
      \texttt{WORKHOME} & Working in household/take care of family members before or after school & \textcolor{negc}{-0.67} & \textcolor{negc}{-0.71} & \textcolor{negc}{-0.59} & \textcolor{negc}{-0.77} & \textcolor{negc}{-0.73} \\
      
      \bottomrule
    \end{tabular}%
  }
\end{table}

% =========================================================
\section{Baseline Comparison Results}
\label{app:baseline}

\subsection{Baseline Definition and Evaluation Protocol}
\label{app:baseline-protocol}

\paragraph{Baseline.}
We define a strong yet simple baseline that uses the same backbone LLM but generates each student persona in a \emph{single pass} (i.e., one-shot holistic profiling) without multi-agent factorization, neuro-symbolic validation, quota scheduling, or semantic deduplication.

\paragraph{Protocol parity.}
Unless otherwise noted, all evaluations follow the exact same pipeline and hyper-parameters as HACHIMI:
(i) the same schema parser and intrinsic evaluator (\S\ref{app:intrinsic-eval}),
(ii) the same cohort construction, sampling sizes, and correlation metrics for CEPS (\S\ref{app:ceps-pipeline}) and PISA (\S\ref{app:pisa-pipeline}),
and (iii) the same student-agent instantiation and prompting template (Figure~\ref{fig:prompt_template}).

\subsection{Intrinsic Comparison on 10,000 Personas}
\label{app:baseline-intrinsic}

We report intrinsic metrics on a 10,000-persona subset generated by the baseline and compare them with a matched 10,000-persona subset from HACHIMI, using the same offline evaluator described in Appendix~\ref{app:intrinsic-eval}.

\begin{table*}[t!]
\centering
\renewcommand{\arraystretch}{1}
\setlength{\tabcolsep}{9pt}

\caption{Summary of intrinsic metrics for HACHIMI personas vs. the matched one-shot single-model baseline.}
\label{tab:baseline-intrinsic-10k}

\resizebox{0.99\textwidth}{!}{
\begin{tabular}{p{4.6cm} c c p{6.6cm}}
\toprule
\textbf{Metric} & \textbf{HACHIMI} & \textbf{Baseline} & \textbf{Notes} \\
\midrule
\# personas & 10{,}000 & 10{,}000 & merged corpus size\\
Error rate & 0.0000 & 0.1203 & fraction with $\geq 1$ hard error \\
Warning rate & 0.0082 & 0.2533 & fraction with $\geq 1$ warning \\
Distinct-1 & 0.3285 & 0.2328 & over values/creativity/psychology \\
Distinct-2 & 0.7893 & 0.4589 & over values/creativity/psychology \\
Near-duplicate pairs & 0 & 157 & SimHash, Hamming $\leq T$ \\
SimHash Hamming (mean) & 29.86 & 21.64 & raw Hamming distance \\
SimHash Hamming (std) & 4.02 & 3.65 & across sampled pairs \\
\bottomrule
\end{tabular}
}
\end{table*}

\paragraph{Intrinsic comparison with a one-shot baseline.}
Table~\ref{tab:baseline-intrinsic-10k} compares HACHIMI with a matched one-shot single-model baseline on 10{,}000 personas, evaluated by the same offline intrinsic evaluator and thresholds as in Appendix~\ref{app:intrinsic-eval}.
HACHIMI exhibits substantially stronger structural reliability: the baseline yields a non-trivial hard-error rate (12.03\%) and a high warning rate (25.33\%), while HACHIMI incurs no hard errors and only a small fraction of soft warnings (0.82\%).
This suggests that single-pass holistic generation is prone to format deviations, missing theory-mandated slots, and cross-field inconsistencies, whereas the Propose--Validate--Revise loop effectively enforces the theory-anchored schema.

\paragraph{Diversity and redundancy.}
HACHIMI also shows higher lexical diversity (Distinct-1/2: 0.3285/0.7893 vs.\ 0.2328/0.4589).
Consistently, SimHash indicates stronger redundancy in the baseline: 157 near-duplicate pairs are detected, and the distribution of pairwise Hamming distances shifts lower (mean 21.64 vs.\ 29.86), implying tighter semantic clustering and template reuse.
Overall, modular generation with validation and deduplication yields personas that are both more schema-faithful and more diverse under the same sampling budget.

% --------------------------
\subsection{Full CEPS Results for the Baseline}
\label{app:baseline-ceps}

Table~\ref{tab:baseline-ceps-full} reports item- and construct-level consistency on CEPS Grade~8 by correlating the 16-dimensional cohort-mean vectors between humans and the one-shot baseline agents, using Pearson $r$ and Spearman $\rho$ (Appendix~\ref{app:ceps-pipeline-CMdetails}). We additionally report $\Delta r$ and $\Delta \rho$, computed as (\textsc{HACHIMI} $-$ Baseline), to quantify the relative advantage of our multi-agent persona generation pipeline.

Overall, \textsc{HACHIMI} exhibits consistently stronger human--agent alignment than the one-shot baseline, with improvements on nearly all targets in both rank-based and linear agreement. The gains are most pronounced on socially grounded and relational signals: teacher attention improves from $\rho{=}0.7647$ and $r{=}0.7202$ with $\Delta \rho{=}+0.1324$ and $\Delta r{=}+0.1390$; parent--child relationship quality shows large gains (father--child: $\Delta \rho{=}+0.1877$, $\Delta r{=}+0.2085$; mother--child: $\Delta \rho{=}+0.3309$, $\Delta r{=}+0.2659$). Notably, help-seeking when in trouble improves substantially ($\Delta \rho{=}+0.5362$, $\Delta r{=}+0.6410$), indicating that \textsc{HACHIMI} better preserves cohort-level patterns for behavioral and support-seeking tendencies.

Importantly, \textsc{HACHIMI} also strengthens alignment on key education-facing outcomes: educational aspiration and parental achievement expectations improve with $\Delta \rho{=}+0.2266$/$+0.1848$ and $\Delta r{=}+0.0856$/$+0.1128$, respectively, suggesting that the multi-agent workflow more faithfully captures population-level ordering and magnitude of academic-expectation constructs. Beyond these, \textsc{HACHIMI} yields broad improvements on perceived subject difficulty and psychosocial constructs, including mathematics difficulty ($\Delta \rho{=}+0.0940$, $\Delta r{=}+0.1180$), Chinese difficulty ($\Delta \rho{=}+0.1574$, $\Delta r{=}+0.1355$), depressive symptoms ($\Delta \rho{=}+0.0842$, $\Delta r{=}+0.0388$), and misbehavior frequency ($\Delta \rho{=}+0.0414$, $\Delta r{=}+0.0683$).

Meanwhile, parental strictness remains negatively correlated for both methods (Baseline: $\rho{=}-0.3324$, $r{=}-0.2733$; $\Delta \rho{=}-0.1823$, $\Delta r{=}-0.2295$), highlighting a challenging construct where static personas may not recover the same cohort gradients as the survey. Taken together, these results indicate that \textsc{HACHIMI} more reliably reproduces CEPS cohort-level structure than the one-shot baseline, especially for constructs tied to classroom experience, relationships, and observable behaviors.

% Preamble (if not already):
% \usepackage{rotating}
% \usepackage{xstring}
% \usepackage{xcolor}

% ---------- helper for delta coloring ----------
% +... -> blue, -... -> red, otherwise (0 / 0.00 / +0.00) -> black
\newcommand{\deltac}[1]{%
  \IfBeginWith{#1}{-}{\textcolor{red}{#1}}{%
    \IfBeginWith{#1}{+}{%
      \IfStrEq{#1}{+0}{#1}{%
        \IfStrEq{#1}{+0.0}{#1}{%
          \IfStrEq{#1}{+0.00}{#1}{%
            \IfStrEq{#1}{+0.000}{#1}{\textcolor{blue}{#1}}%
          }%
        }%
      }%
    }{#1}%
  }%
}

\begin{sidewaystable*}[p] % [p] 更常用：单独浮动到整页；也可用 [t]
\centering
\small
\renewcommand{\arraystretch}{1.05}
\setlength{\tabcolsep}{8pt}
\caption{Full CEPS consistency results for the baseline. We report Pearson $r$ and Spearman $\rho$ between human and baseline agent cohort means (16 cohorts). We additionally report $\Delta r$ and $\Delta \rho$ computed as (\textsc{HACHIMI} $-$ Baseline) on the same targets.}
\label{tab:baseline-ceps-full}

\resizebox{0.98\textheight}{!}{ % 旋转后用 textheight 控制“横向宽度”
\begin{tabular}{l p{4.2cm} c c c c c p{4.8cm}}
\toprule
\textbf{Target} & \textbf{Label / Description} & \textbf{Spearman $\rho$} & \textbf{Pearson $r$} & \textbf{$\Delta \rho$} & \textbf{$\Delta r$} & \textbf{$p_\rho$} & \textbf{Human / Agent mean range} \\
\midrule
% -------- Constructs --------
Construct\_Teacher\_Attention\_Avg & Teacher attention (avg) & 0.7647 & 0.7202 & $\deltac{+0.1324}$ & $\deltac{+0.1390}$ & $1.54\times 10^{-5}$ & 2.6-3.0 / 2.1-3.2 \\
Construct\_Depression & Depressive symptoms (sum) & 0.2276 & 0.2859 & $\deltac{+0.0842}$ & $\deltac{+0.0388}$ & 0.0678 & 16.8-32.3 / 18.1-26.5 \\
Construct\_Parental\_Strictness & Parental strictness (avg) & -0.3324 & -0.2733 & $\deltac{-0.1823}$ & $\deltac{-0.2295}$ & 0.2085 & 2.7-2.9 / 2.1-2.3 \\
Construct\_Prosocial & Prosocial behaviour (avg) & 0.4651 & 0.4684 & $\deltac{+0.1673}$ & $\deltac{+0.0935}$ & 0.0225 & 3.4-3.8 / 3.2-4.5 \\
Construct\_Misbehavior & Misbehaviour frequency (avg) & 0.4615 & 0.3555 & $\deltac{+0.0414}$ & $\deltac{+0.0683}$ & 0.0236 & 1.5-2.3 / 1.2-1.8 \\
Construct\_School\_Bonding & School bonding (avg) & 0.1676 & 0.1807 & $\deltac{+0.1853}$ & $\deltac{+0.0906}$ & 0.3163 & 2.6-3.2 / 3.2-4.0 \\
\midrule
% -------- Single items --------
w2b18 & Educational aspiration & 0.7469 & 0.7798 & $\deltac{+0.2266}$ & $\deltac{+0.0856}$ & $2.77\times 10^{-8}$ & 6.4-7.9 / 6.5-8.2 \\
w2a27 & Parental achievement expectations & 0.7168 & 0.8493 & $\deltac{+0.1848}$ & $\deltac{+0.1128}$ & $1.13\times 10^{-4}$ & 1.7-2.5 / 1.4-3.3 \\
w2b21 & Future confidence & 0.5481 & 0.5667 & $\deltac{+0.0773}$ & $\deltac{+0.1070}$ & 0.0047 & 2.7-3.4 / 2.5-4.0 \\
w2a29 & Parental-expectation pressure & 0.4421 & 0.5244 & $\deltac{+0.0068}$ & $\deltac{+0.0244}$ & 0.0073 & 2.6-3.3 / 2.0-3.0 \\
w2a22 & Father--child relationship quality & 0.2004 & 0.2434 & $\deltac{+0.1877}$ & $\deltac{+0.2085}$ & 0.2583 & 2.2-2.7 / 2.9-3.0 \\
w2a23 & Mother--child relationship quality & 0.3244 & 0.4599 & $\deltac{+0.3309}$ & $\deltac{+0.2659}$ & 0.4035 & 2.5-2.9 / 2.9-3.0 \\
w2b02 & Perceived difficulty in mathematics & 0.7119 & 0.7380 & $\deltac{+0.0940}$ & $\deltac{+0.1180}$ & $1.34\times 10^{-4}$ & 1.8-3.2 / 1.2-3.2 \\
w2b03 & Perceived difficulty in Chinese & 0.2102 & 0.2475 & $\deltac{+0.1574}$ & $\deltac{+0.1355}$ & 0.6845 & 2.4-3.1 / 2.1-3.1 \\
w2b04 & Perceived difficulty in English & 0.6794 & 0.7314 & $\deltac{+0.1235}$ & $\deltac{+0.1139}$ & $3.72\times 10^{-4}$ & 1.7-3.0 / 1.2-3.1 \\
w2c04 & Overall self-rated health & 0.1588 & 0.1752 & $\deltac{+0.0093}$ & $\deltac{+0.0370}$ & 0.3331 & 3.4-4.1 / 3.5-4.8 \\
w2d13 & Help-seeking when in trouble & -0.3091 & -0.4001 & $\deltac{+0.5362}$ & $\deltac{+0.6410}$ & 0.2441 & 2.1-3.2 / 1.6-2.0 \\
\bottomrule
\end{tabular}
}
\end{sidewaystable*}

% --------------------------
\subsection{Full PISA 2022 Results for the Baseline}
\label{app:baseline-pisa}

Table~\ref{tab:baseline-pisa-matrix} reports region-wise consistency on PISA~2022 by correlating the cohort-mean vectors between humans and the one-shot baseline agents, using Pearson $r$ across the same five regions as in Appendix~\ref{app:pisa-pipeline}. We additionally report per-region $\Delta$, computed as (\textsc{HACHIMI} $-$ Baseline), to quantify the relative advantage of our multi-agent persona generation pipeline.

Overall, \textsc{HACHIMI} exhibits consistently stronger human--agent alignment than the one-shot baseline on most education-facing attitudes and classroom relationship constructs, with improvements that are broadly stable across regions. The gains are most pronounced on math self-beliefs and growth-oriented constructs: perceived ease of mathematics (\texttt{MATHEASE}) improves substantially from baseline $r{=}0.45$--$0.63$ with consistently large gains ($\Delta{=}{+}0.27$--${+}0.29$) across \emph{all} regions; mathematics self-efficacy (\texttt{MATHEFF}) is already strong under the baseline ($r{=}0.82$--$0.85$) and is further strengthened by \textsc{HACHIMI} ($\Delta{=}{+}0.13$--${+}0.14$). Similarly, 21st-century math efficacy (\texttt{MATHEF21}) improves from $r{=}0.62$--$0.77$ with $\Delta{=}{+}0.08$--${+}0.15$, preference for mathematics (\texttt{MATHPREF}) improves from $r{=}0.47$--$0.74$ with $\Delta{=}{+}0.11$--${+}0.18$, and persistence on difficult math tasks (\texttt{MATHPERS}) improves from $r{=}0.62$--$0.74$ with $\Delta{=}{+}0.08$--${+}0.17$. Beyond math attitudes, \textsc{HACHIMI} yields broad improvements on growth and creativity-related self-perceptions: growth mindset (\texttt{GROSAGR}) gains are substantial in multiple regions (e.g., S.\ Europe $\Delta_{\mathrm{SE}}{=}{+}0.28$, W.\ Europe $\Delta_{\mathrm{WE}}{=}{+}0.22$), and creative self-efficacy (\texttt{CREATEFF}) improves consistently across all regions, mitigating negative or near-zero baseline correlations (e.g., East Asia: $r{=}{-}0.15$ with $\Delta_{\mathrm{EA}}{=}{+}0.10$; Lat.\ Am.: $r{=}{-}0.38$ with $\Delta_{\mathrm{LA}}{=}{+}0.14$; Mid.\ East: $r{=}{-}0.26$ with $\Delta_{\mathrm{ME}}{=}{+}0.26$; W.\ Europe: $r{\approx}0$ with $\Delta_{\mathrm{WE}}{=}{+}0.29$). Importantly, \textsc{HACHIMI} also strengthens alignment on classroom relational signals: student--teacher relationship quality (\texttt{RELATST}) improves from baseline $r{=}0.33$--$0.46$ with consistently positive gains ($\Delta{=}{+}0.13$--${+}0.29$), suggesting that the multi-agent workflow better preserves cohort-level patterns tied to classroom experience and interpersonal dynamics.

Meanwhile, several constructs remain challenging and exhibit limited or mixed improvements. Motivation to do well in mathematics (\texttt{MATHMOT}) is unstable under the baseline (near-zero or negative in some regions, e.g., Lat.\ Am.\ $r{=}{-}0.28$, Mid.\ East $r{=}{-}0.63$), and \textsc{HACHIMI} yields only marginal gains in most regions ($\Delta{=}{+}0.01$--${+}0.08$) and even a slight decrease in W.\ Europe ($\Delta_{\mathrm{WE}}{=}{-}0.02$). Exposure-related classroom constructs are also heterogeneous: while \texttt{EXPOFA} improves strongly in Lat.\ Am.\ ($\Delta_{\mathrm{LA}}{=}{+}0.19$), \texttt{EXPO21ST} remains particularly difficult, including a strongly negative baseline in S.\ Europe ($r{=}{-}0.79$) with only a small improvement ($\Delta_{\mathrm{SE}}{=}{+}0.03$), and slight regressions in East Asia and W.\ Europe ($\Delta_{\mathrm{EA}}{=}{-}0.02$, $\Delta_{\mathrm{WE}}{=}{-}0.03$). Notably, social connectedness (\texttt{SOCCON}) stays negative or near-zero under both methods and often regresses (e.g., $\Delta_{\mathrm{LA}}{=}{-}0.12$, $\Delta_{\mathrm{ME}}{=}{-}0.29$, $\Delta_{\mathrm{WE}}{=}{-}0.19$), highlighting a boundary where static personas struggle to recover cohort gradients for latent or socially embedded well-being signals. Consistently, well-being and time-use variables show weak alignment and/or regression: psychosomatic symptoms (\texttt{PSYCHSYM}) remain close to zero ($r{=}0.09$--$0.13$) with negligible changes, life satisfaction (\texttt{LIFESAT}) shows mixed deltas including notable decreases in Mid.\ East ($\Delta_{\mathrm{ME}}{=}{-}0.23$) and Lat.\ Am.\ ($\Delta_{\mathrm{LA}}{=}{-}0.11$), and workload/balance measures (\texttt{EXERPRAC}, \texttt{WORKHOME}) consistently regress across regions (e.g., $\Delta$ down to ${-}0.29$ and ${-}0.32$, respectively). Taken together, these results indicate that \textsc{HACHIMI} more reliably reproduces PISA cohort-level structure than the one-shot baseline, especially for math attitudes, growth/creativity self-beliefs, and classroom relationship constructs, while social connectedness, well-being, and time-allocation patterns remain challenging for persona-based inference at population scale.

% --------------------------
% Rotated (landscape) version, without \texttt
\begin{sidewaystable*}[p]
\centering
\small
\definecolor{negc}{gray}{0.6}

% optional helper macros (local)
\newcommand{\dpos}[1]{\textcolor{blue}{#1}}
\newcommand{\dneg}[1]{\textcolor{red}{#1}}
\newcommand{\rneg}[1]{\textcolor{negc}{#1}}

\caption{Baseline alignment on PISA 2022: Pearson correlations ($r$) between human and baseline agent cohort means across regions, with per-region $\Delta$ indicating HACHIMI$-$baseline. Bold indicates $r \ge 0.80$; gray indicates negative alignment. $\Delta>0$ is blue and $\Delta<0$ is red.}
\label{tab:baseline-pisa-matrix}

\resizebox{\textheight}{!}{%
\begin{tabular}{l p{10cm} *{10}{c}}
\toprule
\textbf{Construct} & \textbf{Description (official)} &
\textbf{East Asia} & \textbf{$\Delta_{\mathrm{EA}}$} &
\textbf{S. Europe} & \textbf{$\Delta_{\mathrm{SE}}$} &
\textbf{Lat. Am.} & \textbf{$\Delta_{\mathrm{LA}}$} &
\textbf{Mid. East} & \textbf{$\Delta_{\mathrm{ME}}$} &
\textbf{W. Europe} & \textbf{$\Delta_{\mathrm{WE}}$} \\
\midrule

\multicolumn{12}{l}{\textit{Math Effort \& Efficacy}} \\
MATHEFF  & Mathematics self-efficacy: formal and applied mathematics - response options reversed in 2022 (WLE)
  & \textbf{0.82} & \dpos{+0.13}
  & \textbf{0.85} & \dpos{+0.13}
  & \textbf{0.84} & \dpos{+0.13}
  & \textbf{0.84} & \dpos{+0.13}
  & \textbf{0.83} & \dpos{+0.14} \\
MATHEASE & Perception of Mathematics as easier than other subjects
  & 0.55 & \dpos{+0.28}
  & 0.52 & \dpos{+0.29}
  & 0.45 & \dpos{+0.27}
  & 0.63 & \dpos{+0.28}
  & 0.57 & \dpos{+0.28} \\
MATHEF21 & Mathematics self-efficacy: mathematical reasoning and 21st century skills (WLE)
  & 0.62 & \dpos{+0.15}
  & 0.69 & \dpos{+0.08}
  & 0.76 & \dpos{+0.14}
  & 0.77 & \dpos{+0.12}
  & 0.67 & \dpos{+0.11} \\
MATHPREF & Preference of Math over other core subjects
  & 0.72 & \dpos{+0.11}
  & 0.74 & \dpos{+0.18}
  & 0.47 & \dpos{+0.13}
  & 0.72 & \dpos{+0.18}
  & 0.69 & \dpos{+0.16} \\
MATHPERS & Effort and Persistence in Mathematics (WLE)
  & 0.74 & \dpos{+0.11}
  & 0.62 & \dpos{+0.08}
  & 0.65 & \dpos{+0.11}
  & 0.70 & \dpos{+0.16}
  & 0.62 & \dpos{+0.17} \\
ANXMAT   & Mathematics Anxiety (WLE)
  & 0.62 & \dpos{+0.08}
  & 0.71 & \dpos{+0.06}
  & \textbf{0.80} & \dpos{+0.06}
  & 0.69 & \dpos{+0.08}
  & 0.59 & \dpos{+0.01} \\
MATHMOT  & Motivation to do well in mathematics
  & 0.05 & \dpos{+0.05}
  & 0.01 & \dpos{+0.04}
  & \rneg{-0.28} & \dpos{+0.08}
  & \rneg{-0.63} & \dpos{+0.01}
  & 0.03 & \dneg{-0.02} \\
\addlinespace[0.6em]

\multicolumn{12}{l}{\textit{Curiosity \& Growth}} \\
CURIOAGR & Curiosity (agreement) (WLE)
  & \textbf{0.81} & \dpos{+0.11}
  & \textbf{0.94} & \dpos{+0.02}
  & \textbf{0.81} & \dpos{+0.03}
  & \textbf{0.89} & \dpos{+0.01}
  & \textbf{0.89} & \dpos{+0.03} \\
GROSAGR  & Growth Mindset (WLE)
  & 0.29 & \dpos{+0.06}
  & 0.33 & \dpos{+0.28}
  & 0.75 & \dpos{+0.13}
  & 0.73 & \dpos{+0.16}
  & 0.64 & \dpos{+0.22} \\
CREATOP  & Creativity and Openness to Intellect TBD (WLE)
  & 0.37 & \dpos{+0.24}
  & \textbf{0.86} & +0.00
  & 0.70 & \dpos{+0.02}
  & 0.65 & \dpos{+0.03}
  & \textbf{0.89} & \dpos{+0.06} \\
CREATEFF & Creative self-efficacy (WLE)
  & \rneg{-0.15} & \dpos{+0.10}
  & 0.15 & \dpos{+0.27}
  & \rneg{-0.38} & \dpos{+0.14}
  & \rneg{-0.26} & \dpos{+0.26}
  & \rneg{-0.01} & \dpos{+0.29} \\
EMOCOAGR & Emotional control (agreement) (WLE)
  & 0.15 & \dpos{+0.02}
  & 0.31 & +0.00
  & 0.20 & \dpos{+0.02}
  & 0.27 & \dpos{+0.02}
  & 0.35 & \dneg{-0.06} \\
\addlinespace[0.6em]

\multicolumn{12}{l}{\textit{Classroom \& Belonging}} \\
EXPOFA   & Exposure to Formal and Applied Mathematics Tasks (WLE)
  & 0.37 & \dneg{-0.03}
  & 0.17 & \dpos{+0.02}
  & 0.38 & \dpos{+0.19}
  & 0.66 & \dpos{+0.03}
  & \textbf{0.84} & \dpos{+0.05} \\
RELATST  & Quality of student-teacher relationships (WLE)
  & 0.41 & \dpos{+0.13}
  & 0.41 & \dpos{+0.19}
  & 0.33 & \dpos{+0.15}
  & 0.46 & \dpos{+0.16}
  & 0.46 & \dpos{+0.29} \\
EXPO21ST & Exposure to Mathematical Reasoning and 21st century mathematics tasks (WLE)
  & 0.25 & \dneg{-0.02}
  & \rneg{-0.79} & \dpos{+0.03}
  & 0.40 & \dpos{+0.11}
  & 0.61 & \dpos{+0.04}
  & 0.49 & \dneg{-0.03} \\
BULLIED  & Being bullied (WLE)
  & 0.36 & \dpos{+0.07}
  & 0.54 & \dpos{+0.03}
  & 0.42 & +0.00
  & 0.37 & \dneg{-0.01}
  & 0.36 & \dpos{+0.05} \\
BELONG   & Sense of belonging (WLE)
  & 0.36 & \dpos{+0.02}
  & 0.39 & \dneg{-0.02}
  & 0.54 & \dpos{+0.06}
  & 0.25 & \dpos{+0.05}
  & 0.24 & +0.00 \\
SOCCON   & Social Connections: Ease of Communication About Worries and Concerns (WLE)
  & \rneg{-0.18} & +0.00
  & \rneg{-0.22} & \dneg{-0.07}
  & \rneg{-0.03} & \dneg{-0.12}
  & \rneg{-0.26} & \dneg{-0.29}
  & 0.00 & \dneg{-0.19} \\
\addlinespace[0.6em]

\multicolumn{12}{l}{\textit{Well-being}} \\
PSYCHSYM & Psychosomatic Symptoms (WLE)
  & 0.09 & +0.00
  & 0.10 & \dpos{+0.03}
  & 0.11 & +0.00
  & 0.13 & \dneg{-0.01}
  & 0.11 & \dpos{+0.01} \\
LIFESAT  & Students' Life Satisfaction across Domains (WLE)
  & \rneg{-0.07} & \dneg{-0.04}
  & 0.10 & \dpos{+0.02}
  & \rneg{-0.02} & \dneg{-0.11}
  & \rneg{-0.14} & \dneg{-0.23}
  & 0.02 & \dpos{+0.10} \\
\addlinespace[0.6em]

\multicolumn{12}{l}{\textit{Workload \& Balance}} \\
STUDYHMW & Studying for school or homework before or after school
  & 0.68 & \dpos{+0.04}
  & 0.10 & \dpos{+0.01}
  & \rneg{-0.20} & \dpos{+0.01}
  & \rneg{-0.32} & \dpos{+0.01}
  & 0.36 & +0.00 \\
EXERPRAC & Exercise or practice a sport before or after school
  & \rneg{-0.15} & \dneg{-0.10}
  & \rneg{-0.10} & \dneg{-0.06}
  & 0.08 & \dneg{-0.29}
  & 0.07 & \dneg{-0.24}
  & 0.03 & \dneg{-0.20} \\
WORKHOME & Working in household/take care of family members before or after school
  & \rneg{-0.35} & \dneg{-0.32}
  & \rneg{-0.42} & \dneg{-0.29}
  & \rneg{-0.45} & \dneg{-0.14}
  & \rneg{-0.56} & \dneg{-0.21}
  & \rneg{-0.48} & \dneg{-0.25} \\
\addlinespace[0.3em]

\bottomrule
\end{tabular}%
}
\end{sidewaystable*}

\section{Use of AI Assistants}
GPT-5 was used to polish the appendix language, focusing on grammar and phrasing. All outputs were reviewed and revised by the authors. No AI tools used for scientific content or experiments.

% =========================================================

\end{document}